\PassOptionsToPackage{table}{xcolor}
\documentclass[sigconf]{acmart}

\usepackage[english]{babel}
\usepackage{blindtext}

\renewcommand\footnotetextcopyrightpermission[1]{}
\acmYear{2020}
\copyrightyear{2020}
\setcopyright{none}
\acmConference[IMC '20]{ACM Internet Measurement Conference}{October 27--29, 2020}{Virtual Event, USA}
\acmBooktitle{ACM Internet Measurement Conference (IMC '20), October 27--29, 2020, Virtual Event, USA}
\acmPrice{}
\acmDOI{10.1145/3419394.3423643}
\acmISBN{978-1-4503-8138-3/20/10}



\usepackage{amsmath}

\usepackage[english]{babel}
\usepackage{array}
\usepackage{booktabs}
\usepackage{multirow}
\usepackage{graphicx,subfigure}
\usepackage{float}

\newcommand{\Section}{\S}

\usepackage{enumitem,cleveref}

\newcommand{\myparatight}[1]{\noindent{\bf {#1}:}~}

\newcommand{\myparatightest}[1]{\noindent\textbf{{#1:}}~}

\newcounter{packednmbr}

\newenvironment{packeditemize}{\begin{list}{$\bullet$}{\setlength{\itemsep}{0.5pt}\addtolength{\labelwidth}{-4pt}\setlength{\leftmargin}{\labelwidth}\setlength{\listparindent}{\parindent}\setlength{\parsep}{1pt}\setlength{\topsep}{0pt}}}{\end{list}}

\usepackage{amsfonts}
\usepackage{float}

\newcommand{\Eb}{\mathbb{E}}

\newcommand{\bra}[1]{\left( #1 \right)}
\newcommand{\brb}[1]{\left[ #1 \right]}

\newcommand{\brcc}[1]{\big\{ #1 \big\}}

\newcommand{\brn}[1]{\left\lVert #1 \right\rVert}

\newcommand{\namelong}{DoppelGANger}
\newcommand{\nameshort}{DG\xspace}
\newcommand{\name}{\nameshort}

\newcommand{\wiki}{Wikipedia Web Traffic}
\newcommand{\wikishort}{WWT}
\newcommand{\cluster}{Google Cluster Usage Traces}
\newcommand{\clustershort}{GCUT}
\newcommand{\fcc}{Measuring Broadband America}
\newcommand{\fccshort}{MBA}

\newcommand{\metadata}{metadata}
\newcommand{\metadatas}{metadata}
\newcommand{\Metadata}{Metadata}
\newcommand{\Metadatas}{Metadata}
\newcommand{\measurement}{measurement}
\newcommand{\measurements}{measurements}
\newcommand{\Measurement}{Measurement}
\newcommand{\Measurements}{Measurements}

\newcommand{\sample}{sample}
\newcommand{\samples}{samples}

\allowdisplaybreaks

\usepackage{mathtools}
\usepackage{amsthm}
\usepackage{nicefrac}
\usepackage{xparse}
\usepackage{theoremref}

\NewDocumentCommand{\rademacher}{ O{\mu} O{m} m }{R_#2^{\bra{#1}}\bra{#3}}

\theoremstyle{plain}

\theoremstyle{definition}

\begin{document}
\title{Using GANs for Sharing Networked Time Series Data: Challenges, Initial Promise, and Open Questions}

\author{Zinan Lin}
\affiliation{%
  \institution{Carnegie Mellon University}
  \city{Pittsburgh} 
  \state{PA} 
  \postcode{15206}
}
\email{zinanl@andrew.cmu.edu}

\author{Alankar Jain}
\affiliation{%
  \institution{Carnegie Mellon University}
  \city{Pittsburgh} 
  \state{PA} 
  \postcode{15206}
}
\email{alankarjain91@gmail.com}

\author{Chen Wang}
\affiliation{%
  \institution{IBM}
  \city{New York} 
  \state{NY} 
  \postcode{10022}
}
\email{Chen.Wang1@ibm.com}

\author{Giulia Fanti}
\affiliation{%
  \institution{Carnegie Mellon University}
  \city{Pittsburgh} 
  \state{PA} 
  \postcode{15206}
}
\email{gfanti@andrew.cmu.edu}

\author{Vyas Sekar}
\affiliation{%
  \institution{Carnegie Mellon University}
  \city{Pittsburgh} 
  \state{PA} 
  \postcode{15206}
}
\email{vsekar@andrew.cmu.edu}

\begin{abstract}
Limited data access is a longstanding barrier to data-driven  research and development in the networked systems community. 
In this work, we explore if and how     generative adversarial networks (GANs)  can  be used to incentivize data sharing by  enabling  a generic framework for sharing synthetic datasets    with minimal expert knowledge. 
As a specific target, our  focus in this paper is on time series datasets with \metadata{} (e.g.,  packet loss rate measurements with corresponding ISPs).  We identify  key challenges of  existing  GAN approaches  for such workloads with respect to fidelity (e.g.,  long-term dependencies, complex multidimensional relationships, mode collapse) and privacy (i.e., existing guarantees are poorly understood and can sacrifice fidelity).  
To improve fidelity, we  design a custom workflow called   \namelong{} (\nameshort) and demonstrate that across  diverse real-world datasets  (e.g.,  bandwidth measurements, cluster requests, web sessions) and  use cases (e.g., structural characterization, predictive modeling, algorithm comparison),  \nameshort achieves up to 43\% better fidelity than baseline models. 
Although we do not resolve the privacy problem in this work, we identify fundamental challenges with both classical notions of privacy and recent advances to improve the privacy properties of GANs, and suggest a potential roadmap for addressing these challenges. By shedding light on the  promise and   challenges, we hope our work can rekindle the conversation on  workflows for data sharing.

\end{abstract} 

\maketitle

\section{Introduction}
\label{sec:intro}

Data-driven techniques~\cite{halevy2009unreasonable} are central to  networking and systems research (e.g.,~\cite{mao2016resource,chen2018auto,grandl2015multi,jiang2016via,liu2017hierarchical,montazeri2018homa,baltrunas2014measuring,sundaresan2017challenges,jiang2016via,bischof2017characterizing}). This approach allows network operators and system designers to explore  design choices driven  by empirical  needs, and enable new data-driven management decisions.
However, in practice,  the   benefits of data-driven research are   restricted to those who possess data. Even when  collaborating stakeholders have plenty to gain (e.g., an ISP may need workload-specific optimizations from an equipment vendor), they are reluctant to share datasets for fear of revealing business secrets and/or violating user privacy.    %
 Notable exceptions aside (e.g., \cite{caida,mcgregor2010ripe}), the issue of data access  continues to be a substantial concern in the networking and systems communities.

One  alternative is for data   holders to create and share {\em synthetic} datasets modeled from real traces. 
 There have been  specific successes in our community where   experts  identify the key factors of specific traces that impact downstream applications and create generative models  using statistical toolkits~\cite{sommers2004self,weigle2006tmix,yin2014burse,li2014cluster,ganapathi2010statistics,melamed1992tes,melamed1993applications,melamed1993overview,melamed1998modeling,vishwanath2009swing,sommers2011efficient,denneulin2004synthetic,antonatos2004generating,juan2014beyond,di2014characterizing}. Unfortunately, this approach requires significant human expertise and does not easily  {\em generalize} across workloads and  use cases. 

The overarching question  for our work is:  Can we  create high-fidelity, easily generalizable synthetic datasets for networking applications that require minimal (if any) human expertise  regarding workload characteristics and downstream tasks? Such a toolkit could enhance the potential of data-driven techniques by making it easier to obtain and share data.

This paper explores if  and how we can leverage recent advances in  generative adversarial networks (GANs) \cite{goodfellow2014generative}. 
The primary benefit GANs offer is the ability to learn  high fidelity representations of high-dimensional relationships in datasets; e.g., as evidenced by the excitement   in generating  photorealistic images \cite{karras2017progressive}, among other applications. A secondary benefit is that GANs  allow  users to {\em flexibly} tune generation 	  (e.g.,   augment  anomalous or sparse events), which would not be possible with raw/anonymized datasets.   

To scope our work, we consider an important and broad class of networking/systems datasets---time series \emph{\measurements{}}, associated with multi-dimensional \emph{\metadata{}}; e.g.,    measurements of physical network properties \cite{bauer2009physical,baltrunas2014measuring}  and datacenter/compute cluster usage measurements \cite{benson2010network,reiss2011google,he2013next}. 

 We identify key  disconnects between existing  GAN approaches  and  our  use case on two fronts:

\begin{packeditemize}
    \item With respect to {\em fidelity}, we observe  key challenges for existing GAN techniques to  capture: (a)   complex correlations between \measurements{} and their associated \metadata{} and (b) long-term correlations within time series,  such as diurnal patterns, which are qualitatively different from those found in images. 
 Furthermore,    on datasets with a highly variable dynamic range  GANs  exhibit severe mode collapse~\cite{srivastava2017veegan,lin2018pacgan,arjovsky2017wasserstein,gulrajani2017improved}, wherein the GAN generated data only covers a few classes of data samples and ignores other modes of the distribution. 
\item Second, in terms of {\em privacy}, we observe that the  privacy properties
of GANs are poorly understood. This is especially important as practitioners are often worried that GANs may be ``memorizing'' the data and inadvertently reveal proprietary information or suffer from deanonymization attacks~\cite{shokri2017membership,srivatsa2012deanonymizing, narayanan2008robust,ohm2009broken,sweeney2000simple,backstrom2007wherefore,li2009tradeoff}.
Furthermore,  existing  
privacy-preserving training techniques
may sacrifice the utility of the data~\cite{abadi2016deep, beaulieu2019privacy,esteban2017real,xie2018differentially, xu2019ganobfuscator, frigerio2019differentially}, and it is not clear if they apply in our context. 
\end{packeditemize}

  Our primary   contribution   is the design  of a practical workflow called \namelong{}   (\nameshort) that   synthesizes domain-specific insights and concurrent advances in the GAN literature to tackle the fidelity challenges. 
  First,  to model correlations between \measurements{} and their \metadata{} (e.g., ISP name or location), \name{} {\em decouples the generation of \metadata{} from time series}  and feeds \metadata{} to the time series generator at each time step, and also introduces an auxiliary discriminator for the \metadata{} generation. 
  This contrasts with conventional approaches where these  are generated jointly. 
  Second, to tackle mode collapse, our  GAN  architecture separately generates randomized max and min limits and a normalized time series, which can then be rescaled back to the realistic range.
 Third,  to capture temporal correlations, \name{} outputs \emph{batched} samples rather than singletons. While this idea has been used  in Markov modeling \cite{gonzalez2005modeling}, 
 its use in GANs is relatively preliminary \cite{salimans2016improved,lin2018pacgan} and  not  studied in the context of  time series generation. 
  Across multiple datasets and use cases,   we find  that \name{}: (1) is able to learn structural microbenchmarks of each dataset better than baseline approaches and  (2) consistently outperforms  baseline algorithms  on downstream tasks, such as training prediction algorithms (e.g.,  predictors trained on \name{} generated data have test accuracies  up to 43\% higher).

 Our secondary contribution is an   exploration of privacy tradeoffs of GANs, which is an open challenge in the ML community as well ~\cite{GANchallengeneurips}.  
 Resolving these tradeoffs is beyond the scope of this work.
 However, we  empirically confirm  that an important class of \emph{membership inference} attacks on privacy can be mitigated by training \name{} on larger datasets.
 This may run counter to conventional  release practices, which advocate releasing smaller datasets to avoid leaking user data \cite{reiss2012obfuscatory}. 
A second positive result is that  we  highlight that the decoupled generation architecture of    \name{} workflow can enable  data holders to hide certain attributes of interest (e.g., some specific \metadata{} may  be proprietary). 
 On the flip side, however, we empirically evaluate recent proposals  for GAN training with differential privacy guarantees  \cite{abadi2016deep, beaulieu2019privacy,esteban2017real,xie2018differentially, xu2019ganobfuscator, frigerio2019differentially}
 and show that these  methods destroy  temporal correlations even for moderate privacy guarantees, highlighting the need for  further research on the  privacy front.

\myparatight{Roadmap} In the rest of the paper, we begin by  discussing use cases  and prior work in \Section\ref{sec:strawman}. We provide background on GANs and  challenges in \Section\ref{sec:problem}. We describe the design of \name{} in   \Section\ref{sec:design} and evaluate it in \Section\ref{sec:evaluation}. We  analyze  privacy tradeoffs in \Section\ref{sec:exp-privacy}, before  concluding in \Section\ref{sec:conclude}.

\section{Motivation and Related Work}
In this section, we discuss motivating scenarios and why existing solutions fail to achieve our goals.
\subsection{Use cases and requirements}
\label{sec:use}

While there are many scenarios for data sharing, we consider two  illustrative examples:  (1) {\em Collaboration across stakeholders:} Consider a network operator collaborating with an equipment vendor to design custom workload optimizations. 
Enterprises often impose restrictions on data access between their own divisions and/or with external vendors due to privacy concerns; and  (2) {\em Reproducible, open research:}
Many research proposals rely on datasets  to test and develop ideas. 
However,  policies and business considerations may preclude  datasets from being shared, thus stymieing    reproducibility.

In  such scenarios, we consider three representative tasks:

\noindent{\em (1) Structural characterization:}
Many system designers also need to understand  temporal and/or geographic trends in systems; e.g.,  to understand the shortcomings in existing  systems and  explore remedial solutions  \cite{baltrunas2014measuring,sundaresan2017challenges,jiang2016via,bischof2017characterizing}. 
 In this case, generated data should preserve trends and distributions well enough to reveal such structural insights.

\noindent {\em (2) Predictive modeling:}
A second use case  is to learn predictive models, especially for tasks like resource allocation \cite{fu2007exploring,jiang2016cfa,li2018predictive}.
For these  models to be useful, they should have enough fidelity that a predictor trained on generated data should  make meaningful predictions on real data.

\noindent {\em (3) Algorithm evaluation:}
The design of  resource allocation algorithms for cluster scheduling  and transport protocol design (e.g.,~\cite{mao2016resource,chen2018auto,grandl2015multi,jiang2016via,liu2017hierarchical,montazeri2018homa}) often needs workload data to tune control parameters. 
 A key property for generated data is that if algorithm A performs better than algorithm B on the real data,  the same should hold on the generated data. 

\myparatight{Scope and goals}
Our focus  is on \emph{multi-dimensional time series} datasets,  common in networking and systems applications. Examples include: {1. \em Web traffic traces} of   webpage views with \metadata{} of   URLs, which  can be used  to predict future  views,  page correlations \cite{srivastava2000web}, or generate  recommendations \cite{nguyen2013web,forsati2010effective};   {2. \em Network measurements} of packet loss rate, bandwidth, delay  with \metadata{} such as location  or device type that are  useful for network management \cite{jiang2016cfa}; or {3. \em Cluster usage measurements} of metrics such as CPU/memory usage associated with \metadata{} (e.g.,   server and job type)  that can  inform  resource provisioning \cite{chaisiri2011optimization} and  job scheduling \cite{maqableh2014job}.  At a high level, each example consists  of time series samples (e.g., bandwidth  measurements) with  high-dimensional data points  and associated \metadata{} that can be either numeric or categorical (e.g., IP address, location).
Notably, we do not handle stateful interactions between agents; e.g.,  generating full TCP session packet captures.

Across these  use cases and datasets, we require techniques that can accurately capture  two sources of diversity:
\emph{(1) Dataset diversity:}
For example, predicting CPU usage is very different from predicting network traffic volumes.
\emph{(2) Use case diversity:}
For example, given a website page view dataset, website category prediction focuses on the dependency between the number of page views and its category, whereas page view modeling only needs the temporal characteristics of page views.
 Manually  designing  generative models  for each use case and dataset is   time consuming and  requires significant human expertise. 
Ideally, we  need   generative techniques that are {\em general}   across  diverse datasets and use cases and achieve high {\em fidelity}.

\subsection{Related work and  limitations}
\label{sec:strawman}

In this section, we  focus on non-GAN-based approaches, and defer  GAN-based approaches  to  \Section\ref{sec:prior-gans}. 
 Most   prior work from the networking domain falls in two categories:  simulation models and expert-driven models. A third  approach involves  
machine learned models (not using GANs).

\myparatight{Simulation models}
These  generate data by building a simulator that mimics a real system or network \cite{sommers2004framework,sommers2011efficient,issariyakul2009introduction,moreno2014analysis,di2015gloudsim,magalhaes2015workload,sliwko2016agocs}.
For example, ns-2 \cite{issariyakul2009introduction} is a widely used simulator for networks and  GloudSim \cite{di2015gloudsim} is  a distributed cloud simulator for generating cloud workload and traces.
In terms of fidelity, this class of models is good if the simulator is very close to real systems. 
However, in reality, it is often hard to configure the parameters to simulate a given target dataset.
 Though some data-driven ways of configuring the parameters have been proposed \cite{moreno2014analysis, di2015gloudsim, magalhaes2015workload, sliwko2016agocs}, it is still difficult to ensure that the simulator itself is close to the real system.
Moreover, they do not generalize across datasets and use cases, because a new simulator is needed for each scenario.

\myparatight{Expert-driven models}
These entail capturing  the data using a  \emph{mathematical} model  instead of using a simulation. Specifically,  domain expects determine which parameters are important and which parametric model we should use. Given a model,   the parameters can be manually configured \cite{cooper2010benchmarking,tarasov2016filebench,fio} or learned from data \cite{sommers2004self,weigle2006tmix,yin2014burse,li2014cluster,ganapathi2010statistics,melamed1992tes,melamed1993applications,melamed1993overview,melamed1998modeling,vishwanath2009swing,sommers2011efficient,denneulin2004synthetic,antonatos2004generating,juan2014beyond,di2014characterizing}.
For example, the Hierarchical Bundling Model models inter-arrival times of datacenter jobs better than the widely-used Poisson process \cite{juan2014beyond}. 
Swing \cite{vishwanath2009swing} extracts statistics of user/app/flow/link (e.g., packet loss rate, inter-session times) from data, and then generate traffic by sampling from the extracted distributions.
 In practice, it is challenging to come up with models and parameters   that achieve high fidelity.
For example, BURSE \cite{yin2014burse} explicitly models the burstiness and self-similarity in cloud computing workloads, but does not consider e.g. nonstationary and long-term correlations \cite{calzarossa2016workload}.
Similarly, work in  cloud job  modeling  \cite{juan2014beyond,yin2014burse} characterizes inter-arrival times, but does not model the correlation between job \metadata{} and inter-arrival times.
Such models struggle to generalize because different datasets and use cases require different models.

\begin{figure}[t]
	\centering
	\includegraphics[width=0.9\linewidth]{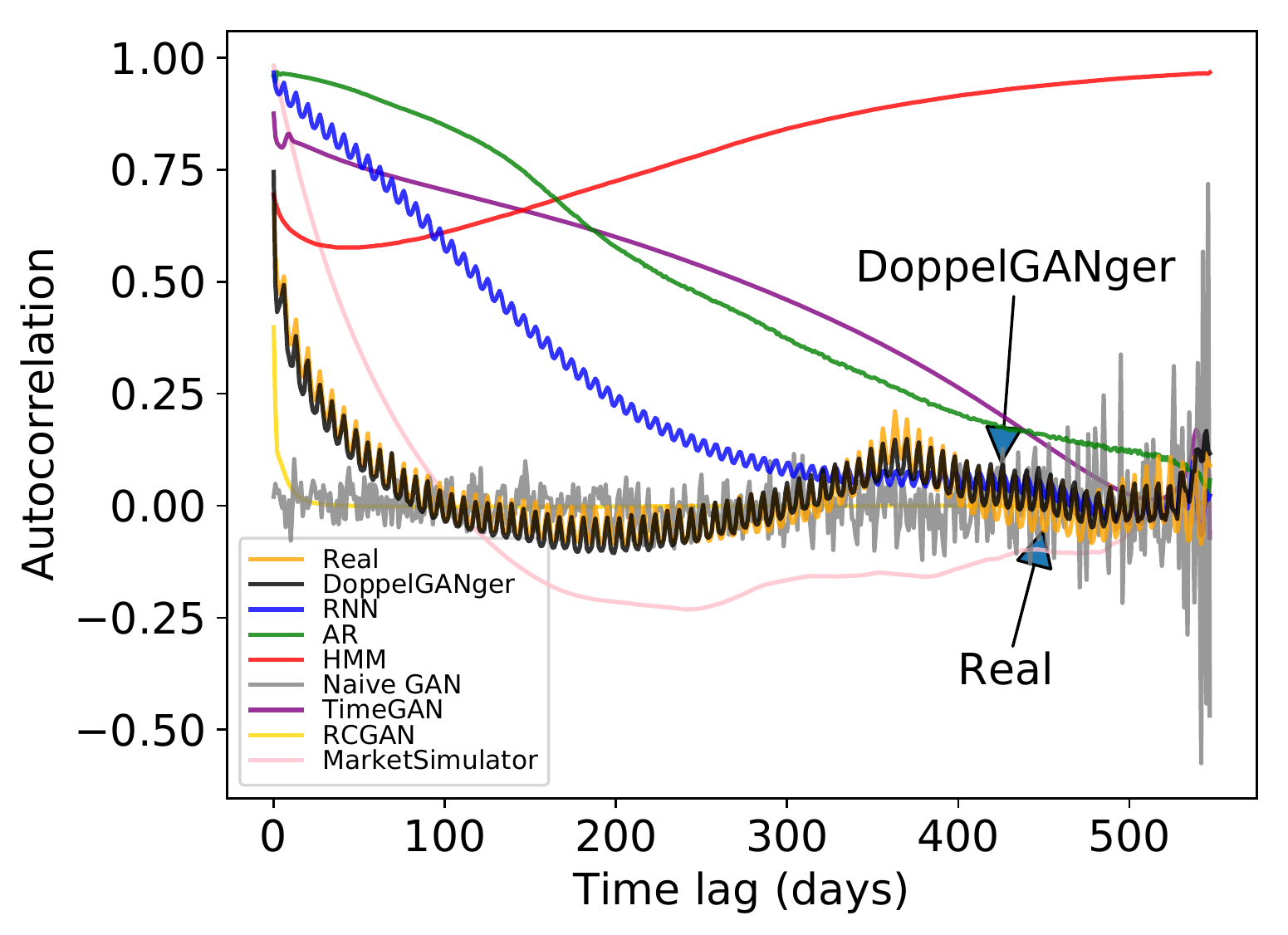}
	\vspace{-0.3cm}
	\caption{Autocorrelation of daily page views for \wiki{} dataset. 
	}
	\vspace{-0.3cm}
	\label{fig:wiki-autocorr}
\end{figure}

\myparatight{Machine-learned models} These are general parametric models, where the parameters can be learned (trained) from data.
 Examples include autoregressive models, Markov models, and recurrent neural networks (RNNs). In theory, these generic statistical models have the potential to  generalize across datasets.
However, they are not general in terms of use cases, because they do not jointly model \metadata{} \emph{and} time series. 
For example, in a network measurement dataset, modeling only the packet loss rate is good for understanding its patterns. 
But if we want to learn which IP prefix has network issues, jointly modeling the \metadata{} (IP address) and the time series (sequence of packet loss rate) is important.  
Additionally, these general-purpose  models have bad fidelity, failing to capture  long-term temporal correlations.
For instance, all these models fail to capture weekly and/or annual patterns in the \wiki~dataset (Figure~\ref{fig:wiki-autocorr}). 
Our point is not to highlight the importance of learning autocorrelations,\footnote{While there are specific tools for estimating autocorrelation, in general this is a hard problem in high dimensions \cite{bai2011estimating,cai2016estimating}.} 
but to show that learning  temporal data distributions (without overfitting to a single statistic, e.g. autocorrelation) is hard.
Note that \name{} is able to learn  weekly and annual correlations  without  special tuning.

\section{Overview} 
\label{sec:problem}
\label{sec:overview}
As we saw, the above  classes of techniques do not achieve good fidelity and generality across datasets and use cases. 
Our overarching  goal is thus to develop a  general framework that can  achieve high fidelity with minimal expertise.

\subsection{Problem formulation}

We abstract the scope of our datasets as follows:
A \emph{dataset} $\mathcal{D}=\brcc{O^1,O^2,\allowbreak...,O^n}$ is defined as a set of \emph{samples} $O^i$ (e.g., the clients). %
Each sample $O^i=(A^i, R^i)$ contains $m$ \emph{\metadatas{}} $A^i=[A^i_1,A^i_2,...,A^i_m]$. For example, \metadata{} $A^i_1$ could represent client $i$'s physical location, and $A^i_2$ the client's ISP.
Note that we can support datasets in which multiple samples have the same set of \metadatas{}. 
The second component of each sample is a time series of \emph{records} $R^i=[R^i_1,R^i_2,...,R^i_{T^i}]$, 
where $R^i_j$ means $j$-th measurement of $i$-th client. Different samples may contain a different number of measurements. 
The number of records for sample $O^i$ is given by $T^i$. 
Each record $R^i_j=(t^i_j,f^i_j)$ contains a \emph{timestamp} $t^i_j$,  and $K$ \emph{\measurements{}} $f^i_j=[f^i_{j,1}, f^i_{j,2},...,f^i_{j,K}]$. For example, $t^i_j$ represents the time when the measurement $f^i_j$ is taken, and $f^i_{j,1}$, $f^i_{j,2}$ represent the ping loss rate and traffic byte counter at this timestamp respectively.
Note that the timestamps are sorted, i.e. $t^i_j<t^i_{j+1}~\forall 1\leq j<T^i$.

This abstraction fits many classes of data that appear in networking applications.
For example, it is able to express web traffic and cluster trace datasets (\Section\ref{sec:evaluation}). Our problem is to take any such dataset as input and learn a model that can generate a new dataset $\mathcal D'$ as output. 
$\mathcal D'$ should exhibit fidelity, and the methodology should be general enough to handle datasets in our abstraction.

\subsection{GANs: background and promise}
\label{sec:gans}
GANs are a data-driven generative modeling technique  \cite{goodfellow2014generative} that  take as input  training data samples and output a model that can produce new samples from the same distribution as the original data. More precisely, if we have a dataset  of $n$ samples $O_1, \ldots, O_n$, where $O_i \in \mathbb{R} ^p$, and each sample is drawn i.i.d. from some distribution $O_i \sim P_O$.
The goal of GANs is to use these samples to learn a model that can draw samples from distribution $P_O$ \cite{goodfellow2014generative}.

\begin{figure}[t]
	\centering
	\begin{minipage}{0.3\textwidth}
		\centering
		\includegraphics[width=\linewidth]{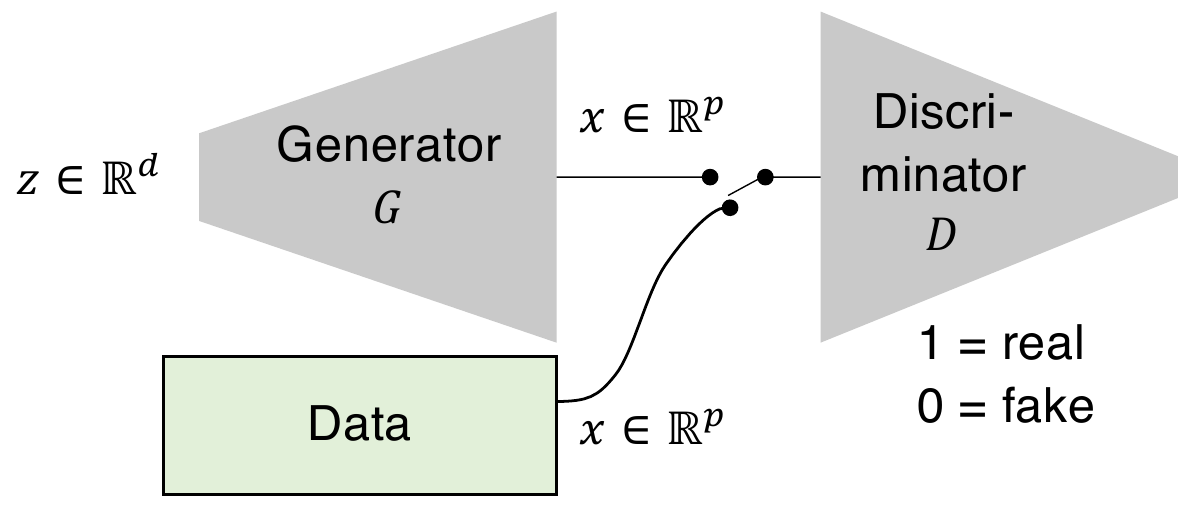}
	\end{minipage}
	\vspace{-0.3cm}
	\caption{
	Original GAN  architecture from \cite{goodfellow2014generative}.}
	\label{fig:gans}
	\label{fig:karras}
	\vspace{-0.5cm}
\end{figure}

 GANs use an adversarial training workflow consisting of a generator $G$ and a discriminator $D$ (Figure \ref{fig:gans}). 
 In practice, both are instantiated with neural networks. 
In the canonical GAN design~\cite{goodfellow2014generative}, the generator maps a noise vector $z\in \mathbb R^d$ to a sample $O \in \mathbb R^p$, where $p \gg d$. 
$z$ is drawn from some pre-specified distribution $P_z$, usually a Gaussian or a uniform.
Simultaneously, we train the discriminator $D: \mathbb R^p \to [0,1]$, which takes samples as input (either real of fake), and classifies each sample as real (1) or fake (0). 
Errors in this classification task are used to train the parameters of both the generator and discriminator through backpropagation.
The loss function for GANs is: 
$
\min_G \max_D \mathbb E_{x\sim p_{x}}[\log D(x)] + \mathbb E_{z\sim p_z}[\log(1-D(G(z)))].
\label{eq:gan}
$
The generator and discriminator are trained alternately, or \emph{adversarially}.
Unlike prior  generative modeling approaches which   likelihood maximization of parametric models (e.g., \S\ref{sec:strawman}), 
GANs make fewer assumptions about the  data structure.

Compared with related work in \Section\ref{sec:strawman}, GANs offer three key benefits.  First, similar to the machine learning models,  GANs can be {\em general}  across datasets. The discriminator  is an universal agent for judging the fidelity of generated samples. Thus,   the discriminator only needs  raw samples and it does not need any other information about the system producing the samples. 
Second, %
GANs can be used to generate \measurements{} and \metadatas{} (\S\ref{sec:challenges}). 
 Thus, GANs have the potential to support a wide range of use cases involving   \measurements{}, \metadatas{}, and cross-correlations between them. Finally, GANs have been used in other domains  for  generating realistic high-fidelity  datasets for complex tasks such as  images  \cite{karras2017progressive},   text \cite{fedus2018maskgan,yu2017seqgan},  and music \cite{mogren2016c,dong2018musegan}.

\subsection{Using GANs to generate time series}

\subsubsection{Prior Work.}
\label{sec:prior-gans}
Using GANs to generate time series is  a popular idea  \cite{fedus2018maskgan,yu2017seqgan,mogren2016c,dong2018musegan,esteban2017real,zec2019recurrent,timegan,fu2007exploring}.
Among the domain-agnostic designs, the generator usually takes prior measurements (generated or real) and noise as inputs and  outputs one measurement at a time  \cite{fu2007exploring,esteban2017real,zec2019recurrent,timegan,yu2017seqgan}. 
These works typically change two aspects of the GAN: the architecture \cite{fu2007exploring,esteban2017real,zec2019recurrent}, the training \cite{yu2017seqgan}, or both \cite{timegan,market-sim}.
The  two most relevant papers to ours are RCGAN \cite{esteban2017real} and TimeGAN \cite{timegan}.
RCGAN is the most similar design to ours; like \name, it uses recurrent neural networks  (RNNs) to generate time series and can condition the generation on \metadata{}. 
However, RCGAN does not itself generate \metadata{} and has little to no evaluation of the correlations across time series and between \metadata{} and \measurements{}. 
We found its fidelity on our datasets to be poor; 
we instead use a different discriminator architecture, loss function, and \measurement{} generation pipeline (\Section\ref{sec:design}).
TimeGAN is the current state-of-the-art, outperforming RCGAN \cite{timegan}.
Like RCGAN, it uses RNNs for both the generator and discriminator.
Unlike RCGAN, it trains an additional neural network that maps time series to vector embeddings, and the generator outputs sequences of  embeddings rather than samples.
Learning to generate transformed or embedded time series is common, both in approaches that rely on  GANs \cite{timegan,corrgan} and those that rely on a different class of generative models called variational autoencoders (VAE) \cite{market-sim}. Our experiments suggest that this approach models long time series poorly (\S\ref{sec:evaluation}).

\subsubsection{Challenges.}
\label{sec:challenges}

Next, we highlight key challenges that  arise in using GANs for our use cases.  While these challenges  specifically stem from our attempts in using GANs in networking- and systems--inspired use cases,  some of these challenges  broadly apply to other use cases as well.

\myparatight{Fidelity challenges} 
 The first  challenge relates to  {\em  long-term temporal correlations}. 
As we see in Figure \ref{fig:wiki-autocorr}, the canonical GAN does poorly in capturing temporal correlations trained on the \wiki{} (\wikishort{}) dataset.\footnote{This uses: (1) a   
dense multilayer perceptron (MLP) which generates \measurements{} and \metadatas{} jointly, (2) an MLP discriminator, and (3)  Wasserstein loss \cite{arjovsky2017wasserstein,gulrajani2017improved},  consistent with prior work~\cite{guibas2017synthetic,choi2017generating,frid2018synthetic,han2018gan}.}   
Concurrent and prior work on using GANs for other time series data has also observed this \cite{fedus2018maskgan,yu2017seqgan, esteban2017real,timegan}.
One approach to address this is  segmenting long datasets into chunks; e.g., 
 TimeGAN~\cite{timegan} chunks datasets into smaller time series each of  24 epochs, and only evaluates the model on  producing new time series of this length  \cite{timegan-code}. 
This is not viable in our domain, as relevant properties of networking/systems data often occur over longer time scales (e.g.,  network measurements) (see Figure \ref{fig:wiki-autocorr}).  Second, \emph{mode collapse} is a well-known problem in GANs where they generate only a few modes of the underlying distribution~\cite{srivastava2017veegan,lin2018pacgan,arjovsky2017wasserstein,gulrajani2017improved}.  It is particularly exacerbated in our time series use cases because of the high variability in the range of \measurement{} values.   Finally, we need to capture \emph{complex  relations between \measurements{} and \metadatas{}} (e.g., packet loss rate and ISP),  and  across different \measurements{} (e.g., packet loss rate and byte counts). 
 As such, state-of-the-art approaches either do not co-generate attributes with time series data or their accuracy for such tasks is  unknown~\cite{esteban2017real,zec2019recurrent,timegan},  and directly   generating joint  \metadata{} with \measurements{} samples
 does not  converge (\Section\ref{sec:evaluation}).   Further,  generating independent time series for each \measurement dimension will break their   correlations.

\myparatight{Privacy challenges} 
In addition to the above fidelity challenges, we also observe key challenges with respect to \emph{privacy}, and reasoning about the
privacy-fidelity tradeoffs of using GANs.  A meta question, not unique to our work, is finding the right definition of privacy for each use case.
  Some    commonly used definitions in the community are  notions of {\em differential privacy} (i.e., how much does any single data point contribute to a model)~\cite{dwork2008differential}  and {\em membership inference} (i.e., was a specific sample in the datasets)~\cite{shokri2017membership}. 
  However, these definitions can hurt fidelity \cite{sankar2013utility,bagdasaryan2019differential} without defending against relevant attacks \cite{fredrikson2014privacy,hitaj2017deep}.
 In our networking/systems use cases, we may also want to  even hide {\em specific features} and  avoid releasinng aggregate  statistical characteristics of proprietary data (e.g., number of users,   server load, meantime to failure). 
 Natural questions arise:    First, can GANs  support these flexible notions of privacy in practice, and if so under what configurations?   Second, there are emerging proposals to extend   GAN training  (e.g.,~\cite{xie2018differentially, xu2019ganobfuscator, frigerio2019differentially}) to offer some privacy guarantees. Are their privacy-fidelity tradeoffs sufficient to be practical for  networking datasets?

\section{\namelong{} Design}

\label{sec:design}

In this section, we describe how we tackle fidelity shortcomings of time series GANs. 
Privacy is discussed in Section~\ref{sec:exp-privacy}.    Recall that existing approaches have issues in capturing temporal effects and  relations between \metadatas{} and \measurements. In what follows, we present our solution starting from the canonical GAN strawman and extend it to address these challenges. 
Finally, we summarize the design and  guidelines  for users to use our workflow.

\subsection{Capturing long-term effects}
\label{sec:design-long}

Recall that the canonical GAN generator architecture is a fully-connected multi-layer perceptron (MLP), which we use in our strawman solution (\S\ref{sec:challenges}). As we saw, this architecture fails to capture long-term correlations (e.g., annual correlations in page views).  

\myparatight{RNN primer and limitations} Similar to prior efforts, we posit that the main   reason is that   MLPs are not well suited for time series.
 A better choice  is to use recurrent neural networks (RNNs), which are   designed to model time series and have been widely used in the GAN literature to generate time series \cite{mogren2016c,esteban2017real,yu2017seqgan,zec2019recurrent,timegan}. Specifically, we use a  variant  of RNN called long short-term memory (LSTM) \cite{hochreiter1997long}.
 
 \begin{figure}[t]
	\centering
	\includegraphics[width=0.55\linewidth]{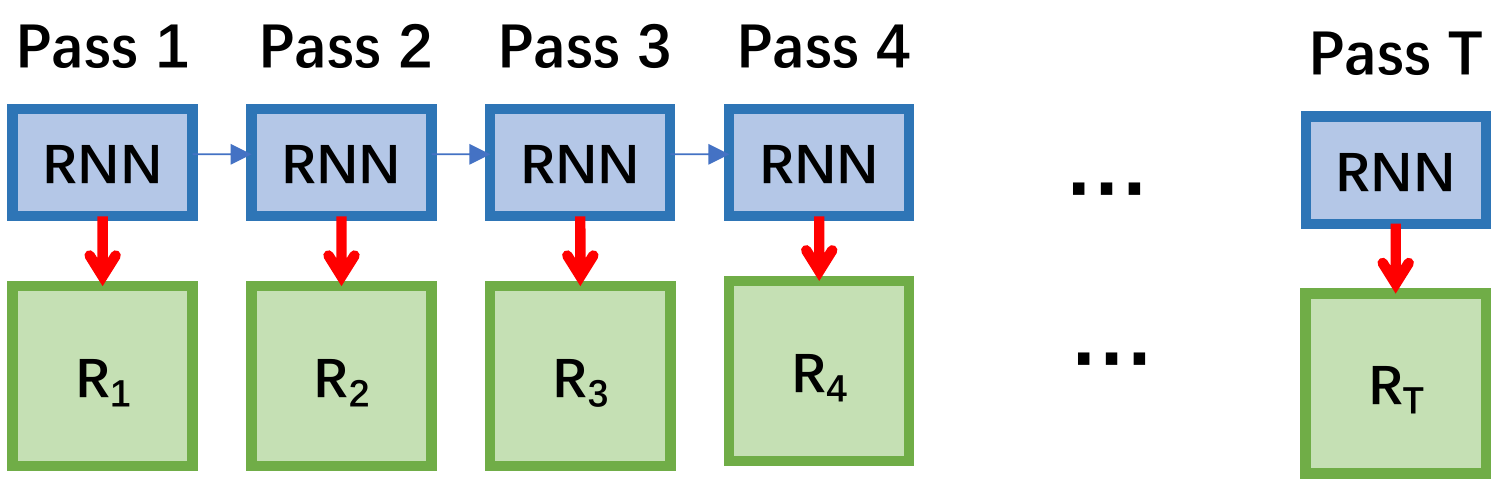}
	\put(-170,20){(a)}
	
	\includegraphics[width=0.55\linewidth]{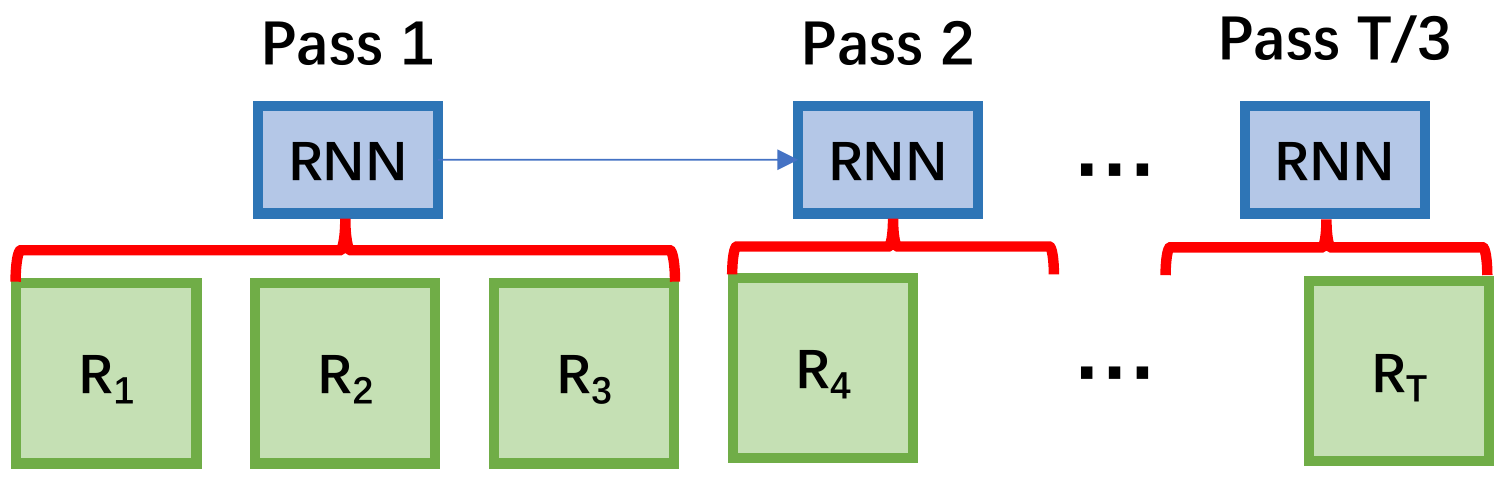}
	\put(-170,20){(b)}
	\vspace{-0.2cm}
	\caption{(a) The usual way of generating time series. (b) Batch generation with $S=3$. The RNN is a single neural network, even though many units are illustrated.  
    This \emph{unrolled} representation  conveys that the RNN is being used  many times to generate samples.}
	\vspace{-0.3cm}
	\label{fig:batch_gen}
\end{figure}
 
At a high level, RNNs work as follows (Figure \ref{fig:batch_gen} (a)). Instead of generating the entire time series at once,  RNNs generate one record $R_j^i$ at a time (e.g., page views on the $j$th day), and then run $T^i$ (e.g., the number of days) passes to generate the entire time series. 
The key difference in a RNN from  traditional neural units is that  RNNs have an internal state that implicitly encodes  all past states of the signal.  Thus,  when generating $R_j^i$, the RNN unit can incorporate the patterns in $R_1^i,...,R_{j-1}^i$ (e.g., all page views before the $j$-th day). 
Note that RNNs can learn correlations across the dimensions of a time series, and produce  multi-dimensional outputs.

However, we empirically find that RNN generators still struggle to capture temporal correlations when the length exceeds a few hundred epochs.  
The reason is that for long time series, RNNs take too many passes to generate the entire sample; the more passes taken, the more temporal correlation RNNs tend to forget. 
Prior work copes with this problem in three ways.
The first is to generate only short sequences \cite{mogren2016c,yu2017seqgan,timegan};
long  datasets are evaluated on chunks of tens of samples \cite{timegan,timegan-code}.
The second approach is to train on small datasets, 
 where rudimentary designs may  be able to effectively memorize  long term effects (e.g. unpublished work \cite{blog} generates time series of length 1,000, from a dataset of about $100$ time series).
 This approach leads to memorization \cite{arora2017gans}, which defeats the purpose of training a model. 
 A third approach assumes an auxiliary raw data time series as an additional input during the generation phase to help generate long time series \cite{zec2019recurrent}. 
 This again defeats the purpose of synthetic data generation.

\begin{figure}[t]
    \centering
    \includegraphics[width=0.7\linewidth]{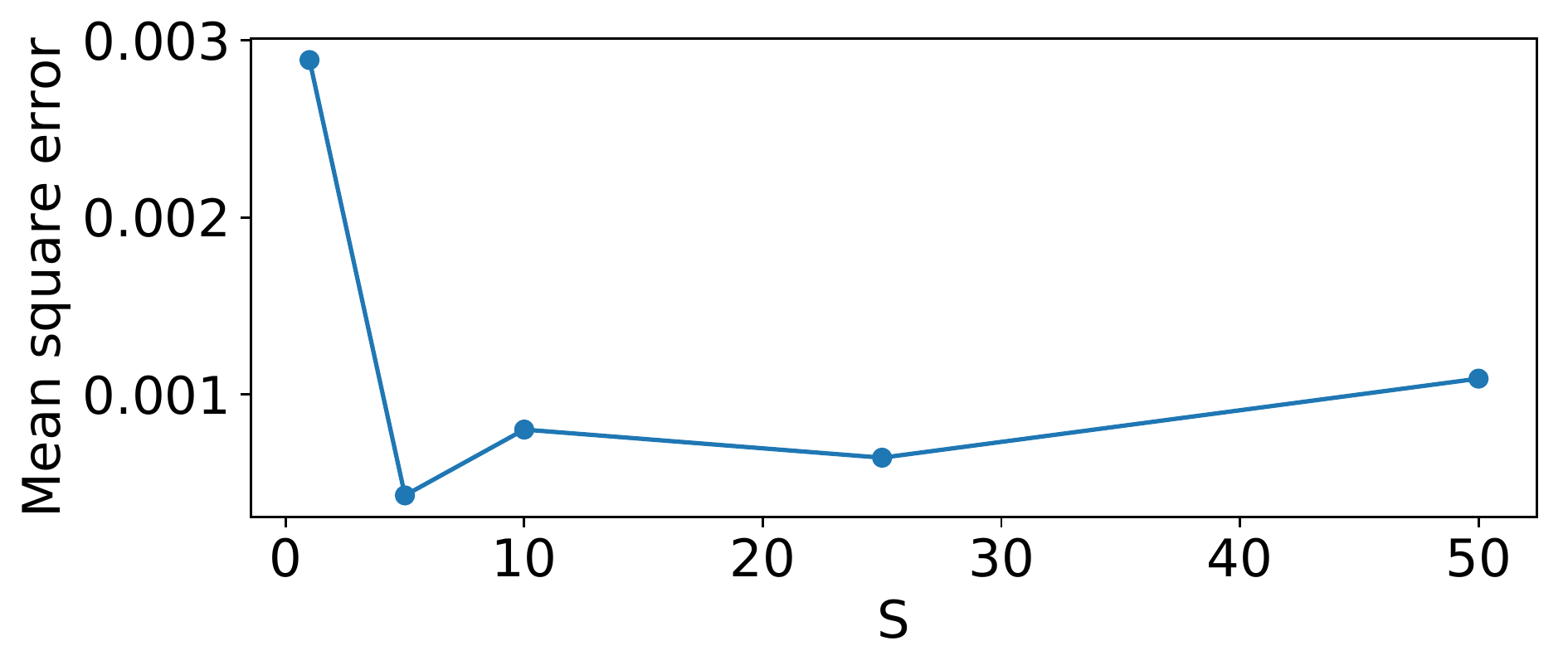}
    \vspace{-0.4cm}
    \caption{Error vs.\ batch parameter.}
    \vspace{-0.5cm}
    \label{fig:s_param}
\end{figure}

\myparatight{Our approach} To reduce the number of RNN passes, we propose to use a simple yet effective idea called  \emph{batch generation}.
At each pass of the RNN, instead of generating one record (e.g., page views of one day), it generates $S$ records (e.g., page views of  $S$ consecutive days), where $S$ is a tunable parameter (Figure \ref{fig:batch_gen} (b)).\footnote{
Our batch generation differs from  two similarly-named concepts. 
\emph{Minibatching} is a standard practice of computing gradients on small sets of samples rather than the full dataset for efficiency \cite{hinton2012neural}.
\emph{Generating batches of sequences} in SeqGAN \cite{yu2017seqgan} involves generating multiple time series during GAN training 
to estimate the reward of a generator policy in their reinforcement learning framework.
Both are orthogonal to our batch  generation.
}
This effectively reduces the total number of RNN passes by a factor of $S$. 
As $S$ gets larger, the difficulty of synthesizing a batch of records at a single RNN pass also increases. 
This induces a natural  trade-off between the number of RNN passes and the single pass difficulty.
For example, Figure \ref{fig:s_param} shows the mean square error between the autocorrelation of our generated signals and real data on the \wikishort{} dataset.
Even  a small (but larger than 1) $S$ gives substantial improvements in signal quality. In practice, we find that $S=5$ works well for many datasets and  a simple autotuning of this hyperparameter similar to this experiment can be used in practice (\Section\ref{sec:design:combine}).

 The above workflow of using RNNs with batch generation  ignores the timestamps from generation.  
In practice, for some datasets, the timestamps may be important in addition to  timeseries samples; e.g.,  if  derived features such as inter-arrival times of requests may be important for downstream systems and networking tasks.  To this end, we support two simple alternatives. First, if the raw timestamps are not important, we can  assume that they are equally spaced and are the same for all \samples{} (e.g., when the dataset is daily page views of websites).
   Second, if the derived temporal properties are critical, we can simply use the  initial timestamp of each \sample{} as an additional \metadata{} (i.e., start time of the \sample{}) and the inter-arrival times between consecutive records as an additional \measurement{}.

\subsection{Tackling mode collapse}
\label{sec:design-mode-collapse}

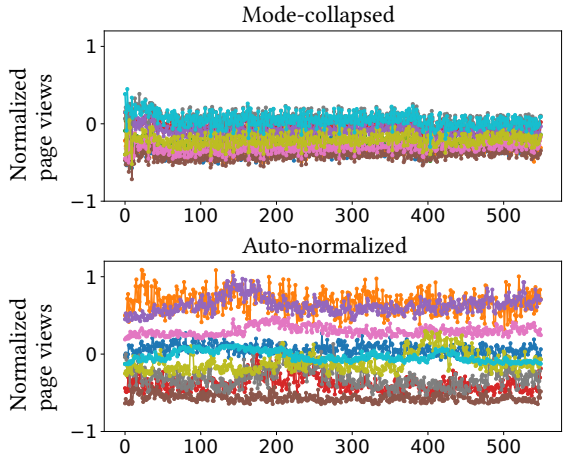
\begin{figure}[t]
	
	\centering
	\begin{minipage}{.8\linewidth}
		\centering
		\includegraphics[width=\linewidth]{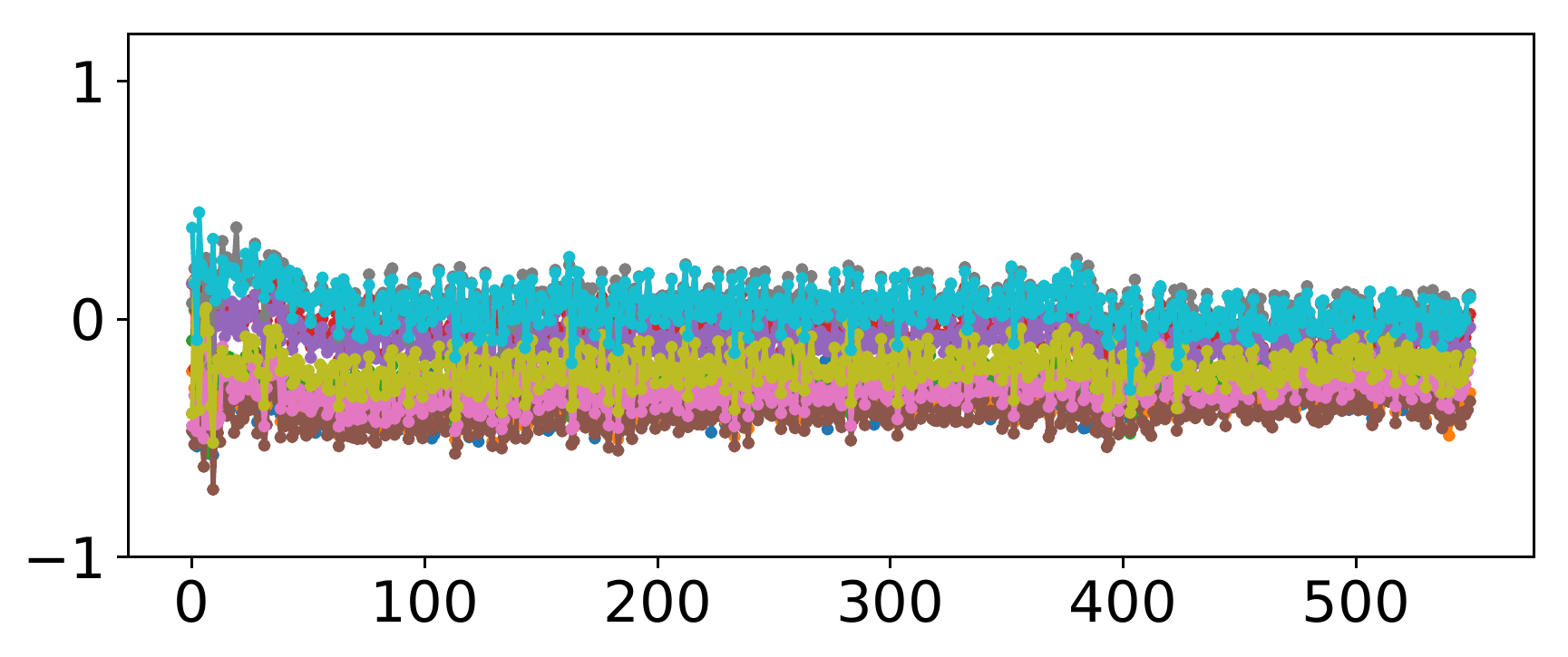}
		\put(-125,80){Mode-collapsed}
		\put(-213,20){\rotatebox{90}{Normalized}}
		\put(-203,20){\rotatebox{90}{page views}}
	\end{minipage}%
	
	\centering
	\begin{minipage}{.8\linewidth}
		\centering
		\includegraphics[width=\linewidth]{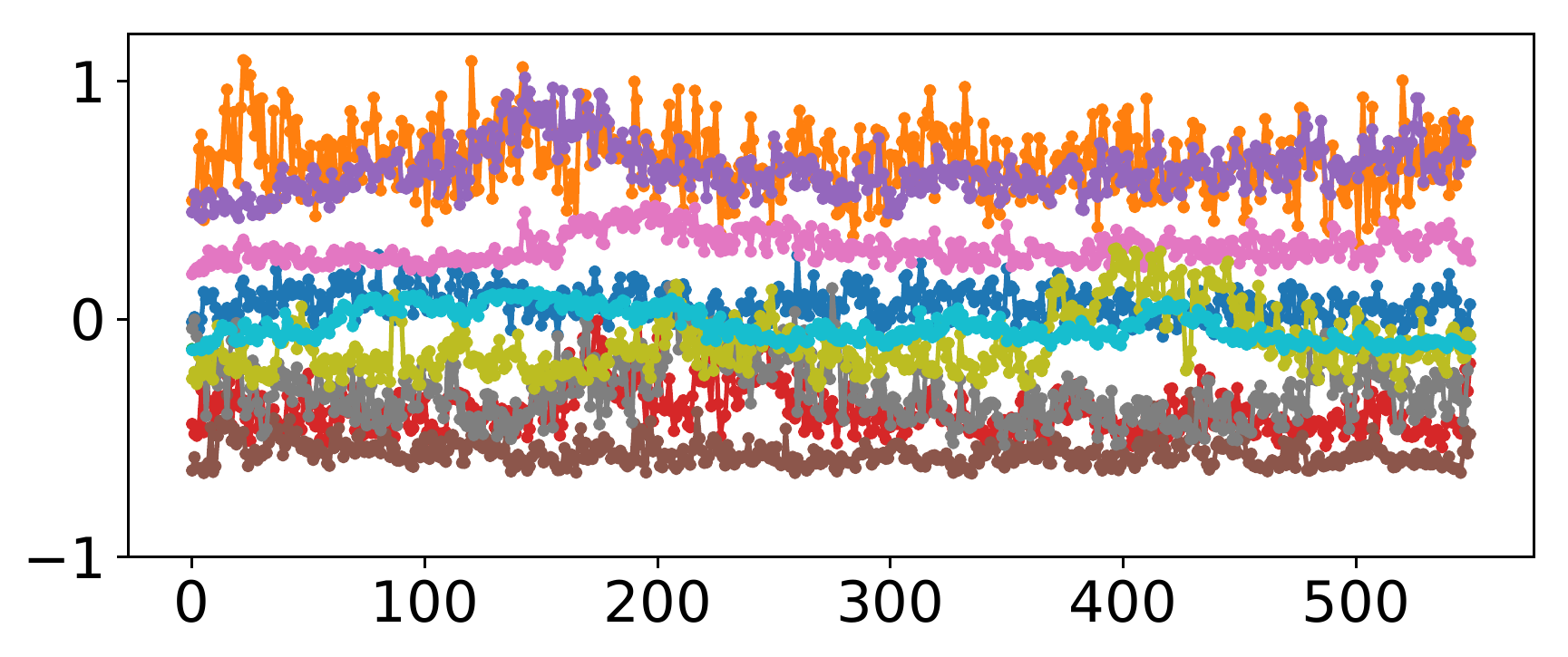}
		\put(-125,80){Auto-normalized}
		\put(-213,20){\rotatebox{90}{Normalized}}
		\put(-203,20){\rotatebox{90}{page views}}
	\end{minipage}%

	\caption{Without auto-normalization, generated  samples show telltale signs of mode collapse as they have similar shapes and amplitudes.}
	\label{fig:modecollapse}
	\vspace{-0.5cm}
\end{figure}

Mode collapse is a well-known  problem  \cite{goodfellow2016nips}, where the GAN outputs homogeneous samples despite being trained on a diverse dataset.
For example, suppose we train on web traffic data that includes three distinct kinds of signals, corresponding to different classes of users. 
A mode-collapsed GAN might learn to generate only one of those traffic types.

For instance, \cref{fig:modecollapse} (top) plots synthetic time series from a GAN trained on  the \wikishort{} dataset, normalized and shifted to $[-1,1]$. 
The generated samples are heavily mode-collapsed, 
exhibiting similar amplitudes, offsets, and shapes.\footnote{While  mode collapse can  happen both in \measurements{} or in \metadata{},  we observed substantially more mode collapse in the \measurements{}.}

\myparatight{Existing work and limitations}
Alleviating mode collapse  is an active research topic in the GAN community (e.g.,  \cite{srivastava2017veegan,lin2018pacgan,arjovsky2017wasserstein,gulrajani2017improved}).  
We experimented with a number of state-of-the-art techniques for mitigating mode collapse \cite{lin2018pacgan,gulrajani2017improved}. 
However, these did not resolve the problem on our datasets.

\myparatight{Our approach}
   Our intuition is that  
  unlike   images or medical data, where value ranges tend  to be similar across samples, networking datasets  exhibit much higher range variability. 
  Datasets 
    with a large range (across samples) appear to worsen  mode collapse because they have a more diverse set of modes,  making them harder to learn.  
    For example, in the \wikishort{} dataset, 
some web pages consistently  have $>$2000 page views per day, whereas  others always have  $<$10.

Rather than using a general solution for mode collapse, we build on this insight to develop a custom  \emph{auto-normalization} heuristic.
  Recall that each time series of \measurements{} $f^i$ (e.g., network traffic volume measurement of a client) is also assigned some \metadatas{} $A^i$ (e.g., the connection technology of the client, cable/fiber). 
 Suppose our dataset has two time series with different offsets: $f^1(t) = \sin(t)$  and $f^2(t)=\sin(t)+100$ and no \metadata{}, so  $A^1=A^2=()$.
 We have $\min(f^1) = -1$, $\max(f^1)=2$, $\min(f^2)=99$, $\max(f^2)=101$. 
 A standard normalization approach (e.g. as in \cite{timegan-code}) would be to simply normalize this data by the \emph{global} min and max, store them as global constants, and train on the normalized data.
 However, this is just scaling and shifting by a constant; from the GAN's perspective, the learning problem is the same, so mode collapse still occurs.

 Instead, we normalize each time series signal  \emph{individually}, and store the min/max  as ``fake"  \metadata{}. 
 Rather than training on the original $(f^i, A^i)$ pairs, we  train on $\tilde f^1(t) = \sin(t)$, $\tilde A^1=(-1,1)$,  $\tilde f^2(t)=\sin(t)$, $\tilde A^2=(99,101)$.\footnote{In reality, we store $\tilde A^i=(\max\{f^i\}  \pm \min\{f^i\})/2$ to ensure that  our generated min is  always less than our max.} 
 Hence, the  GAN learns to generate these two fake \metadatas{} defining the max/min  for each time series individually, which are then used to rescale \measurements{}  to a realistic range. 
 
     Note that this approach differs from typical feature normalization in two ways: (1) it normalizes each sample individually, rather than normalizing over the entire dataset, and (2) it treats the maximum and minimum value of each time series as a random variable to be learned (and generated).
 In this way, all time series have the same range during generation, which alleviates the mode collapse problem. 
Figure \ref{fig:modecollapse} (bottom) shows that by training \name{} with auto-normalization on the  \wikishort{} data, we generate samples with a broad range of amplitudes, offsets, and shapes.

\subsection{Capturing attribute relationships}
\label{sec:design-correlation}
 So far, we have only discussed how to generate time series.
 However, \metadatas{} can strongly influence the characteristics of \measurements{}. 
 For example, fiber users tend to use more traffic than cable users. 
 Hence, we need a mechanism  to  model the \emph{joint} distribution  between \measurements{} and \metadatas{}. 
 As discussed in \S\ref{sec:challenges}, naively generating concatenated  \metadata{} $A^i$ and \measurements{} $R^i$ does not learn the correlations between them well. 
We hypothesize that this is because jointly generating \metadatas{} and \measurements{} using a single generator is too difficult.

\myparatight{Existing work and limitations} 
A few papers have tackled this problem, mostly in the context of generating multidimensional data.
The dominant approach  in the literature is to train a variant of GANs called conditional GANs (CGANs), which learn to produce data conditioned on  a user-provided input label.  
For example, prior works \cite{esteban2017real,fu2019time,zec2019recurrent} learn a conditional model in which the user inputs the desired \metadata{}, and the architecture generates  \measurements{} conditioned on the attributes; generating the attributes as well is a simple extension \cite{esteban2017real}.
TimeGAN claims to co-generate \metadata{} and \measurements{}, 
but it does not evaluate on any datasets that include \metadata{} in the paper, nor does the released code handle \metadata{} \cite{timegan,timegan-code}.

\myparatight{Our approach} 
We start by decoupling this problem  into two  sub-tasks: generating \metadatas{} and generating \measurements{} {\em conditioned}  on \metadatas{}: 
$
P(A^i, R^i) = P(A^i) \cdot P(R^i|A^i)
$, each using a dedicated generator;
this is is conceptually similar to prior approaches \cite{esteban2017real,zec2019recurrent}. 
More specifically, we use a  standard multi-layer perceptron (MLP) network for  generating the \metadatas{}. 
This is a good choice,  as MLPs are good at modeling (even high-dimensional) non-time-series data. 
For \measurement{} generation, we use the RNN-based architecture from \S\ref{sec:design-long}. 
To  preserve the hidden relationships between the \metadata{} and the \measurements{},  the generated \metadata{} $A^i$ is added as an input to the RNN at every step. 

Recall from  \cref{sec:design-mode-collapse} that we treat the max and min of each time series as \metadata{} to  alleviate mode collapse.
Using this conditional framework, we divide the generation of max/min \metadata{} into three steps: (1) generate ``real" \metadatas{} using the MLP generator (\S\ref{sec:design-correlation}); 
(2) with the generated \metadatas{} as inputs, generate the two ``fake" (max/min) \metadatas{} using another MLP; (3) with the generated real and fake \metadatas{} as inputs, generate \measurements{} using the architecture in \S\ref{sec:design-long} (see Figure \ref{fig:architecture}).

Unfortunately, a decoupled architecture alone  does  not  solve the problem.
Empirically, we  find that when the average length of \measurements{} is long (e.g., in the \wikishort{} dataset, each sample consists of 550 consecutive daily page views), the fidelity of generated data---especially the \metadata{}---is poor. To understand why, recall that a GAN  discriminator  judges the fidelity of generated samples and provides feedback for the generator to improve. 
When the total dimension of samples (\measurements{} + \metadatas{}) is large, judging sample fidelity is hard. 

Motivated by this, we introduce an \emph{auxiliary discriminator} which discriminates only on \metadatas{}. 
The losses from two discriminators are combined by a weighting parameter $\alpha$:
$
	\min_G \max_{D_1,D_2} \mathcal{L}_1(G,D_1) + \alpha \mathcal{L}_2(G,D_2)
$
where $\mathcal{L}_i$, $i\in\{1,2\}$ is the Wasserstein loss of the original and the auxiliary discriminator respectively.
The generator effectively learns from this auxiliary discriminator to generate high-fidelity \metadata{}. 
Further, with the help of the original discriminator, the generator can learn to generate \measurements{} well. 

\begin{figure}[htb]
	\centering
	\begin{minipage}{0.65\linewidth}
		\centering
		\includegraphics[width=\linewidth]{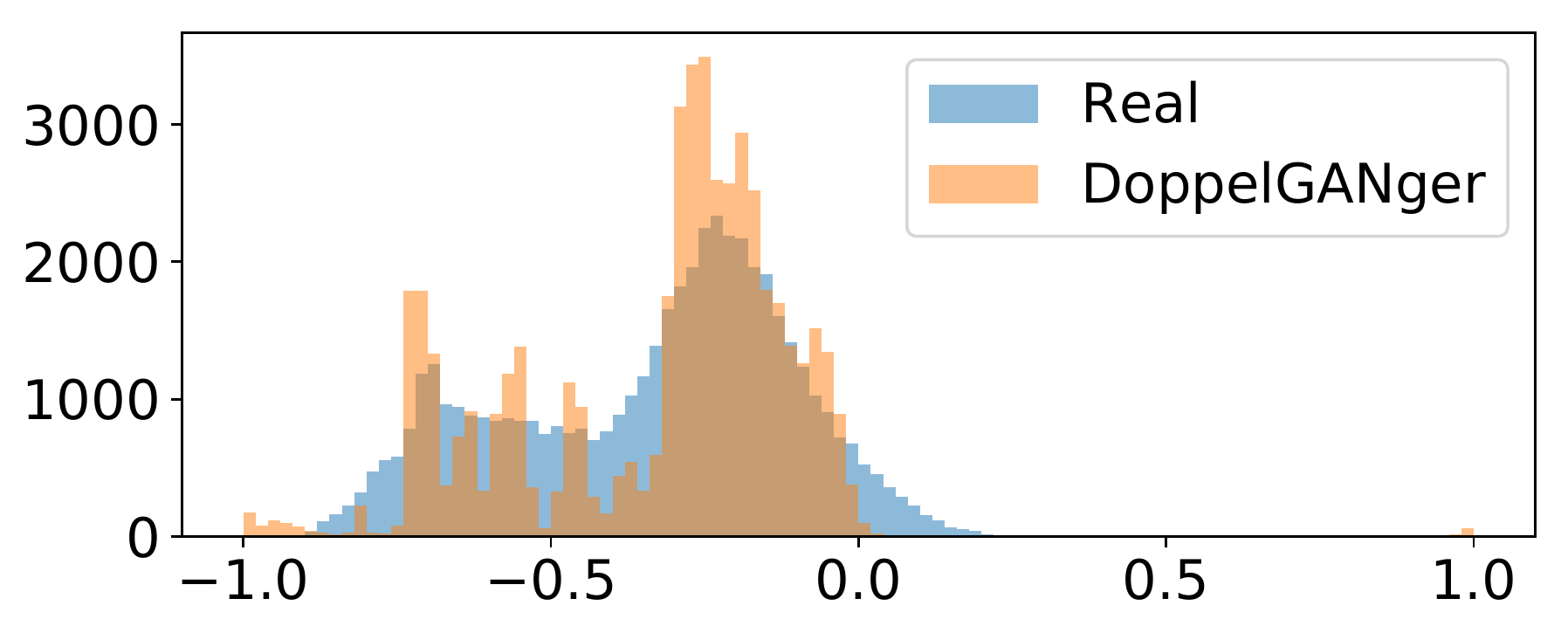}
		\includegraphics[width=\linewidth]{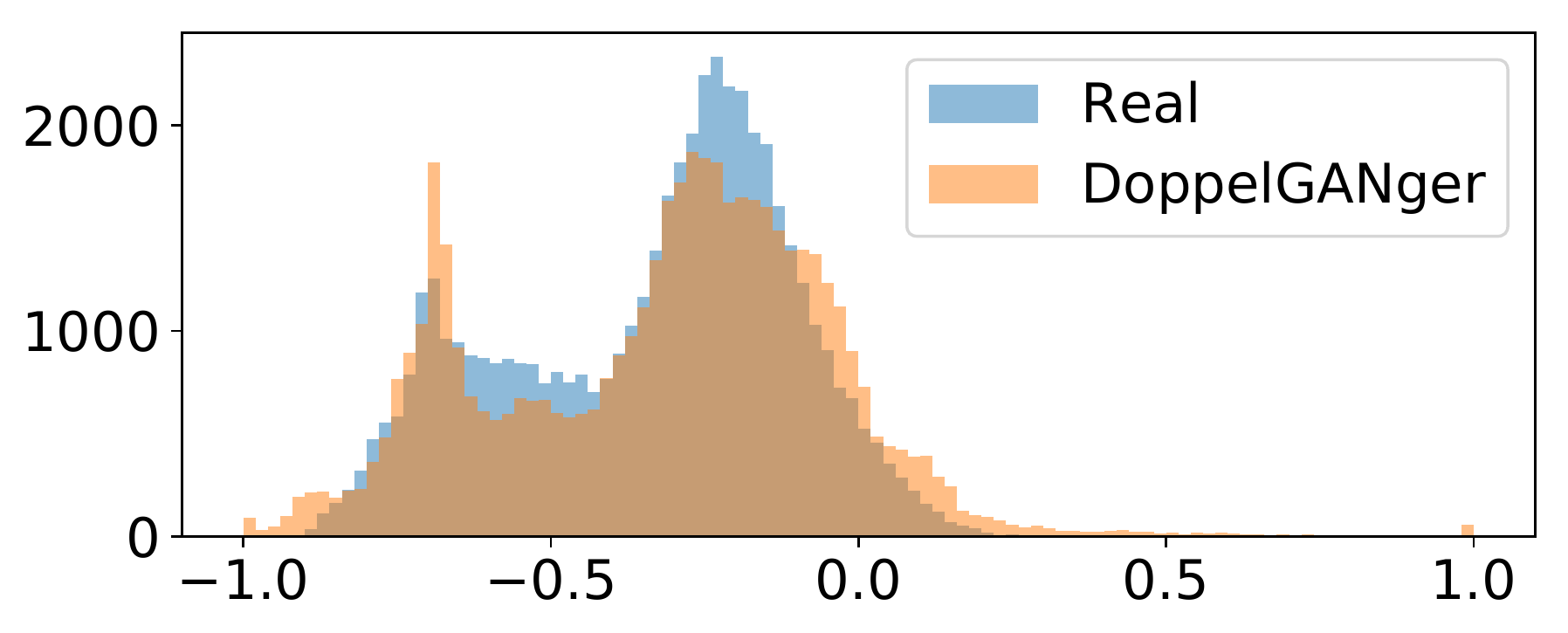}
		\includegraphics[width=\linewidth]{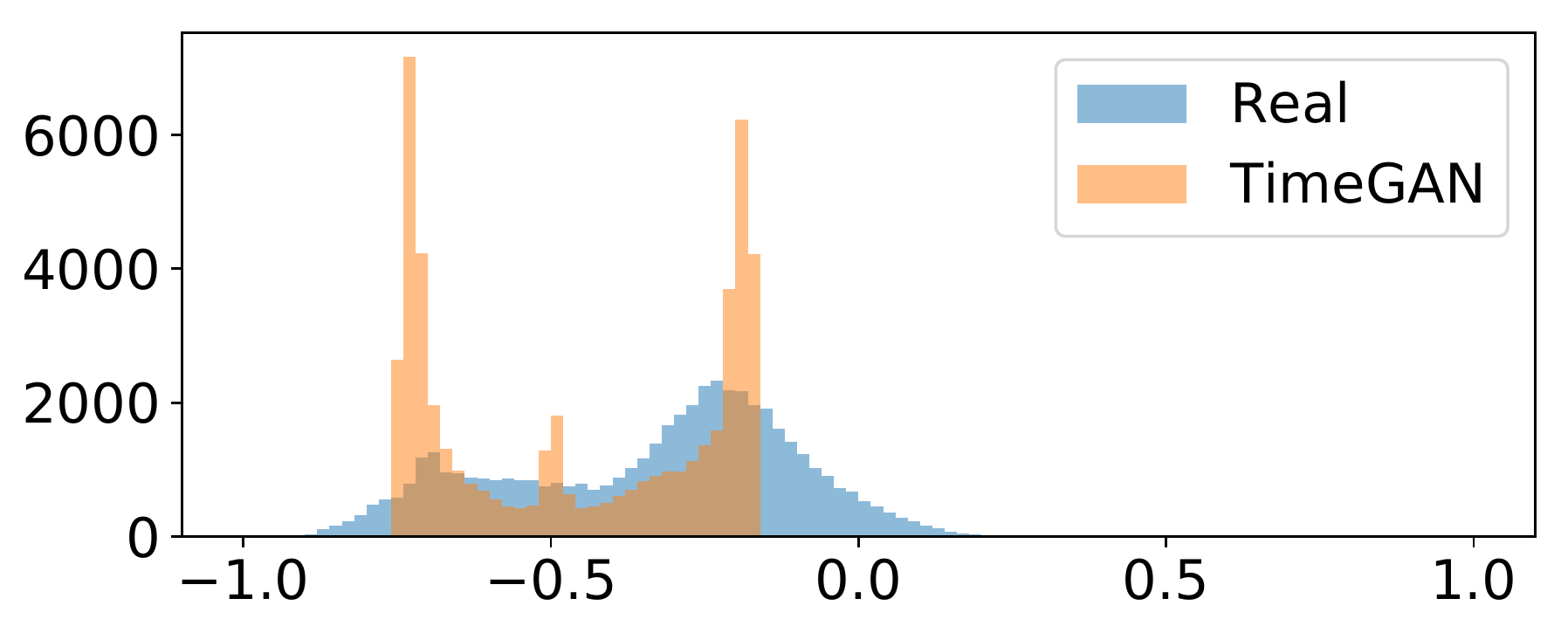}
		\includegraphics[width=\linewidth]{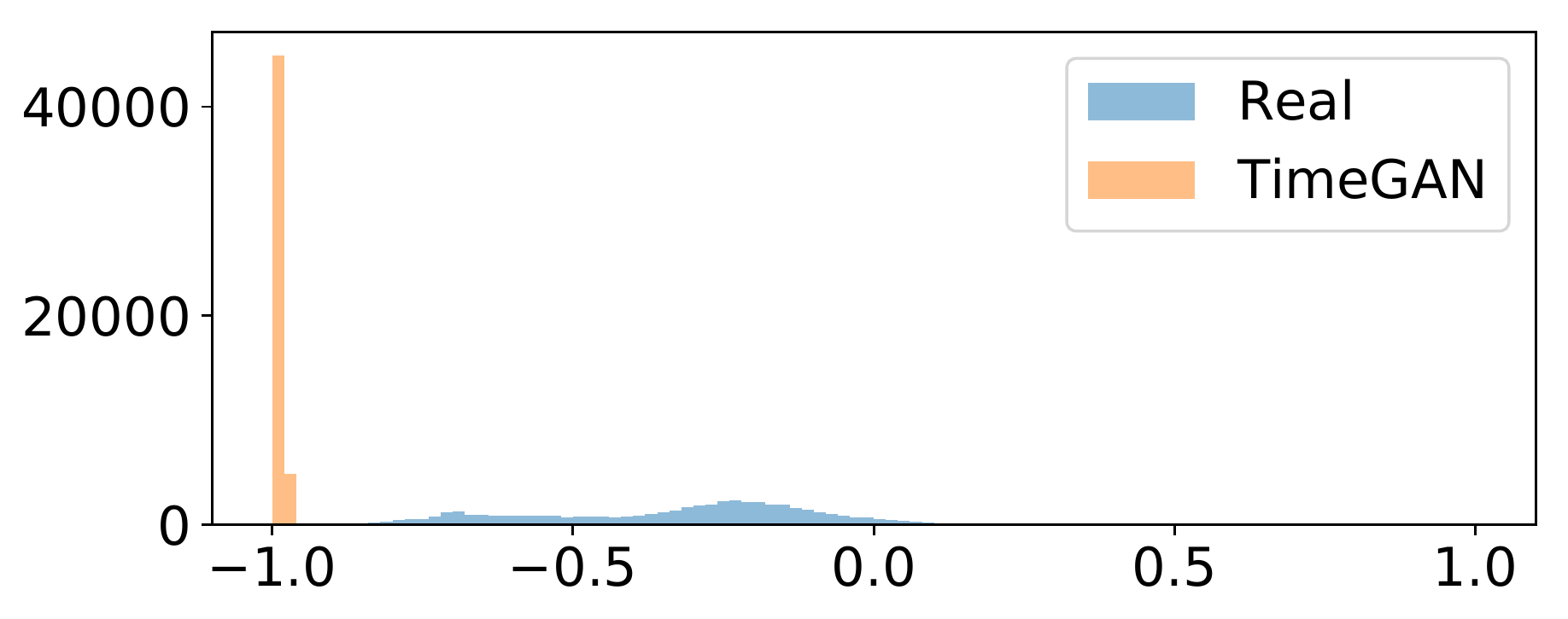}
	\end{minipage}
	\caption{ Distribution of (max+min)/2  of (a) \nameshort{} without and (b) \nameshort{} with the auxiliary discriminator, (c) TimeGAN, and (d) RCGAN (\wikishort~data).}
	\label{fig:(max+min)/2}
\end{figure}
Empirically, we find that this architecture improves the data fidelity significantly.
Figure \ref{fig:(max+min)/2} shows a histogram of the (max$+$min)/2 \metadata{} distribution from \name{} on the \wikishort{} dataset.
That is, for each time series, we extract the maximum and minimum value, and compute their average; then we compute a histogram of these averages over many time series. 
This distribution implicitly reflects how well \name{} reproduces the range of time series values in the dataset.
We observe that adding the auxiliary discriminator significantly improves the fidelity of the generated distribution, particularly in the tails  of the true distribution.

\subsection{Putting it all together}
\label{sec:design:combine}
\begin{figure}[th]
	\centering
	\includegraphics[width=0.95\linewidth]{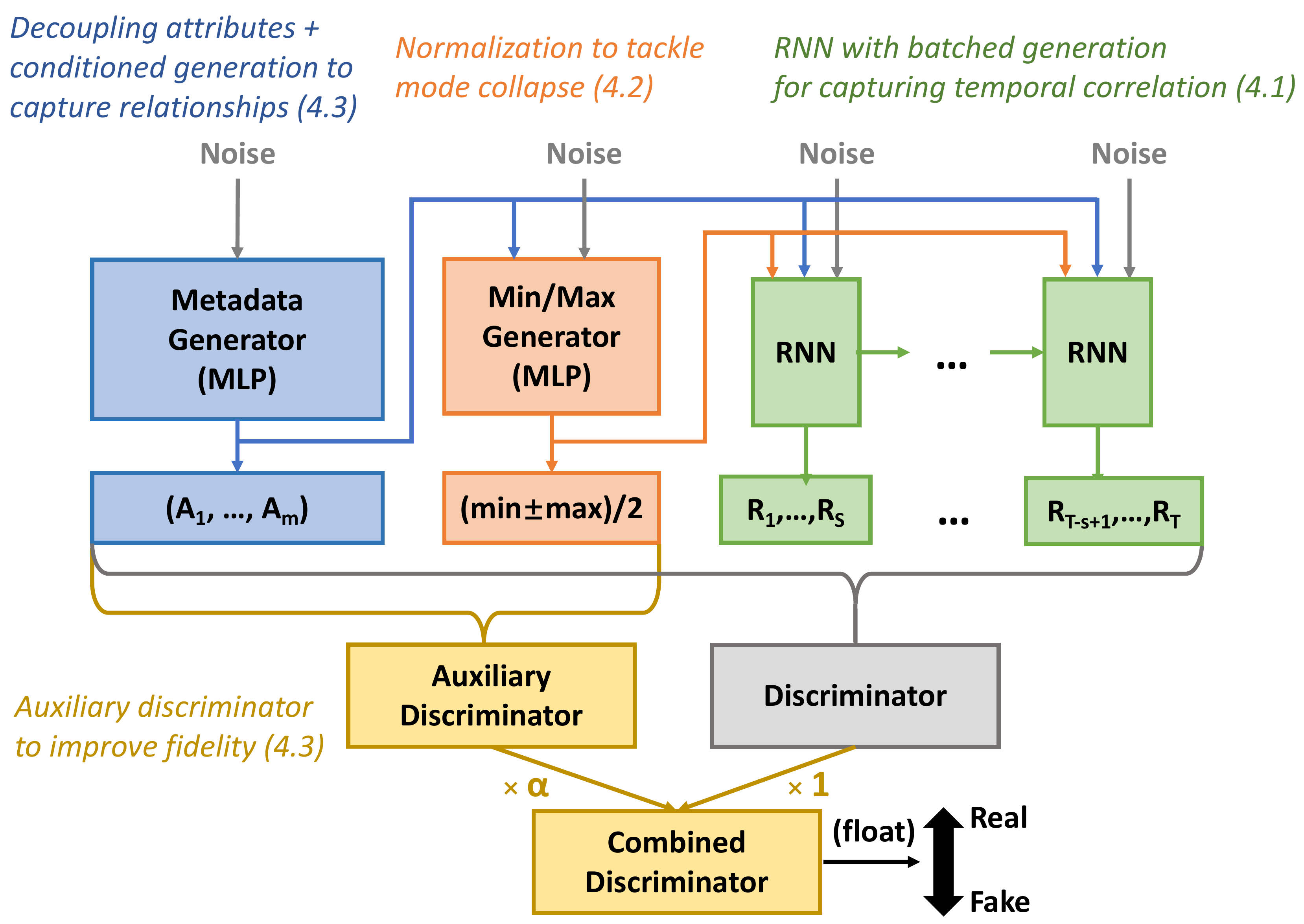}
	\caption{Architecture of \name{} highlighting key ideas and extensions to canonical GAN approaches.}
	\vspace{-0.3cm}
	\label{fig:architecture}
\end{figure}

The overall \name{} architecture is in Figure \ref{fig:architecture},   highlighting the key differences from canonical approaches. First, to capture the correlations between \metadata{} and \measurements{}, we use a decoupled generation of \metadata{} and \measurements{} using an auxiliary discriminator, and conditioning the  \measurements{} based on the \metadata{} generated.  Second, to address the mode collapse problem for the \measurements{}, we add the fake metadata capturing the min/max values for each generated sample. Third, we use a batched RNN generator to capture the temporal correlations and synthesize long time series that are representative.

The training phase requires two primary inputs: the  \emph{data schema} (i.e.,   \metadata{}/\measurement{} dimensions, indicating whether they are categorical or numeric) and the {\em training data}.    
 The only minimal tuning that  data holders sharing a dataset using \name{} need to be involved in is selecting  the \measurement{} batch size $S$ (\S\ref{sec:design-long})  controls the number of \measurements{} generated at each RNN pass. Empirically, setting $S$ so that $T/S$ (the number of steps RNN needs to take) is around 50 gives good results, whereas prior time series GANs use $S=1$ \cite{esteban2017real,beaulieu2017privacy,timegan,zec2019recurrent}.  Optionally,    data holders can specify  sensitive \metadatas{}, whose distribution  can be masked or request additional   privacy settings  to be used (\Section\ref{sec:exp-privacy}).   We envision data holders sharing the {\em generative model} with the data users.  Users can then flexibly use this model and also optionally   specify different \metadata{} distribution (e.g., for amplifying rare  events) if needed. That said, our workflow  also accommodates a more restrictive mode of sharing, where the  holder uses \name{} to generate synthetic data internally and then releases the generated data without   sharing the model.\footnote{From a privacy   perspective, model and  data  sharing may suffer similar information leakage risks~\cite{hayes2019logan}, but this may be a pragmatic choice some providers can make nonetheless.}
 
 The code and a detailed documentation (on data format, hyper-parameter setting, model training, data generation, etc.) are available at \url{https://github.com/fjxmlzn/DoppelGANger}.

\section{Fidelity Evaluation}
\label{sec:evaluation}

We evaluate the fidelity of \name{}  on three datasets, whose properties are summarized in Table \ref{tbl:data}\footnote{Our choice of using public datasets  is  to enable others to independently validate and reproduce our work.}. 
\subsection{Setup}
\subsubsection{Datasets}

These datasets are chosen to exhibit different combinations of  challenges: %
 (1) correlations within time series and \metadata{}, (2) multi-dimensional \measurements{}, and/or (3) variable \measurement{} lengths.

\begin{table}[t]
	\centering
	\begin{tabular}{|m{1.55cm}|m{1.5cm}|m{2.2cm}|m{1.3cm}|}
		\hline
		Dataset & Correlated in time \& metadata & Multi-dimensional \measurements{} & Variable-length signals\\
		\hline
		\wikishort{} \cite{goog-web-traffic} 
		& x & & \\ \hline
		\fccshort{} \cite{mba-data} 
		& x & x & \\ \hline
		\clustershort{} \cite{reiss2011google}  & x & x & x \\ \hline
	\end{tabular}
	\caption{Challenging properties of studied datasets.}\label{tbl:data}
	\vspace{-0.8cm}
\end{table}

\myparatightest{\wiki{} (\wikishort{})} This dataset tracks the number of daily views of Wikipedia articles, starting from July 1st, 2015 to December 31st, 2016 \cite{goog-web-traffic}.
In our language, each \emph{sample} is a  page view counter for one Wikipedia page, with three \emph{\metadatas{}}: Wikipedia domain, type of access (e.g., mobile, desktop), and type of agent (e.g., spider). 
Each \sample~has one \measurement{}:
the number of daily page views.

\myparatightest{\fcc{} (\fccshort{})}
This dataset was collected by United States Federal Communications Commission (FCC) \cite{mba-data} and consists of several measurements such as round-trip times and packet loss rates from several clients in geographically diverse homes to different servers using different protocols (e.g. DNS, HTTP, PING).
Each \emph{sample} consists of measurements from one device.
Each sample has three \emph{\metadatas{}}: Internet connection technology, ISP, and US state. A record contains UDP ping loss rate (min.\  across measured servers) and total traffic (bytes sent and received), reflecting client's aggregate Internet usage.

\myparatightest{\cluster{} (\clustershort{})}
This dataset \cite{reiss2011google} contains usage traces of a Google Cluster of 12.5k machines over 29 days in May 2011.
We use the logs containing measurements of task resource usage, and the exit code of each task. 
Once the task starts,  the system measures its resource usage (e.g. CPU usage, memory usage) per second, and logs  aggregated statistics every 5 minutes (e.g., mean, maximum). 
Those resource usage values are the \emph{\measurements{}}. 
When the task ends, its end event type (e.g. FAIL, FINISH, KILL) is also logged. 
Each task has  one end event type, which we treat as an \emph{\metadata{}}. 

More details of the datasets (e.g. dataset schema) are attached in Appendix \ref{app:dataset}.

\subsubsection{Baselines}
\label{sec:eva-baseline}
We only compare \name{} to the baselines in \S\ref{sec:strawman} that are general---the machine-learned models. (In the interest of reproducibility, we provide complete configuration details for these models in Appendix~\ref{app:implementation}.)

\myparatightest{Hidden Markov models (HMM) (\S \ref{sec:strawman})} While HMMs have been used for generating time series data, there is no  natural way to jointly generate \metadatas{} and time series in HMMs. 
Hence, we infer a separate multinomial distribution for the \metadatas{}. 
During generation, \metadatas{} are randomly drawn from the multinomial distribution on training data, independently of the time series. 

\myparatightest{Nonlinear auto-regressive (AR) (\S \ref{sec:strawman})}
Traditional AR models can only learn to generate \measurements{}. In order to jointly learn to generate \metadatas{} and \measurements{}, we design the following more advanced version of AR: we learn a function $f$ such that $R_t = f(A, R_{t-1}, R_{t-2}, ..., R_{t-p})$.
To boost the accuracy of this baseline, we use a multi-layer perceptron version of $f$.
During generation, $A$ is randomly drawn from the multinomial distribution on training data, and the first record $R_1$ is drawn a Gaussian distribution learned from training data.

\myparatightest{Recurrent neural networks (RNN) (\S \ref{sec:strawman})}
In this model, we train an RNN via teacher forcing \cite{williams1989learning} by feeding in the true time series at every time step and predicting the value of the time series at the next time step. Once trained, the RNN can be used to generate the time series by using its predicted output as the input for the next time step.
A traditional RNN can only learn to generate \measurements{}. We design an extended  RNN takes \metadata{} $A$ as an additional input. 
During generation, $A$ is randomly drawn from the multinomial distribution on training data, and the first record $R_1$ is drawn a Gaussian distribution learned from training data.

\myparatightest{Naive GAN (\S \ref{sec:challenges})} We include the naive GAN architecture (MLP generator and discriminator) in all our evaluations. %

\myparatightest{TimeGAN \cite{timegan}} 
 Note that the state-of-the-art TimeGAN \cite{timegan-code} does not jointly generate metadata and high-dimensional time series of different lengths, so several of our evaluations cannot be run on TimeGAN.
However, we modified  the TimeGAN implementation directly \cite{timegan-code} to run on the \wikishort{} dataset (without metadata) and compared against it.

\myparatight{RCGAN \cite{esteban2017real}}
RCGAN does not generate \metadata{}, and only deals with time series of the same length, so again, several of our evaluations cannot be run on RCGAN. To make a comparison, we used the version without conditioning (called RGAN \cite{esteban2017real}) from the official implementation \cite{rcgan-code} and evaluate it on the \wikishort{} dataset (without \metadata{}).

\myparatightest{Market Simulator \cite{market-sim}} We also compare against a VAE-based approach \cite{market-sim} designed to generate synthetic financial market data, since its code is publicly available. %

\subsubsection{Metrics}
Evaluating GAN fidelity is notoriously difficult \cite{lucic2018gans,xu2018empirical};  the most widely-accepted metrics  are designed for labelled image data \cite{salimans2016improved,heusel2017gans} and cannot be applied to our datasets. 
Moreover, numeric metrics do not always capture the qualitative problems of generative models. 
We therefore evaluate \name{} with a combination of qualitative and quantitative microbenchmarks and downstream tasks that are tailored to each of our datasets.
Our \emph{microbenchmarks} evaluate how closely a statistic of the generated data matches the real data. E.g., the statistics could be attribute distributions or autocorrelations, and the similarity can be evaluated qualitatively or by computing an appropriate distance metric (e.g., mean square error, Jensen-Shannon divergence). 
Our \emph{downstream tasks} use the synthetic data to reason about the real data, e.g., attribute prediction or algorithm comparison. 
In line with the recommendations of \cite{lucic2018gans}, these tasks can be evaluated with quantitative, task-specific metrics like prediction accuracy. 
Each metric is explained in more detail inline.

\subsection{Results}
\label{sec:fidelity}

\begin{figure}
    \centering
    \includegraphics[width=.7\linewidth]{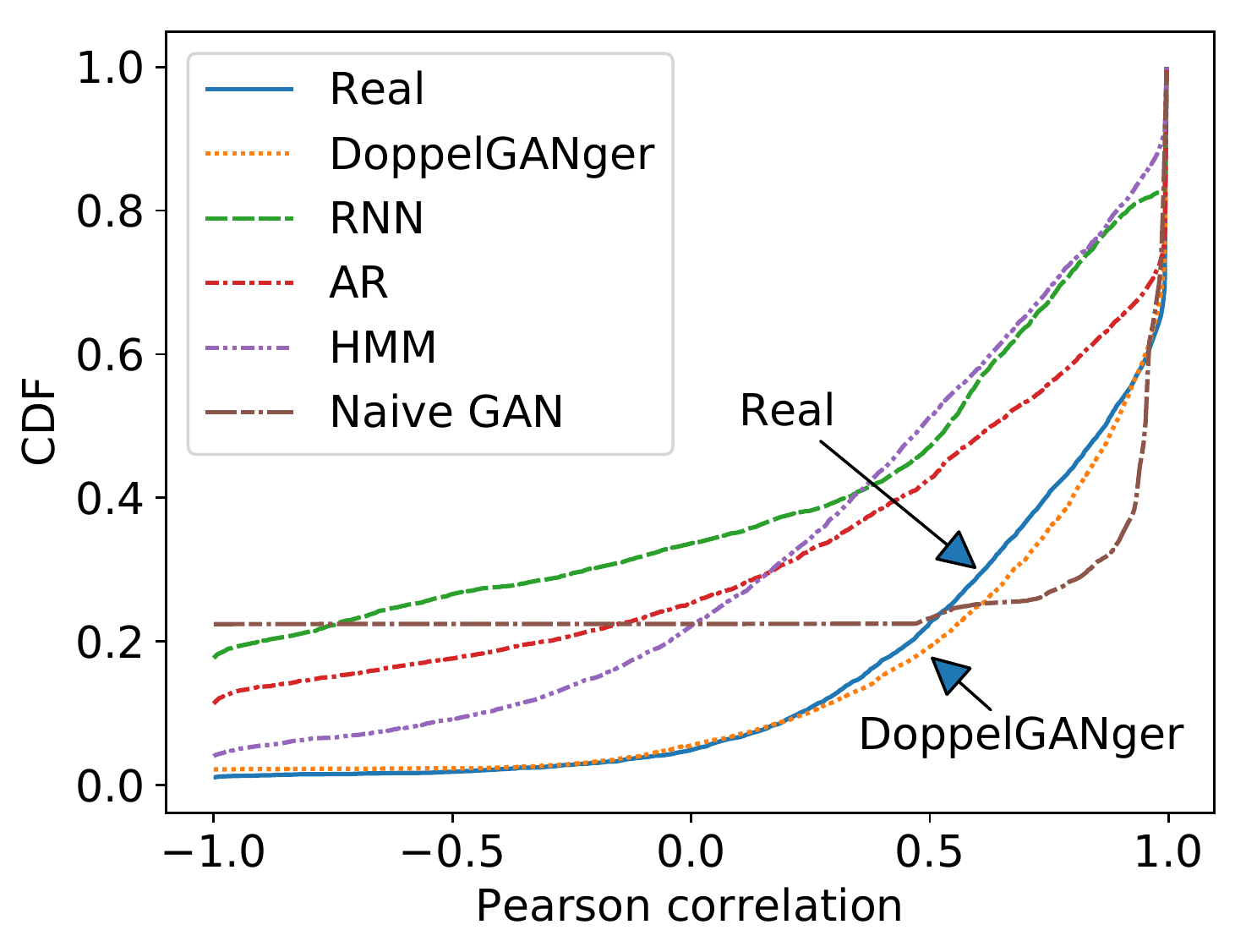}
    \caption{CDF of Pearson correlation between CPU rate and assigned memory usage from \clustershort{}.}
    \vspace{-0.3cm}
    \label{fig:google-cluster-mea-corr}
\end{figure}

\myparatight{Structural characterization}
In line with prior recommendations \cite{melamed1993applications}, we explore how \name{} captures structural data properties like temporal correlations, \metadata{} distributions, and \metadata{}-\measurement{} joint distributions.\footnote{
Such properties are sometimes ignored in the ML literature in favor of downstream performance metrics; however, in systems and networking, we argue such microbenchmarks are  important.} 

\noindent\underline{Temporal correlations:}
To show how \name{} captures temporal correlations,
Figure \ref{fig:wiki-autocorr} shows the average autocorrelation for the \wikishort{} dataset for real and synthetic datasets (discussed in \S \ref{sec:strawman}). 
As mentioned before, the real data has a short-term weekly correlation and a long-term annual correlation. 
\name{} captures both, as evidenced by the periodic weekly spikes and the local peak at roughly the 1-year mark, unlike our baseline approaches.
It also exhibits a 91.2\% lower mean square error from the true data autocorrelation  than the closest baseline (RCGAN).

The fact that \name{} captures these correlations is surprising, particularly  since we are using an  RNN  generator. 
Typically,  RNNs are able to reliably  generate time series of length around 20, while the length of \wikishort{} \measurements{} is 550. 
We believe this is due to a combination of adversarial training (not typically used for RNNs) and our batch generation.
Empirically, eliminating either feature hurts the learned autocorrelation. %
TimeGAN and RCGAN, for instance, use RNNs and adversarial training but does not batch generation, though its performance may be due to other architectural differences \cite{timegan,esteban2017real}.
E.g.,  \wikishort{} is an order of magnitude longer than the time series it evaluates on  \cite{timegan-code,rcgan-code}.

\begin{figure}[t]
	\centering
	\includegraphics[width=0.7\linewidth]{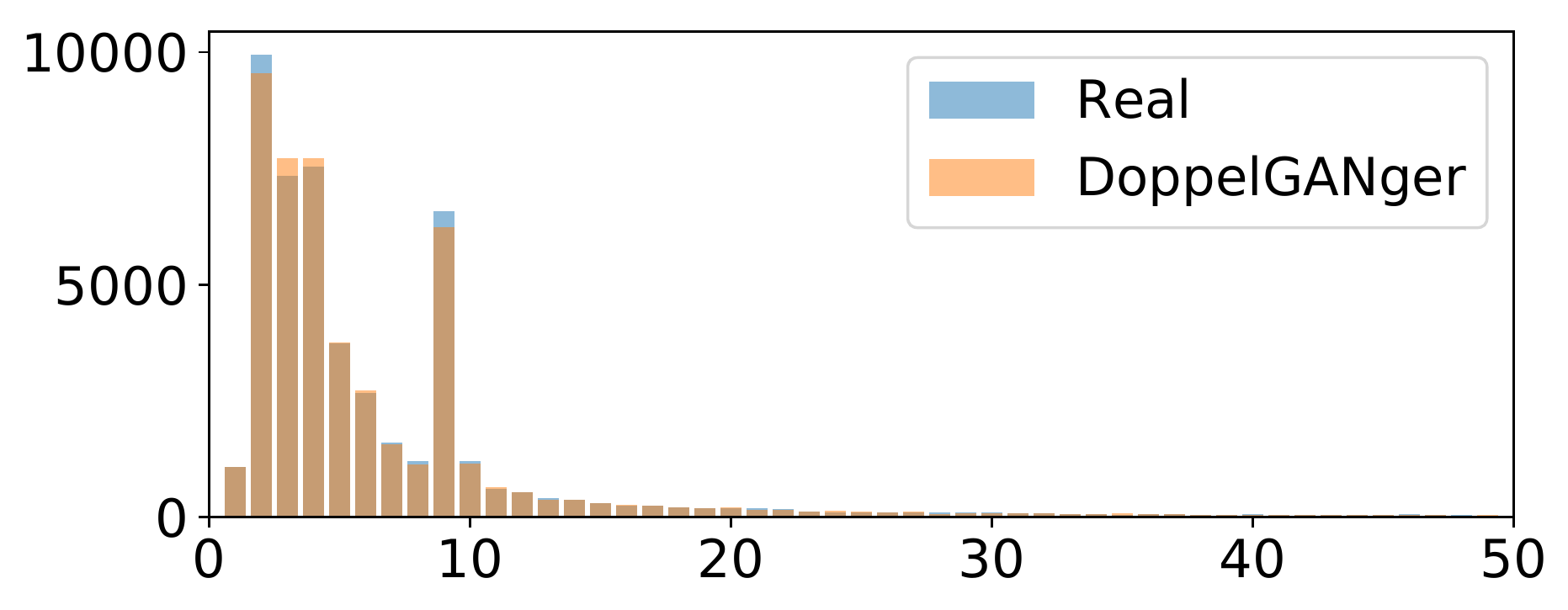}
	\includegraphics[width=0.7\linewidth]{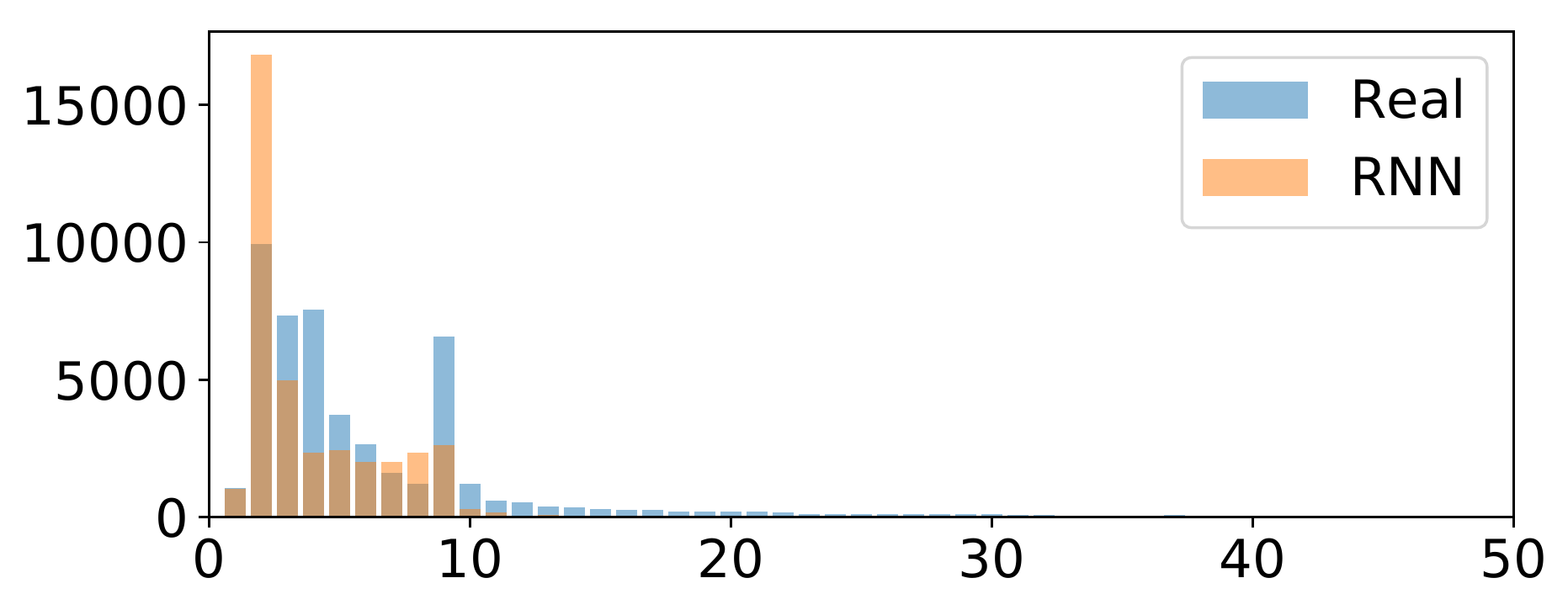}
	\put(-130,-5){Task duration (seconds)}
	\put(-180,45){\rotatebox{90}{Count}}
	\vspace{-0.3cm}
	\caption{Histogram of task duration for the \cluster{}. RNN-generated data misses the second mode, but DoppelGANger captures it.}
	\vspace{-0.6cm}
	\label{fig:google-length-hist}
\end{figure}

Another aspect of learning temporal correlations is generating time series of the right length.
Figure \ref{fig:google-length-hist} shows the duration of tasks in the \clustershort{} dataset for real and synthetic datasets generated by \name{} and RNN. 
Note that TimeGAN generates time series of different lengths by first generating time series of a maximum length and then truncating according to the empirical length distribution from the training data \cite{timegan-code}. 
Hence we do not compare against TimeGAN because the comparison is not meaningful; 
it perfectly reproduces the empirical length distribution, but not because the generator is learning to reproduce time series lengths. 

\name{}'s length distribution fits the real data well, capturing the bimodal pattern in real data, whereas RNN fails. 
Other baselines are even worse at capturing the length distribution (Appendix \ref{app:fidelity}).
We observe this regularly; while \name{} captures multiple data modes, our baselines tend to capture one at best.
This may be due to the naive randomness in the other baselines. 
RNNs and AR models incorporate too little randomness, causing them to learn simplified duration distributions; HMMs instead are too random: they maintain too little state to generate meaningful results. 

\noindent\underline{Cross-\measurement{} correlations:}
To evaluate correlations between the dimensions of our measurements,
we computed the Pearson correlation between the CPU and memory measurements of generated samples from  the \clustershort{} dataset.
Figure  \ref{fig:google-cluster-mea-corr} shows the CDF of these correlation coefficients for different time series.  
We observe that \name{} much more closely mirrors the true auto-correlation coefficient distribution than any of our baselines.

\noindent\underline{\Measurement{} distributions:}
As discussed in \S\ref{sec:design-correlation} and Figure \ref{fig:architecture}, \nameshort{} captures the distribution of (max+min)/2 of page views in \wikishort{} dataset. As a comparison, TimeGAN and RCGAN have much worse fidelity. TimeGAN captures the two modes in the distribution, but fails to capture the tails. RCGAN does not learn the distribution at all. In fact, we find that RCGAN has severe mode collapse in this dataset: all the generated values are close to -1. Some possible reasons might be: (1) The maximum sequence length experimented in RCGAN is 30 \cite{rcgan-code}, whereas the sequence length in \wikishort{} is 550, which is much more difficult; (2) RCGAN used different numbers of generator and discriminator updates per step in different datasets \cite{rcgan-code}. We directly take the hyper-parameters from the longest sequence length experiment in RCGAN's code \cite{rcgan-code}, but other fine-tuned hyper-parameters might give better results. Note that unlike RCGAN, \nameshort{} is able to achieve good results in our experiments without tuning the numbers of generator and discriminator updates.

\begin{figure}[t]
	\centering
	\includegraphics[width=0.6\linewidth]{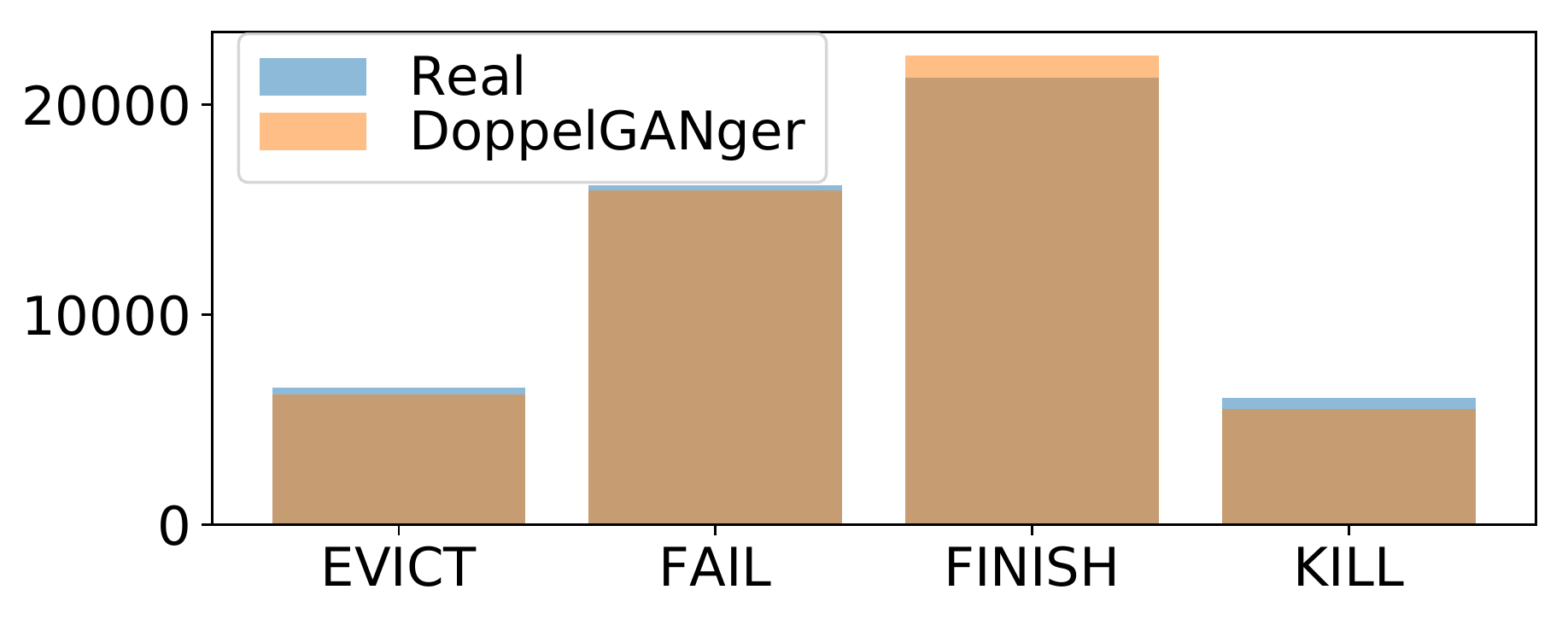}
	\includegraphics[width=0.6\linewidth]{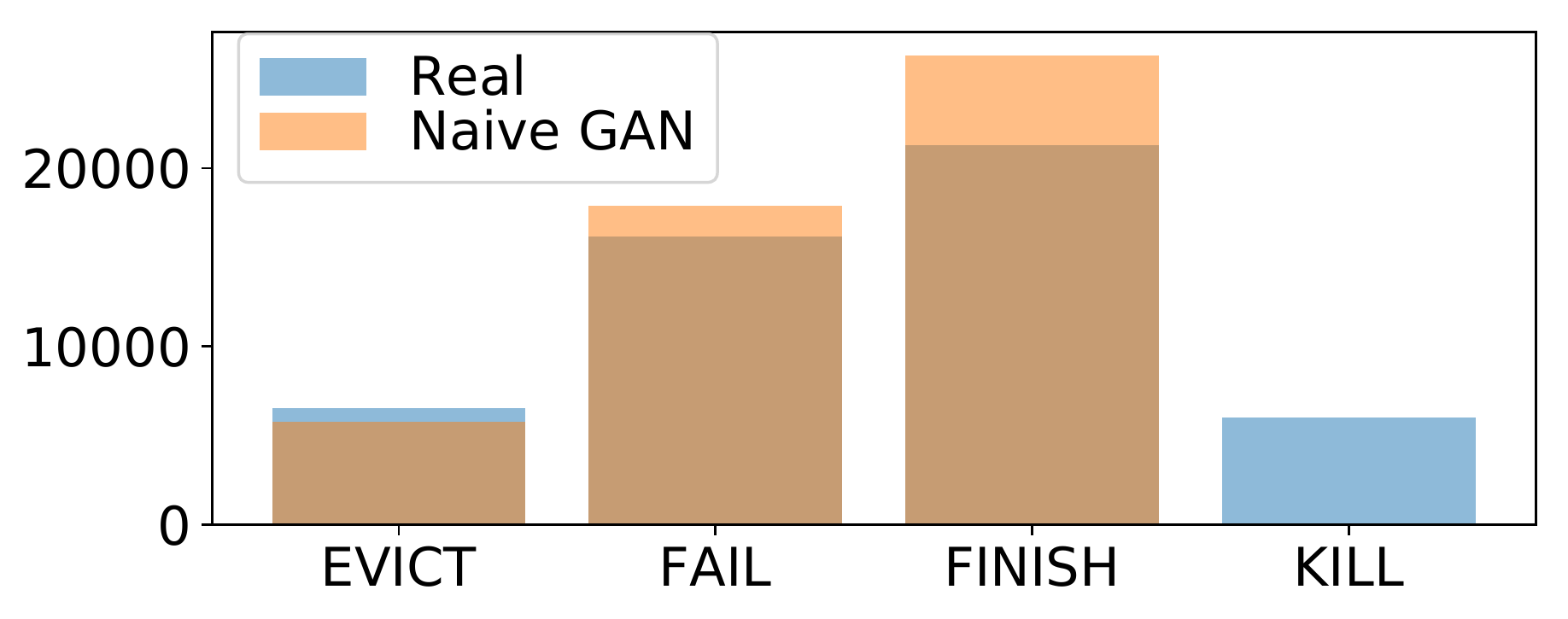}
	\put(-100,-5){End Event Type}
	\put(-160,40){\rotatebox{90}{Count}}
	\vspace{-0.3cm}
	\caption{Histograms of end event types from \clustershort{}.}
	\label{fig:google-exit-hist}
\end{figure}

\noindent\underline{\Metadata{} distributions:}
Learning correct \metadatas{} distributions is necessary for learning \measurement{}-\metadata{} correlations. 
As mentioned in \S \ref{sec:eva-baseline}, for our HMM, AR, and RNN baselines, \metadatas{} are randomly drawn from the multinomial distribution on training data because there is no clear way to jointly generate \metadatas{} and \measurements{}. 
Hence, they trivially learn a perfect \metadata{} distribution. 
Figure \ref{fig:google-exit-hist} shows that \name{} is also able to mimic the real distribution of end event type distribution in \clustershort{} dataset, while naive GANs miss a category entirely; this appears to be due to mode collapse, which we mitigate with our second discriminator. 
Results on other datasets are in Appendix \ref{app:fidelity}.

\noindent\underline{\Measurement{}-\metadata{} correlations:}
Although our HMM, AR, and RNN  baselines learn perfect \metadata{} distributions, it is substantially more challenging (and important) to learn the joint \metadata{}-\measurement{} distribution.
To illustrate this, we compute the CDF of total bandwidth for DSL and cable users in \fccshort{} dataset. 
Table \ref{tbl:fcc-wd} shows the Wasserstein-1 distance between the generated CDFs and the ground truth,\footnote{
Wasserstein-1 distance is the integrated absolute error between 2 CDFs.}
showing that \name{} is closest to the real distribution.
CDF figures are attached in Appendix \ref{app:fidelity}.

\begin{table}[t]
	\centering
	\setlength\tabcolsep{3pt}
	\begin{tabular}{c|c|c|c|c|c}
		\toprule
		& DoppelGANger & AR & RNN & HMM & Naive GAN\\
		\midrule
		DSL  & \textbf{0.68} & 1.34 & 2.33 & 3.46& 1.14\\
		Cable  & \textbf{0.74}  & 6.57 & 2.46& 7.98 & 0.87\\
		\bottomrule
	\end{tabular}
	\caption{Wasserstein-1 distance of total bandwidth distribution of DSL and cable users. Lower is better.}
	\vspace{-0.6cm}
	\label{tbl:fcc-wd}
\end{table}

\noindent  \underline{\name{} does not overfit:}
In the context of GANs, overfitting is equivalent to memorizing the training data, which is a common concern  with GANs \cite{arora2017gans,odena2017conditional}. 
To evaluate this, we ran an experiment inspired by the methodology of \cite{arora2017gans}: for a given generated \name{} sample, we find its nearest samples in the training data. 
We  observe significant differences (both in square error and qualitatively) between the generated samples and the nearest neighbors on all datasets, suggesting that \name{} is not memorizing. 
Examples can be found in Appendix \ref{app:fidelity}.

\begin{figure}[t]
	\centering
	\includegraphics[width=0.8\linewidth]{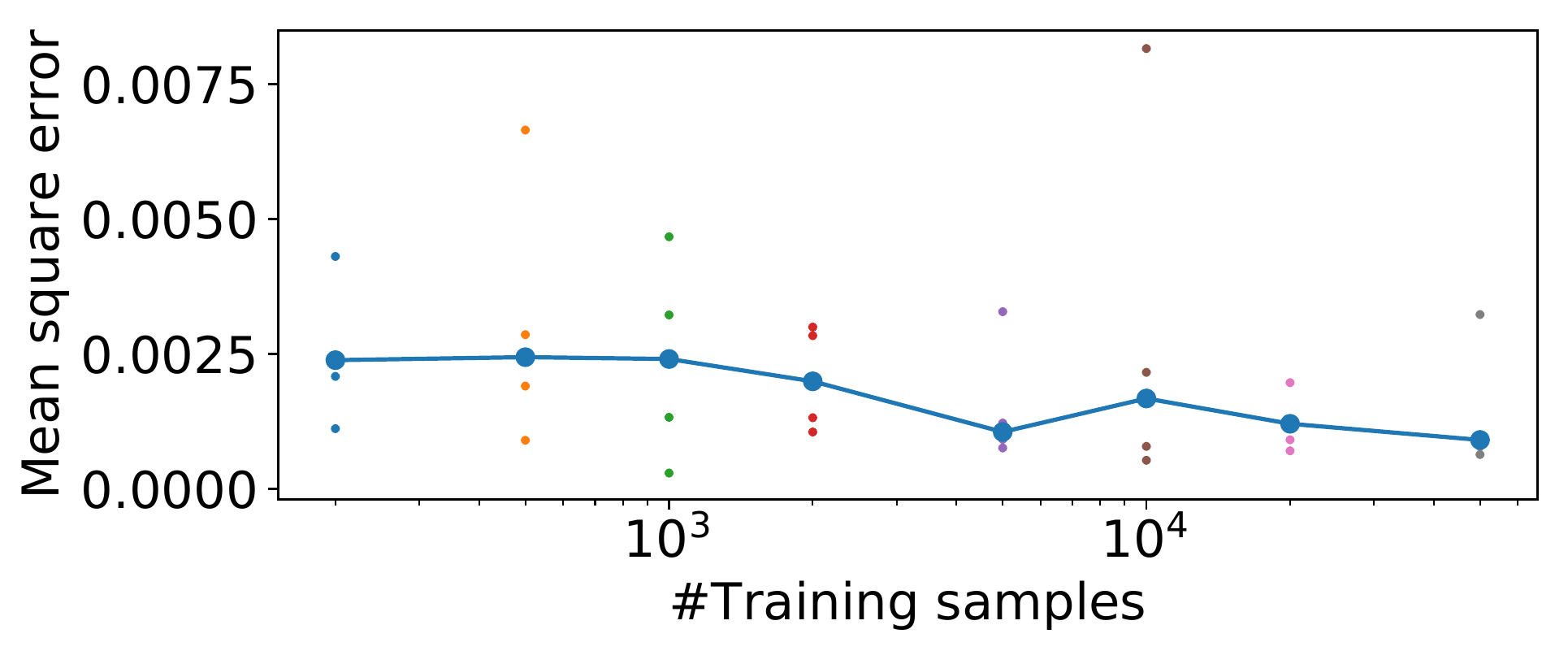}
	\caption{Mean square error of autocorrelation of the daily page views v.s. number of training samples for \wikishort{} dataset. For each training set size, 5 independent runs are executed and their MSE are plotted in the figure. The line connects the median MSE of the 5 independent runs.}
	\label{fig:wiki-autocorr-mse-training-size}
\end{figure}

\noindent\underline{Resource costs:}
\name{} has two main costs: training data and training time/computation. 
In Figure \ref{fig:wiki-autocorr-mse-training-size}, we plot the mean square error (MSE) between the generated samples' autocorrelations and the real data's autocorrelations on the \wikishort{} dataset as a function of training set size.
MSE is sensitive to training set size---it decreases by 60\% as the training data grows by 2 orders of magnitude.
However, Table \ref{tbl:wiki-autocorr-mse} shows that \name{} trained on 500 data points (the size that gives DG the worst performance) still outperforms all  baselines trained on 50,000 samples in autocorrelation MSE.
Figure \ref{fig:wiki-autocorr-mse-training-size} also illustrates variability between models; due to GAN training instability, different GAN models with the same hyperaparameters can have different fidelity metrics.
Such training failures can typically be detected early in the training proccess. 

\begin{table}[t]
	\begin{tabular}{l|c}
		\toprule
		Method & MSE\\
		\midrule
		RNN & 0.1222\\
		AR & 0.2777\\
		HMM & 0.6030\\
		Naive GAN & 0.0190\\
		TimeGAN & 0.2384\\
		RCGAN & 0.0103\\
		MarketSimulator & 0.0324\\
		\hline
		DoppelGANger & 0.0009\\
		DoppelGANger (500 training samples) & 0.0024\\
		\bottomrule
	\end{tabular}
	\caption{Mean square error (MSE) of autocorrelation of the daily page views for \wikishort{} dataset (i.e. quantitative presentation of Figure \ref{fig:wiki-autocorr}). Each model is trained with multiple independent runs, and the median MSE among the runs is presented.  Except the last row, all models are trained with 50000 training samples.}
	\label{tbl:wiki-autocorr-mse}
\end{table}

With regards to training time, Table \ref{tab:training-time} lists the training time for \name{} and other baselines. 
All models were trained on a single NVIDIA Tesla V100 GPU with 16GB GPU memory and an Intel Xeon Gold 6148 CPU with 24GB RAM. 
These implementations have not been optimized for performance at all, but we find that on the \wikishort{} dataset, \name{} requires 17 hours on average to train, which is $3.4\times$ slower than the fastest benchmark (Naive GAN) and $15.2\times$ faster than the slowest benchmark (TimeGAN). 

\begin{table}[t]
	\begin{tabular}{l|c}
		\toprule
		Method & Average training time (hrs)\\
		\midrule
		Naive GAN & 5\\
		Market Simulator & 6\\
		HMM & 8\\
		RNN & 22\\
		RCGAN & 29\\
		AR & 93\\
		TimeGAN & 258\\
		\hline
		DoppelGANger & 17\\
		\bottomrule
	\end{tabular}
	\caption{Average training time (hours) of synthetic data models on the \wikishort{} dataset. 
	All models are trained with 50000 training samples.}
	\vspace{-0.5cm}
	\label{tab:training-time}
\end{table}

\myparatight{Predictive modeling}
Given time series \measurements{}, users may want to predict whether an event $E$ occurs in the future, or even forecast the time series itself.
For example, in \clustershort{} dataset, we could predict whether a particular job will complete successfully.
In this use case, we want to show that models trained on generated data generalize to real data.

\begin{figure}[t]
	\centering
	\includegraphics[width=0.7\linewidth]{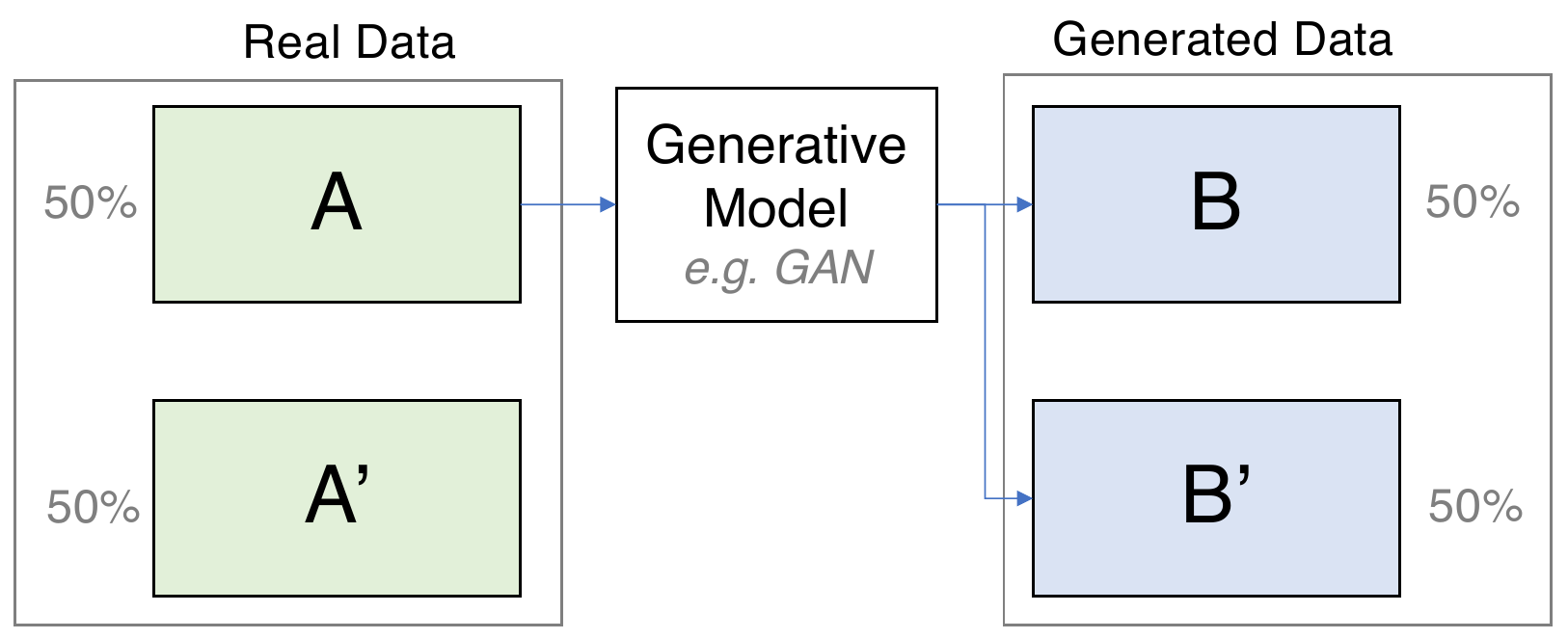}
	\vspace{-0.3cm}
	\caption{Predictive modeling setup:   Using training data $A$, we  generate  samples $B\cup B'$. Subsequent experiments  train  downstream tasks on $A$ or $B$, our training sets, and then test on $A'$ or $B'$.}
	\vspace{-0.5cm}
	\label{fig:evaluation-set}
\end{figure}

We first partition our dataset, as shown in Figure \ref{fig:evaluation-set}.
We split real data into two sets of equal size: a training set A and a test set A'. 
We then train a generative model (e.g., \name{} or a baseline) on training set A. 
We generate datasets B and B' for training and testing. 
Finally, we evaluate event prediction algorithms by training a predictor on A and/or B, and testing on A' and/or B'.
This allows us to compare the generalization abilities of the prediction algorithms both within a class of data (real/generated), and generalization across classes (train on generated, test on real) \cite{esteban2017real}.

\begin{figure}[t]
	\centering
	\includegraphics[width=0.9\linewidth]{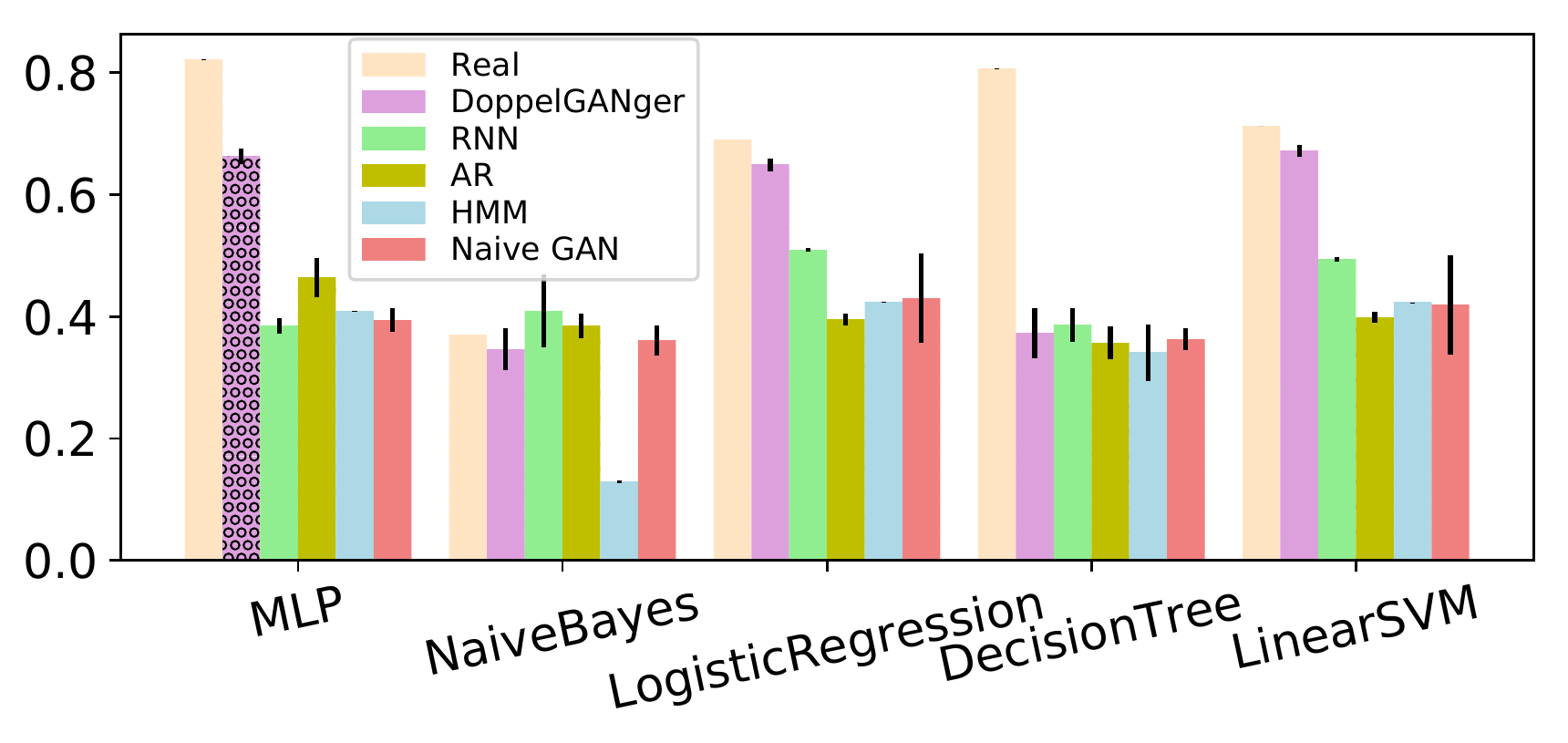}
	\put(-225,40){\rotatebox{90}{Accuracy}}
	\vspace{-0.4cm}
	\caption{ Event-type prediction accuracy on \clustershort{}.}
	\vspace{-0.45cm}
	\label{fig:google-acc}
\end{figure}

We first predict the task end event type on \clustershort{} data (e.g., EVICT, KILL) from time series observations. 
Such a predictor may be useful for cluster resource allocators.
This prediction task reflects the correlation between the time series and underlying \metadatas{} (namely, end event type). 
For the predictor, we trained various algorithms to demonstrate the generality of our results: multilayer perceptron (MLP), Naive Bayes, logistic regression, decision trees, and a linear SVM.
Figure \ref{fig:google-acc} shows the test accuracy of each predictor when trained on generated data and tested on real. 
Real data expectedly has the highest test accuracy.
However, we find that \name{} performs better than other baselines for all five classifiers.
For instance, on the MLP predictive model, \name{}-generated data has 43\% higher accuracy than the next-best baseline (AR), and 80\% of the real data accuracy.
The results on other datasets are similar. (Not shown for brevity; but  in Appendix \ref{app:downstream} for completeness.)

\myparatight{Algorithm comparison}
We evaluate whether algorithm rankings are preserved on generated data 
on the \clustershort{} dataset by training different classifiers (MLP, SVM, Naive Bayes, decision tree, and logistic regression) to do end event type classification. 
We also evaluate this on the \wikishort{} dataset by training different regression models (MLP, linear regression, and Kernel regression) to do time series forecasting (details in Appendix \ref{app:downstream}).  
For this use case, users have  only generated data,
so we want the ordering (accuracy) of algorithms on real data to be preserved when we train and test them on generated data. 
In other words, for each class of generated data, we train each of the predictive models on $B$ and test on $B'$.
This is different from Figure \ref{fig:google-acc}, where we trained on generated data ($B$) and tested on real data ($A'$).
We compare this ranking with the ground truth ranking, in which the predictive models are trained on $A$ and tested on $A'$.
We then compute the Spearman's rank correlation coefficient \cite{spearman1904proof}, which compares how the ranking in generated data is correlated with the groundtruth ranking. 
Table \ref{tbl:rank} shows that 
\name{} and AR achieve the best rank correlations.
This  result is misleading because AR models exhibit minimal randomness, so \emph{all} predictors achieve the same high accuracy; the AR model achieves near-perfect rank correlation despite producing low-quality samples; this highlights the importance of considering rank correlation together with other fidelity metrics.
More results (e.g., exact prediction  numbers) are in Appendix \ref{app:downstream}.

\begin{table}[t]
	\centering
	\setlength\tabcolsep{3pt}
	\begin{tabular}{c|c|c|c|c|c}
			\toprule
			& DoppelGANger & AR & RNN & HMM & Naive GAN\\
			\midrule
			\clustershort{} & \textbf{1.00} &\textbf{1.00} &\textbf{1.00} & 0.01 & 0.90\\
			\wikishort{} & \textbf{0.80}  & \textbf{0.80} & 0.20 & -0.60 & -0.60\\
			\bottomrule
	\end{tabular}
	\caption{Rank correlation of predication algorithms on \clustershort{} and \wikishort{} dataset. Higher is better.}
	\vspace{-0.8cm}
	\label{tbl:rank}
\end{table}

\subsection{Other case studies}
\begin{figure}
    \centering
    \includegraphics[width=0.75\linewidth]{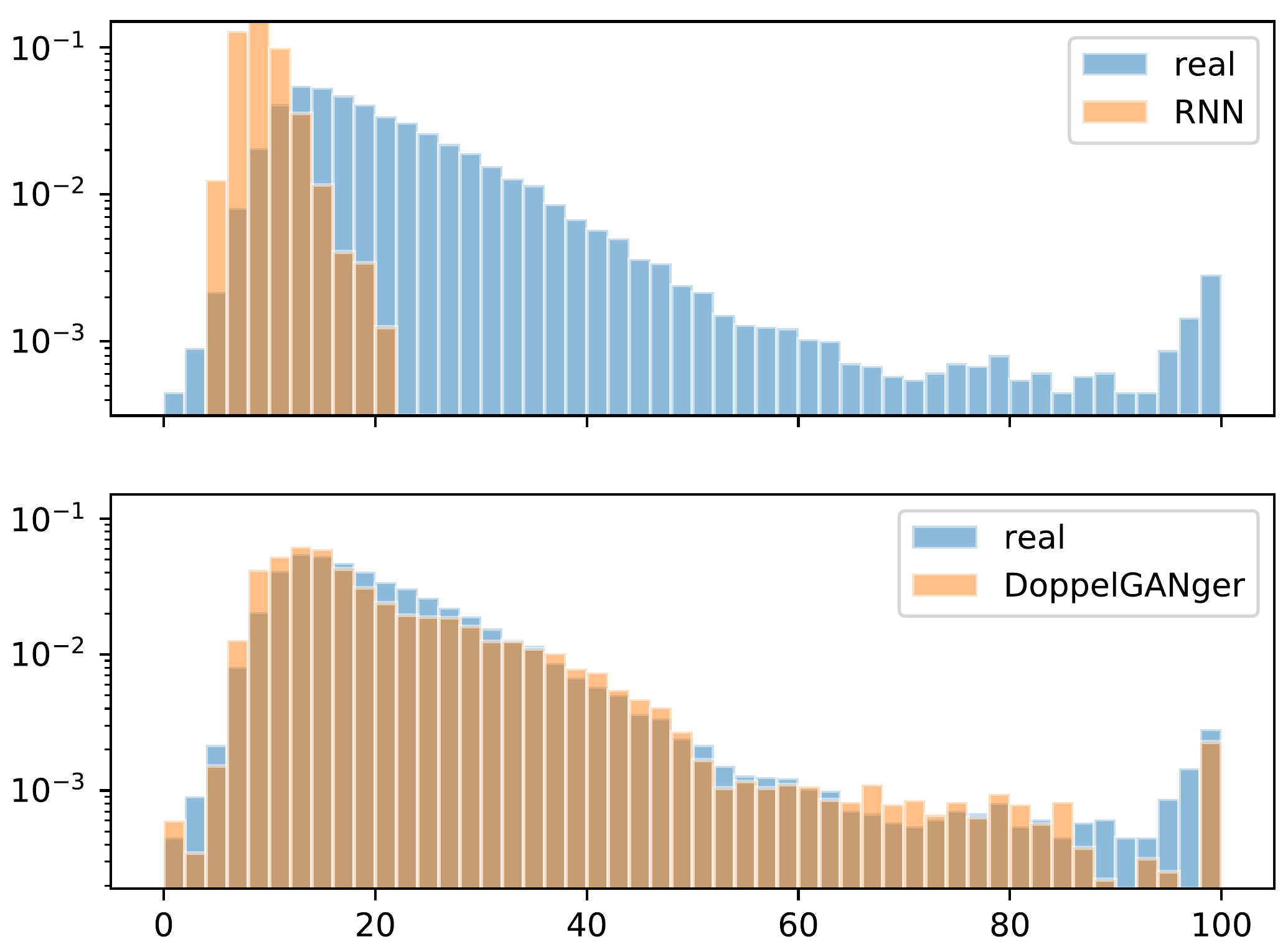}
    \vspace{-0.2cm}
    \caption{Maximum CPU usage.}
    \vspace{-0.3cm}
    \label{fig:cpu-ibm}
\end{figure}

 \name{} is being evaluated by several independent users, though \name{} has not yet been used to release any datasets to the best of our knowledge.
A large public cloud provider (IBM) has internally validated the fidelity of \name{}. 
IBM  stores time series data of resource usage measurements for different containers
used in the cloud's machine learning service. 
They trained \name{} to generate resource usage  \measurements{},  with the container image name as  \metadata{}. 
Figure \ref{fig:cpu-ibm} shows the learned distribution of containers' maximum CPU usage (in percent). 
We show the maximum because developers usually size containers according to their peak usage. 
\name{} captures this challenging distribution very well, even in the heavy tail.
Additionally, other users outside of the networking/systems domain such as banking, sensor data, and natural resource modeling have also been using \name{} \cite{hazy,boogiesoftware}.

\section{Privacy Analysis and Tradeoffs}
\label{sec:exp-privacy}
Data holders' privacy concerns often fit in two categories: protecting business secrets (\S\ref{sec:design-private-attribute}) and protecting user privacy (\S\ref{sec:design-dp-gans}). 
In this section, we illustrate what GANs can and cannot do in each of these topics.

\subsection{Protecting business secrets} 
\label{sec:design-private-attribute}
In our discussions with major data holders, a primary concern about data sharing is leaking information about the types of resources available and in use at the enterprise. 
Many such business secrets tend to be embedded in the \metadatas{} (e.g., hardware types in a compute cluster).
Note that in this case the correlation between hardware type and other \metadatas{}/\measurements{} might be important for downstream applications, so data holders cannot simply discard hardware type from data. 

There are two ways to change or obfuscate the \metadata{} distribution. The naive way is to rejection sample the \metadata{} to a different desired distribution. This approach is clearly inefficient. The architecture in \S\ref{sec:design-correlation} naturally allows data holders to obfuscate the \metadata{} distribution in a much simpler way. After training on the original dataset, the data holders can retrain \emph{only} the \metadata{} generator to any desired distribution as the \metadata{} generator and \measurement{} generator are isolated.
Doing so requires synthetic \metadata{} of the desired distribution, but it does not require new  time series data.

A major open  question  is how to realistically tune attribute distributions. 
Both  of the above approaches to obfuscating attribute distributions keep  the conditional  distribution $\mathbb P(R^i | A^i)$ unchanged,  even as we alter the marginal  distribution  $\mathbb P(A^i)$.
While this may be a reasonable approximation for small  perturbations, changing the marginal   $\mathbb P(A^i)$ may affect the conditional  distribution in general. 
For example, $A^i$ may represent the fraction  of large jobs in a system, and $R^i$ the memory usage  over time; 
if  we increase $A^i$, at some point we should encounter resource exhaustion. 
However, since the GAN  is trained only on input data,  it cannot predict such effects. 
Learning to train GANs with  simulators that can  model physical constraints of  systems may be a good way to combine  the statistical learning properties of GANs with systems that encode domain knowledge.

\begin{figure}[ht]
	\centering
	\includegraphics[width=0.75\linewidth]{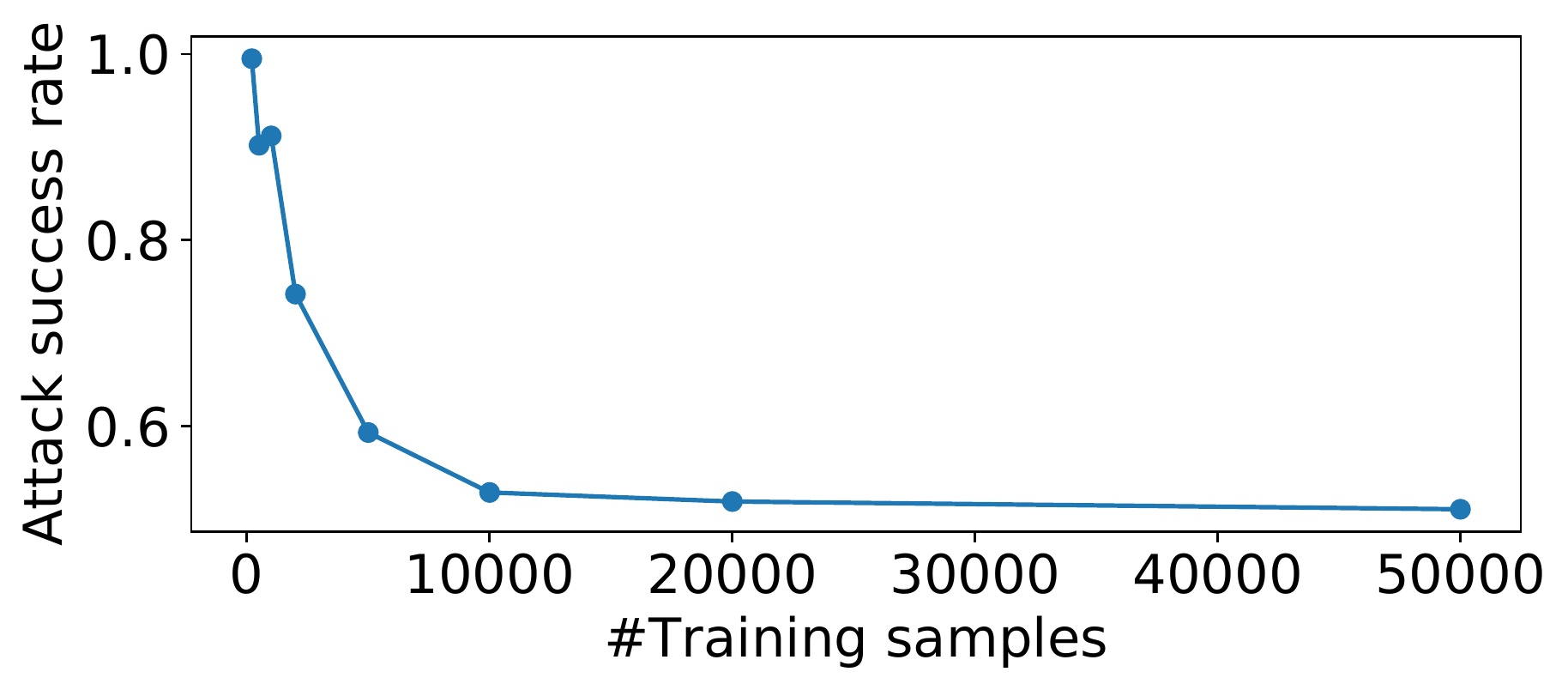}
	\vspace{-0.3cm}
	\caption{Membership inference attack against \name{} in \wikishort{} dataset vs.\ training set size. }
	\vspace{-0.45cm}
	\label{fig:wiki-membership-num-training}
\end{figure}

\begin{figure*}[ht]
    \centering
    \includegraphics[width=0.85\linewidth]{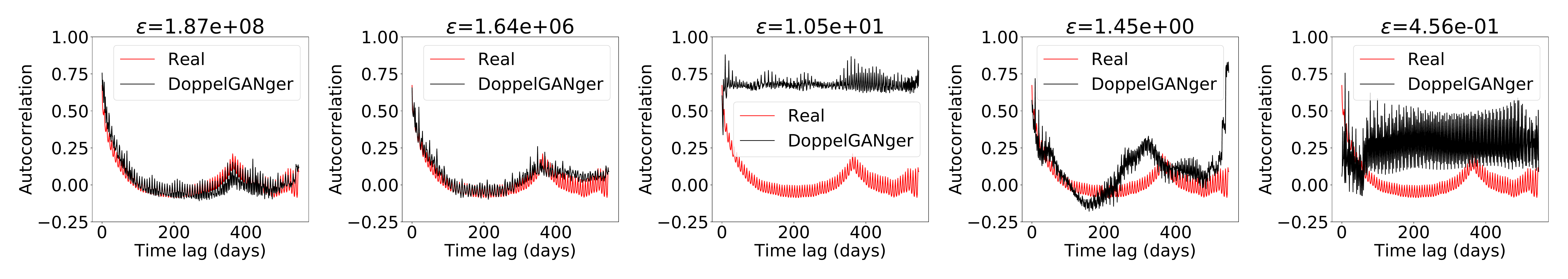}
    \vspace{-0.3cm}
    \caption{Autocorrelation for real and DP-GANs with different values of $\epsilon$.}
    \label{fig:dp_one}
	\vspace{-0.3cm}
\end{figure*}
\subsection{Protecting user privacy}
\label{sec:design-dp-gans}
User privacy is a  major concern with regards to any data sharing application, and  generative models pose unique challenges.
 For example,  recent work has shown  that generative models may memorize a specific user's information from the training data (e.g. a social security number) and leak it during data generation \cite{carlini2019secret}.
 Hence, we need methods for evaluating and protecting user privacy.
One challenge is the difficulty of defining what it means to be privacy-preserving. 
Although there are many competing privacy definitions \cite{k-anon,dwork2008differential,sankar2013utility}, a common theme is \emph{deniability}: 
 released data or models should look similar whether a given user's data is included or not.

In this section,  we show how two of the most common notions of deniability  relate to GANs. 
 
\noindent \textbf{Differential privacy:}
A leading metric for measuring user privacy  today is
differential privacy (DP) \cite{dwork2008differential}. 
In the context of machine learning, DP states that the trained model should not depend too much on any individual user's data. 
More precisely, a model $\mathcal M$ is $(\epsilon,\delta)$ differentially private if for any pair of training datasets $\mathcal D$ and $\mathcal D'$ that differ in  the record of a single user,  and for any input $z$, it holds that
$$
\mathcal M(z; \mathcal D) \leq e^\epsilon \mathcal M(z; \mathcal D') + \delta,
$$
where $\mathcal M(z; \mathcal D)$ denotes a model trained on $\mathcal D$ and evaluated on $z$. Smaller values of $\epsilon$ and $\delta$ give more privacy. 

Recent work has explored how to make deep learning models differentially private \cite{abadi2016deep} by clipping and adding noise to the gradient updates in stochastic gradient descent to ensure that any single example in the training dataset does not disproportionately influence the model parameters.
Researchers have applied this technique to GANs \cite{xie2018differentially, xu2019ganobfuscator, frigerio2019differentially} to generate privacy-preserving time series \cite{esteban2017real, beaulieu2019privacy}; we generally call this approach DP-GAN. 
These papers argue that such DP-GAN architectures give privacy at minimal cost to utility. 
A successive paper using DP student-teacher learning achieved better performance than DP-GAN on single-shot data \cite{pate-gan}, but we do not consider it here because  its architecture is ill-suited to generating time series.

To evaluate the efficacy of DP-GANs in our context, we trained DoppelGANger with DP-GANs for the WWT dataset using TensorFlow Privacy~\cite{andrew2019tensorflow}.\footnote{We were unable to implement DP-TimeGAN for comparison, as their code uses loss functions that are not supported by Tensorflow Privacy  \cite{timegan-code}.} 
Figure \ref{fig:dp_one} shows the autocorrelation of the resulting time series for different values of the privacy budget, $\epsilon$. 
Smaller values of $\epsilon$ denote more privacy; $\epsilon \approx 1$ is typically considered a reasonable operating point.
As $\epsilon$ is reduced (stronger privacy guarantees), autocorrelations become progressively worse. In this figure, we show results only for the 15th epoch, as the results become only worse as training proceeds  (not shown). 
These results highlight an important point:  although DP-GANs seems to destroy our  autocorrelation plots, \emph{this was not always evident from downstream metrics, such as predictive accuracy in \cite{esteban2017real}}. 
This highlights the need to evaluate generative time series models  qualitatively \emph{and} quantitatively; prior work has focused mainly  on  the latter \cite{esteban2017real, beaulieu2019privacy,pate-gan}. 
Our results  
suggest that  DP-GAN mechanisms require significant improvements for privacy-preserving time series generation.

\myparatight{Membership  inference}
Another common way of evaluating user deniability is through  \emph{membership inference attacks} \cite{shokri2017membership,hayes2019logan,chen2019gan}.
Given a trained machine learning model and set of data samples, the goal of such an attack is to infer whether those samples were in the training dataset.
The attacker does this by training a classifier to output whether each  sample was in the training data.
Note that differential privacy should protect against such an attack; the stronger the privacy parameter, the lower the success rate.
However, DP alone does quantify the efficacy of membership  inference attacks.

To  understand this question further,  we measure \name{}'s vulnerability to membership inference attacks \cite{hayes2019logan} on the \wikishort{} dataset. 
As in \cite{hayes2019logan}, our metric is success rate, or the percentage of successful trials in guessing whether a sample is in the training dataset. 
Naive random guessing gives 50\%, 
while we found an attack success rate of 51\%, suggesting robustness to membership inference \emph{in this case}.
However,  when we decrease the training set size, the attack success rate increases (Figure \ref{fig:wiki-membership-num-training}). 
For instance, with 200 training samples, the attack success rate is as high as 99.5\%. 
Our results suggest a practical guideline:  to be more robust against membership attacks, use more training data. 
This contradicts the common practice of subsetting for better privacy \cite{reiss2012obfuscatory}. 

\myparatight{Summary and implications}
The privacy properties of GANs appear to be mixed. 
On  the positive side, GANs can be used to obfuscate attribute distributions for masking business secrets, and there appear to be simple techniques for preventing common membership inference attacks (namely,  training on more data).
However, the primary challenge appears to be preserving good fidelity while providing strong \emph{theoretical} privacy guarantees.
We find that existing approaches for providing DP destroy fidelity beyond recognition, and are not a viable solution.

More broadly, we believe there is value  to exploring different privacy metrics. 
DP is designed for datasets where one sample corresponds to one user.
However, many datasets of interest have multiple time series from the same user,
and/or  time series may not correspond to  users at all. 
In these cases,  DP gives hard-to-interpret guarantees, while destroying data fidelity 
(particularly for uncommon signal classes \cite{bagdasaryan2019differential}).
Moreover, it does not defend against other attacks like model inversion \cite{fredrikson2014privacy,hitaj2017deep}.
So it is not clear whether DP is a good metric for this class of data.

\section{Conclusions}
\label{sec:conclude}

While  \name{} is a promising    general  workflow for data sharing, 
  the privacy properties of  GANs   require further research for data holders to confidently use such workflows.   
  Moreover, many networking datasets require significantly more complexity than \name{} is currently able to handle, such as causal interactions between stateful agents.
  Another direction of interest is to enable ``what-if" analysis, in which practitioners can model changes in the underlying system and generate associated data.
  Although \name{} makes some what-if analysis easy (e.g., slightly altering the attribute distribution), larger changes may alter the physical system model such that the conditional distributions learned by \name{} are invalid (e.g., imagine simulating a high-traffic regime with a model trained only on low-traffic-regime data). 
  Such what-if analysis is likely to require physical system modeling/simulation, while GANs may be able to help model individual agent behavior.
  We hope that the initial promise  and open questions  inspire further work from  theoreticians and practitioners to help break the  impasse in data sharing. 
  
   \noindent {\bf Data/code release and Ethics:}   While we highlight  privacy concerns in using GAN-based models,   the code/data we are releasing does not raise any ethical issues as the public datasets  do not contain any personal information.

\section*{Acknowledgements}
The authors would like to thank Shivani Shekhar for assistance reviewing and evaluating related work, 
and Safia
Rahmat, Martin Otto, and Abishek Herle for valuable discussions.
The authors would also like to thank Vijay Erramilli and the reviewers for insightful suggestions.
This work was supported in part by faculty research awards from Google and JP Morgan Chase, as well as a gift from Siemens AG.
This research was sponsored in part by National Science Foundation Convergence Accelerator award 2040675 and the U.S. Army Combat
Capabilities Development Command Army Research Laboratory under Cooperative Agreement
Number W911NF-13-2-0045 (ARL Cyber Security CRA).
The views and conclusions contained in this document are
those of the authors and should not be interpreted as representing the official policies, either expressed or implied, of the
Combat Capabilities Development Command Army Research
Laboratory or the U.S. Government. The U.S. Government
is authorized to reproduce and distribute reprints for Government purposes notwithstanding any copyright notation here
on.
This work used the Extreme Science and Engineering Discovery Environment (XSEDE) \cite{xsede}, which is supported by National Science Foundation grant number ACI-1548562. Specifically, it used the Bridges system \cite{bridges}, which is supported by NSF award number ACI-1445606, at the Pittsburgh Supercomputing Center (PSC).

\bibliographystyle{ACM-Reference-Format}
\bibliography{main}

\clearpage
\appendix
\section*{Appendix}

\section{Datasets}
\label{app:dataset}

\myparatightest{Google Cluster Usage Trace}
Due to the substantial computational requirements of training GANs and our own resource constraints, we did not use the entire dataset. 
Instead, we uniformly sampled a subset of 100,000 tasks and used their corresponding measurement records to form our dataset. 
This sample was collected after filtering out the following categories:
\begin{packeditemize}
	\item 197 (0.17\%) tasks don't have corresponding end events (such events may end outside the data collection period)
	\item 1403 (1.25\%) tasks have discontinuous measurement records (i.e., the end timestamp of the previous measurement record does not equal the start timestamp of next measurement record)
	\item 7018 (6.25\%) tasks have an empty measurement record
	\item 3754 (3.34\%) tasks have mismatched end times (the timestamp of the end event does not match the ending timestamp of the last measurement).
\end{packeditemize}
The maximum \measurement~length in this dataset is 2497, however, 97.06\% samples have length within 50.
The schema of this dataset is in Table \ref{tbl:google-schema}.

\myparatightest{Wikipedia Web Traffic}
The original datasets consists of 145k \samples. After removing samples with missing data, 117k \samples~are left, from which we sample 100k \samples~for our evaluation. All samples have \measurement~length 550.
The schema of this dataset is in Table \ref{tbl:web-schema}.

\myparatightest{FCC MBA}
We used the latest cleaned data published by FCC MBA in December 2018 \cite{mba-data}. This datasets contains hourly traffic measurements from 4378 homes in September and October 2017. However, a lot of measurements are missing in this dataset. Considering period from 10/01/2017 from 10/15/2017, only 45 homes have complete network usage measurements every hour. This small sample set will make us hard to understand the actual dynamic patterns in this dataset. To increase number of valid \samples, we take the average of measurements every 6 hours for each home. As long as there is at least one measurement in each 6 hours period, we regard it as a valid \sample. Using this way, we get 739 valid \samples~with measurements from 10/01/2017 from 10/15/2017, from which we sample 600 \samples~for our evaluation. All samples have \measurement~length 56.
The schema of this dataset is in Table \ref{tbl:mba-schema}.

\begin{table}[htb]
	\begin{tabular}{m{2.3cm}|m{2.5cm}|m{2.5cm}}
		\toprule
		\multicolumn{1}{c|}{\textbf{\Metadatas}}& \multicolumn{1}{c|}{\textbf{Description}} & \multicolumn{1}{c}{\textbf{Possible Values}}\\
		\midrule
		end event type & The reason that the task finishes & FAIL, KILL, EVICT, etc.\\ 
		
		\hline \hline
		\multicolumn{1}{c|}{\textbf{\Measurements}}& \multicolumn{1}{c|}{\textbf{Description}} & \multicolumn{1}{c}{\textbf{Possible Values}}\\
		\midrule
		CPU rate & Mean CPU rate & float numbers \\
		\hline
		maximum CPU rate & Maximum CPU rate & float numbers \\
		\hline
		sampled CPU usage & The CPU rate sampled uniformly on all 1 second measurements & float numbers \\
		\hline
		canonical memory usage & Canonical memory usage measurement & float numbers \\
		\hline
		assigned memory usage & Memory assigned to the container & float numbers\\
		\hline
		maximum memory usage & Maximum canonical memory usage & float numbers\\
		\hline
		unmapped page cache & Linux page cache that was not mapped into any userspace process & float numbers\\
		\hline
		total page cache & Total Linux page cache& float numbers \\
		\hline
		local disk space usage & Runtime local disk capacity usage & float numbers \\
		
		\hline \hline
		\multicolumn{2}{c|}{\textbf{Timestamp Discription}} & \multicolumn{1}{c}{\textbf{Possible Values}}\\
		\hline
		\multicolumn{2}{m{4.9cm}|}{The timestamp that the measurement was conducted on. Different task may have different number of measurement records (i.e. $T^i$ may be different)} & 2011-05-01 01:01, etc.\\
		\bottomrule
	\end{tabular}
	\caption{Schema of \clustershort{} dataset. \Metadatas{} and \measurements{} are described in more detail in \cite{reiss2011google}.}
	\label{tbl:google-schema}
\end{table}

\begin{table}[htb]
	\begin{tabular}{m{2.3cm}|m{2.5cm}|m{2.5cm}}
		\toprule
		\multicolumn{1}{c|}{\textbf{\Metadatas}}& \multicolumn{1}{c|}{\textbf{Description}} & \multicolumn{1}{c}{\textbf{Possible Values}}\\
		\midrule
		Wikipedia domain & The main domain name of the Wikipedia page  & zh.wikipedia.org, commons.wikimedia.org, etc.\\ 
		\hline
		access type & The access method & mobile-web, desktop, all-access, etc. \\
		\hline
		agent & The agent type & spider, all-agent, etc.\\
		
		\hline \hline
		\multicolumn{1}{c|}{\textbf{\Measurements}}& \multicolumn{1}{c|}{\textbf{Description}} & \multicolumn{1}{c}{\textbf{Possible Values}}\\
		\midrule
		views & The number of views & integers\\
		
		\hline \hline
		\multicolumn{2}{c|}{\textbf{Timestamp Discription}} & \multicolumn{1}{c}{\textbf{Possible Values}}\\
		\hline
		\multicolumn{2}{m{4.9cm}|}{The date that the page view is counted on} & 2015-07-01, etc.\\
		
		\bottomrule
	\end{tabular}
	\caption{Schema of \wikishort~dataset.}
	\label{tbl:web-schema}
\end{table}

\begin{table}[htb]
	\begin{tabular}{m{2.3cm}|m{2.5cm}|m{2.5cm}}
		\toprule
		\multicolumn{1}{c|}{\textbf{\Metadatas}}& \multicolumn{1}{c|}{\textbf{Description}} & \multicolumn{1}{c}{\textbf{Possible Values}}\\
		\midrule
		technology & The connection technology of the unit & cable, fiber, etc.\\ 
		\hline
		ISP & Internet service provider of the unit & AT\&T, Verizon, etc. \\
		\hline
		state & The state where the unit is located& PA, CA, etc.\\
		
		\hline \hline
		\multicolumn{1}{c|}{\textbf{\Measurements}}& \multicolumn{1}{c|}{\textbf{Description}} & \multicolumn{1}{c}{\textbf{Possible Values}}\\
		\midrule
		ping loss rate & UDP ping loss rate to the server that has lowest loss rate within the hour & float numbers\\
		\hline
		traffic byte counter & Total number of bytes sent and received in the hour (excluding the traffic due to the activate measurements) & integers \\
		
		\hline \hline
		\multicolumn{2}{c|}{\textbf{Timestamp Discription}} & \multicolumn{1}{c}{\textbf{Possible Values}}\\
		\hline
		\multicolumn{2}{m{4.9cm}|}{The time of the measurement hour} & 2015-09-01 1:00, etc.\\
		
		\bottomrule
	\end{tabular}
	\caption{Schema of \fccshort~dataset.}
	\label{tbl:mba-schema}
\end{table}

\section{Implementation Details}
\label{app:implementation}
\myparatightest{\name}
\Metadata~generator and min/max generator are MLPs with 2 hidden layers and 100 units in each layer. \Measurement~generator is 1 layer of LSTM with 100 units. Softmax layer is applied for categorical \measurement~and \metadata~output. Sigmoid or tanh is applied for continuous \measurement~and \metadata~output, depending on whether data is normalized to [0,1] or [-1,1] (this is configurable). The discriminator and auxiliary discriminator are MLP with 4 hidden layers and 200 units in each layer. Gradient penalty weight was 10.0 as suggested in \cite{gulrajani2017improved}. The network was trained using Adam optimizer with learning rate of 0.001 and batch size of 100 for both generators and discriminators.

\noindent\underline{Loss functions:} As mentioned in \S\ref{sec:challenges},  Wasserstein loss has been widely adopted for improving training stability and alleviating mode collapse.
In our own empirical explorations, we find that Wasserstein loss is better than the original loss for generating categorical variables.
Because categorical variables are prominent in our domain, we use Wasserstein loss.

In order to train the two discriminators simultaneously, we combine the loss functions of the two discriminators by a weighting parameter $\alpha$. More specifically, the loss function is
\begin{align}
	\min_G \max_{D_1,D_2} \mathcal{L}_1(G,D_1) + \alpha \mathcal{L}_2(G,D_2)
\end{align} 
where $\mathcal{L}_i$, $i\in\{1,2\}$ is the Wasserstein loss of the original and second discriminator, respectively:
$
	\mathcal{L}_i = \Eb_{x~\sim p_x}\brb{T_i(D_i(x))} - \Eb_{z\sim p_z} \brb{D_i( T_i(G(z)) )} - \lambda \Eb_{\hat{x}\sim p_{\hat{x}}}\brb{\bra{ \brn{\nabla_{\hat{x}}D_i(T_i(\hat{x}))}_2 - 1 }^2}
$.
Here $T_1(x)=x$ and $T_2(x)=$ \metadata~part of $x$. $\hat{x} := tx + (1-t)G(z)$ where $t\sim \text{Unif}[0,1]$. 
Empirically, we find that $\alpha=1$ is enough for getting good fidelity on \metadatas.
As with all GANs, the generator and discriminators are trained alternatively until convergence. 
Unlike naive GAN architectures, we did not observe problems with training instability, and on our datasets, convergence required only up to 200,000 batches (400 epochs when the number of training samples is 50,000).

\begin{figure}[t]
    \centering
    \includegraphics[width=0.6\linewidth]{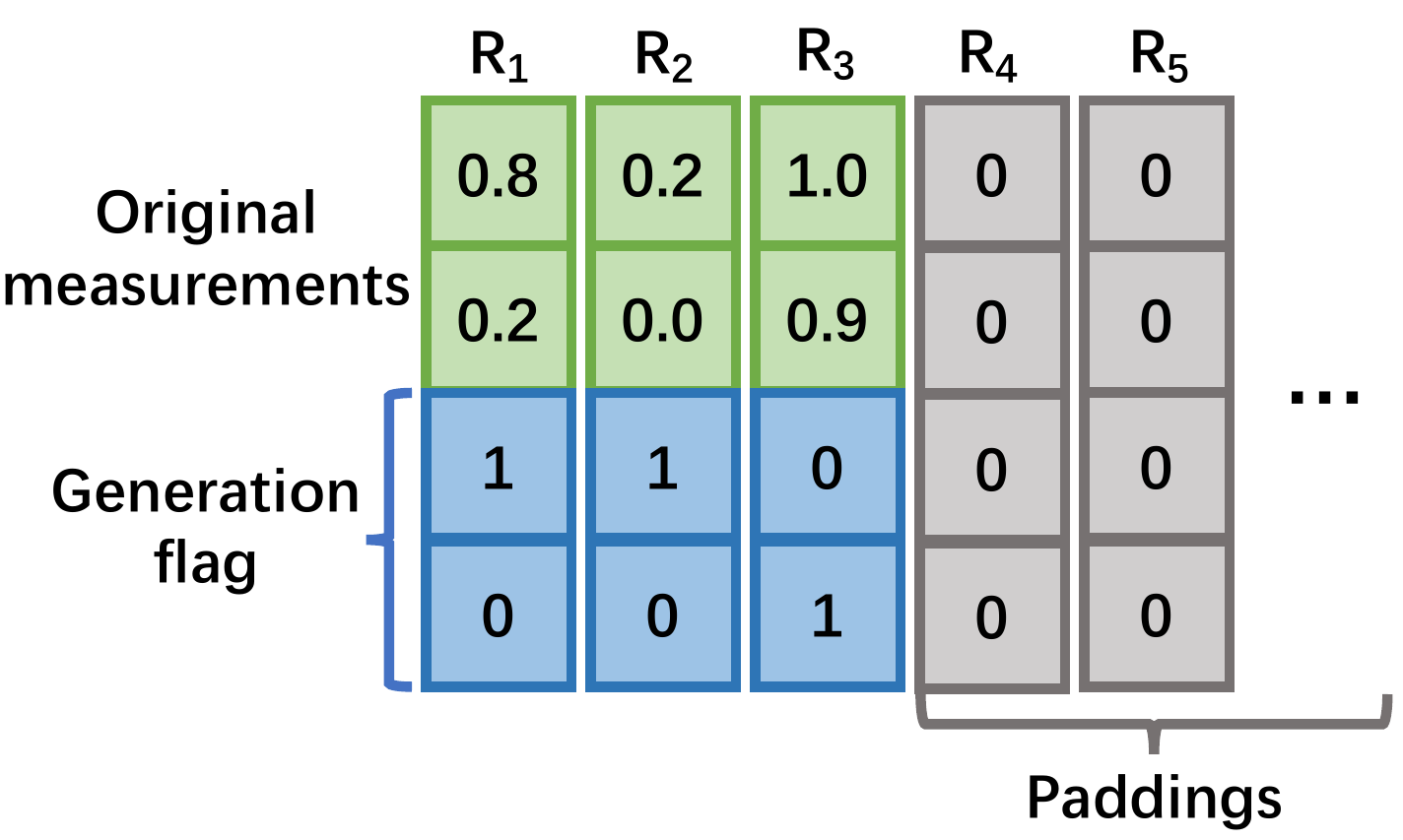}
    \caption{Generation flag for enabling variable length sequence.}
    \label{fig:generation-flag}
\end{figure}

\noindent\underline{Generation flag for variable lengths:}
Time series may have different lengths. For example, in \clustershort~dataset, different jobs have different duration (Figure \ref{fig:google-length-hist}).
We want to learn to generate sequences of the right length organically (without requiring the user to specify it).
The simplest solution to pad all time series with 0 to the same length. However, that would introduce a confusion on whether a zero value means the \measurements~(say, daily page view) is zero, or it is actually a padding token. Therefore, along with the original \measurements, we add generation flag to each time step: $[1, 0]$ if the time series does not end at this time step, and $[0,1]$ if the time series ends exactly at this time step (Figure \ref{fig:generation-flag}).
The generator outputs generation flag $[p_1,p_2]$ through a softmax output layer, so that $p_1,p_2\in [0,1]$ and $p_1+p_2=1$. $[p_1,p_2]$ is used to determine whether we should continue unrolling the RNN to the next time step. 
One way to interpret this is that $p_1$ gives the probability that the time series should continue at this time step. Therefore, 
if $p_1<p_2$, we stop generation and pad all future \measurements~with 0's; if $p_1>p_2$, we continue unrolling the RNN to generate \measurements~for the next time step(s). 
The generation flags are also fed to the discriminator as part of the features, so the generator can also learn sample length characteristics.

We also want to highlight that if the user wants to control the length of the generated samples, our architecture can also support this by iterating the RNN generator for the given desired number of steps.

\myparatightest{AR}
We used $p=3$, i.e., used the past three samples to predict the next. The AR model was an MLP with 4 hidden layers and 200 units in each layer. The MLP was trained using Adam optimizer \cite{kingma2014adam} with learning rate of 0.001 and batch size of 100.

\myparatightest{RNN}
For this baseline, we used LSTM (Long short term memory) \cite{hochreiter1997long} variant of RNN. It is 1 layers of LSTM with 100 units. The network was trained using Adam optimizer with learning rate of 0.001 and batch size of 100.

\myparatightest{Naive GAN}
The generator and discriminator are MLPs with 4 hidden layers and 200 units in each layer. Gradient penalty weight was 10.0 as suggested in \cite{gulrajani2017improved}. The network was trained using Adam optimizer with learning rate of 0.001 and batch size of 100 for both generator and discriminator.

\section{Additional Fidelity Results}
\label{app:fidelity}

\begin{figure}[t]
	\centering
	\includegraphics[width=0.7\linewidth]{figure_nsdi20/google_cluster/length_hist/DoppelGANger/length_hist_NSDI2020.pdf}
	\includegraphics[width=0.7\linewidth]{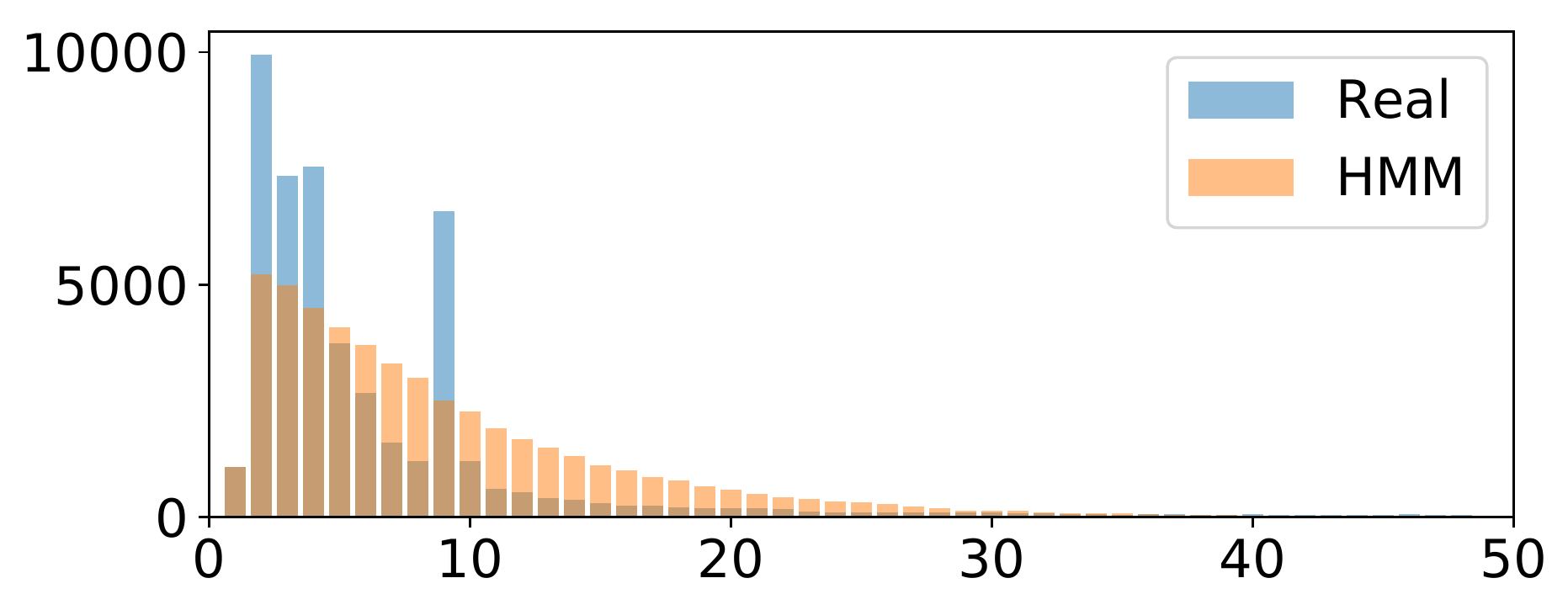}
	\begin{minipage}{1.0\linewidth}
		\centering
		\includegraphics[width=0.7\linewidth]{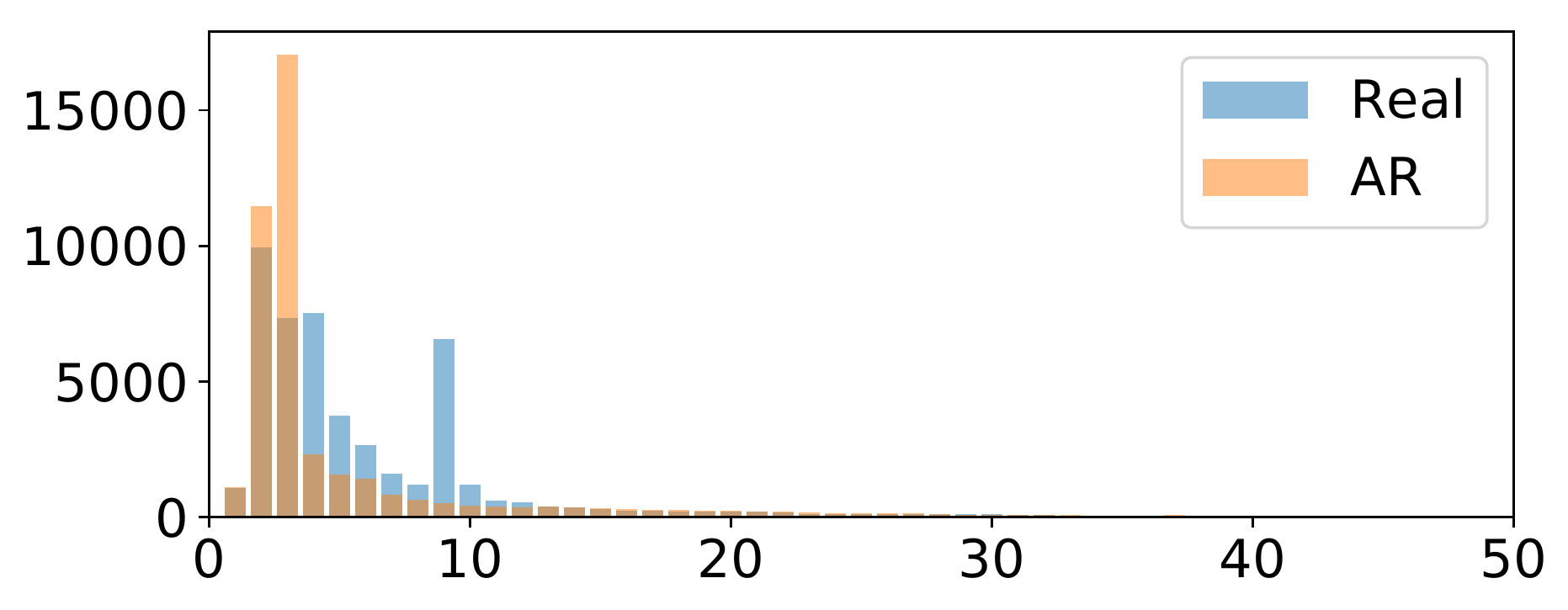}
		\put(-190,25){\rotatebox{90}{Count}}
	\end{minipage}
	\includegraphics[width=0.7\linewidth]{figure_nsdi20/google_cluster/length_hist/RNN/length_hist_NSDI2020.pdf}
	\begin{minipage}{1.0\linewidth}
		\centering
		\includegraphics[width=0.7\linewidth]{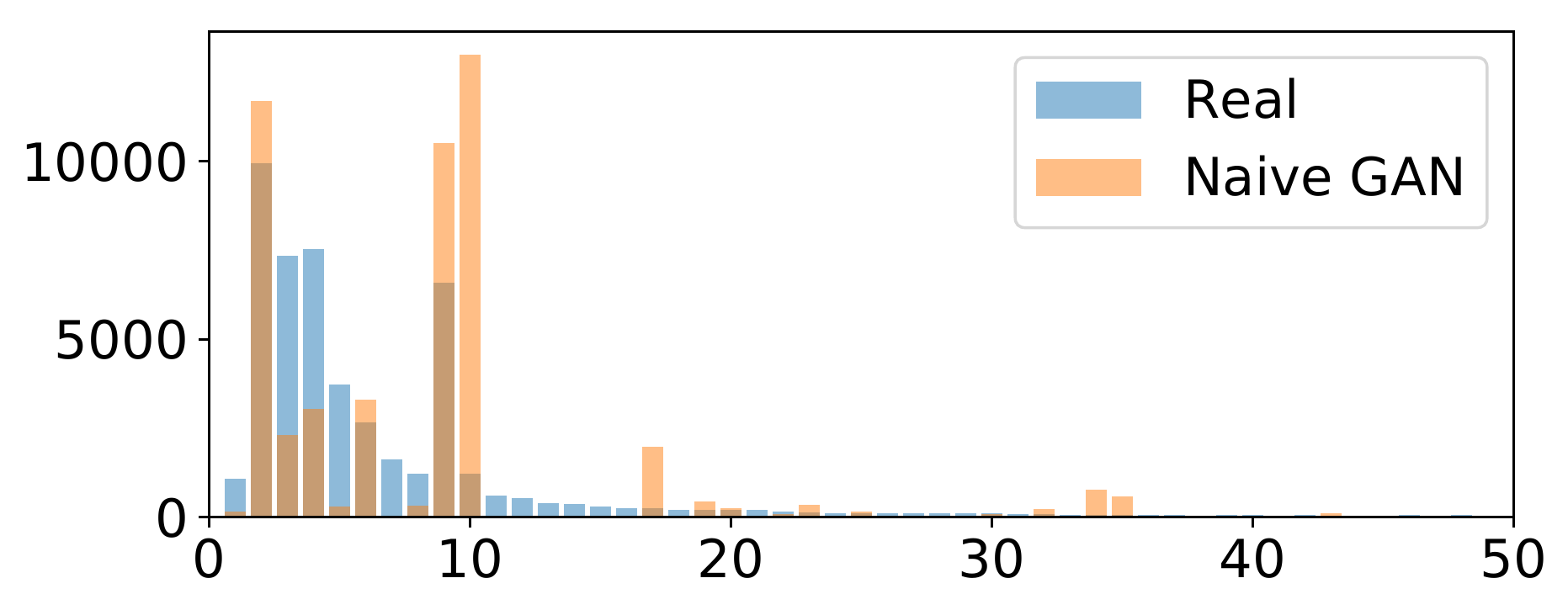}
		\put(-130,-5){Task duration (seconds)}
	\end{minipage}
	\caption{Histogram of task duration for the \clustershort~dataset. DoppelGANger gives the best fidelity.}
	\label{fig:google-length-hist-complete}
\end{figure}

\myparatightest{Temporal lengths}
Figure \ref{fig:google-length-hist-complete} shows the length distribution of \name~ and baselines in \clustershort~dataset. It is clear that \name~ has the best fidelity.

\begin{figure}[t]
	\centering
	\begin{minipage}{1.0\linewidth}
		\centering
		\includegraphics[width=0.7\linewidth]{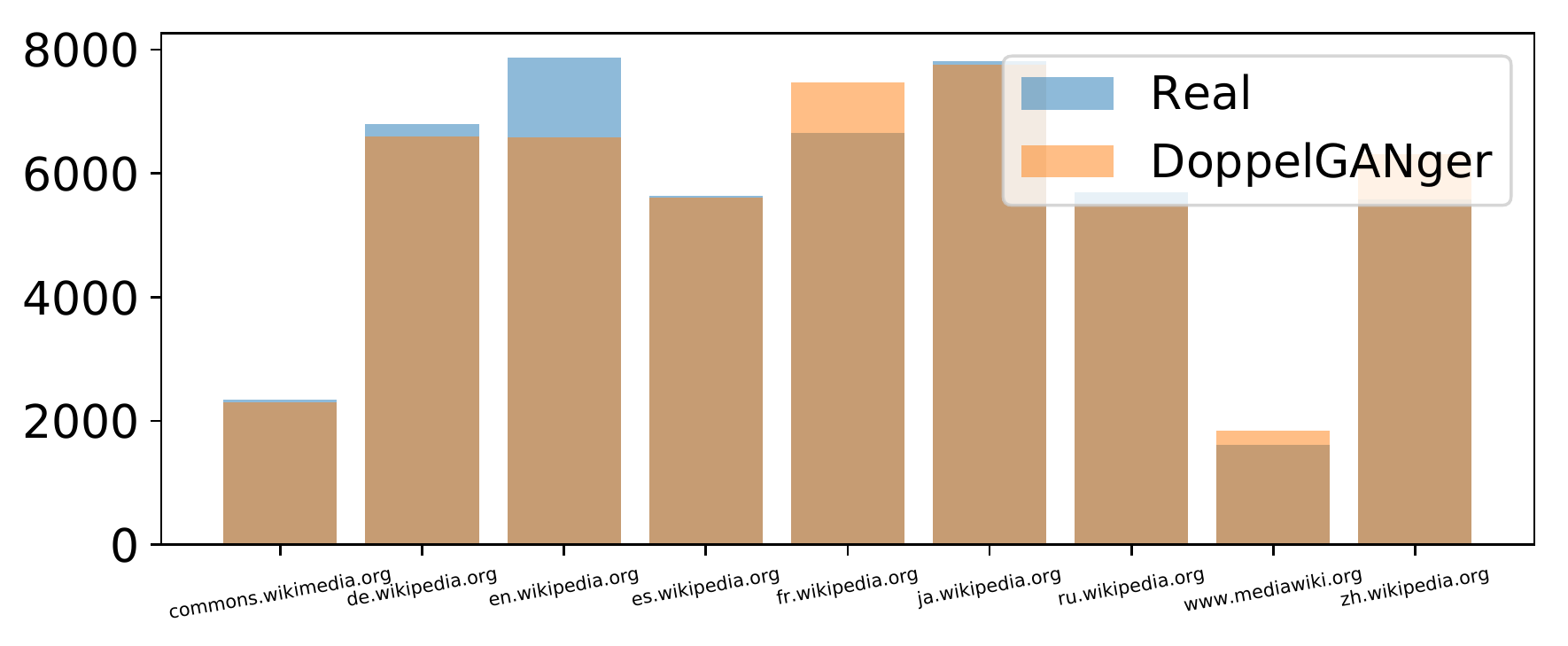}
		\includegraphics[width=0.7\linewidth]{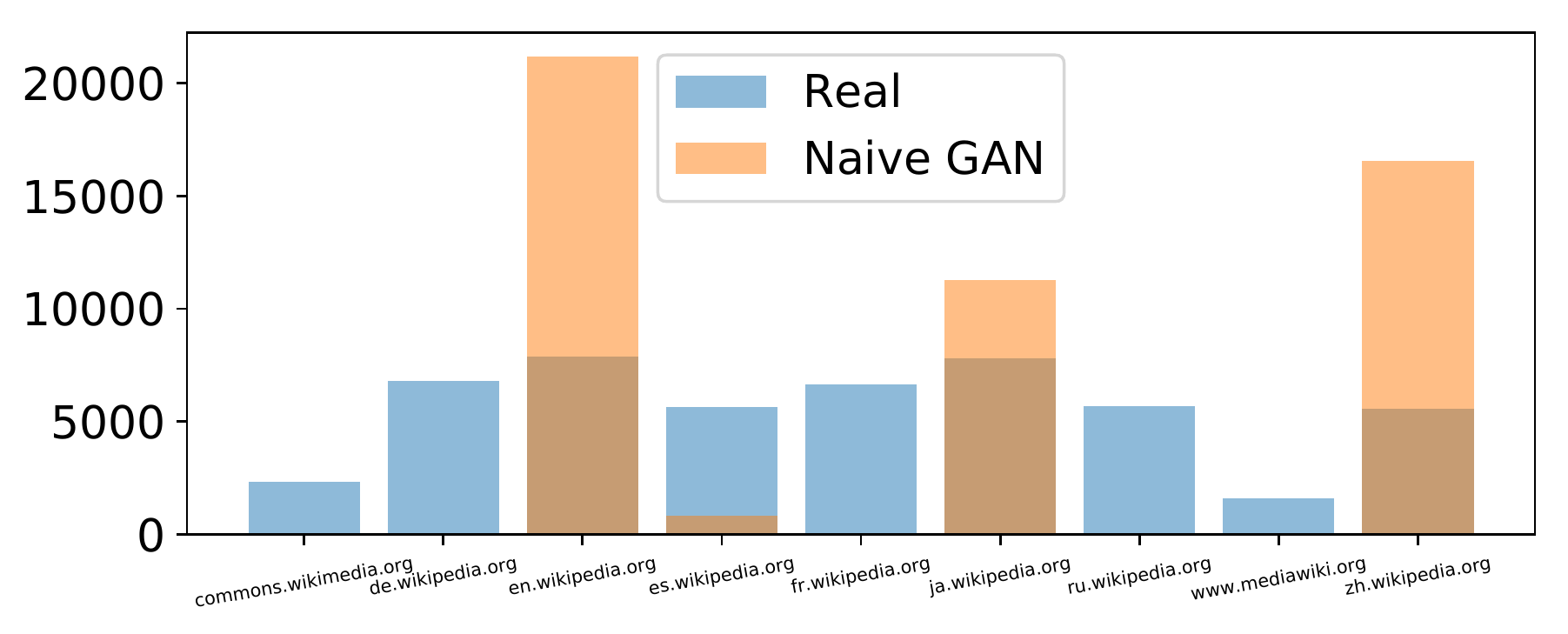}
		\put(-110,-5){Wikipedia domain}
		\put(-180,45){\rotatebox{90}{Count}}
	\end{minipage}
	\caption{Histograms of Wikipedia domain from \wikishort~dataset.}
	\label{fig:wiki-domain-hist}
\end{figure}

\myparatightest{\Metadata{} distributions}
Figure \ref{fig:wiki-domain-hist}
shows the histogram of Wikipedia domain of Naive GAN and \name{}. \name{} learns the distribution fairly well, whereas naive GAN cannot. 

\begin{figure}[t]
	\centering
	\begin{minipage}{1.0\linewidth}
		\centering
		\includegraphics[width=0.7\linewidth]{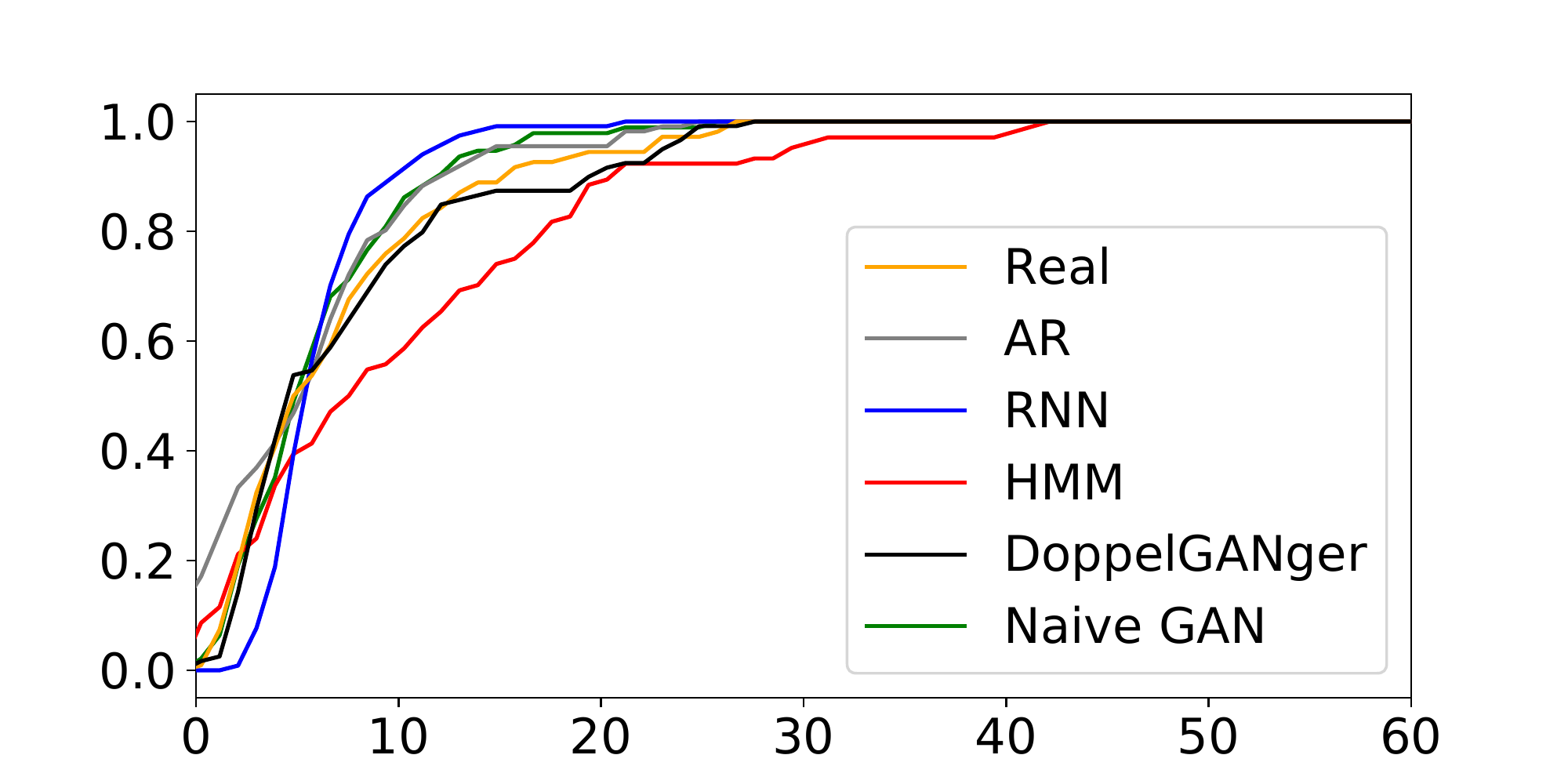}
		\put(-175,35){(a)}
	\end{minipage}
	\begin{minipage}{1.0\linewidth}
		\centering
		\includegraphics[width=0.7\linewidth]{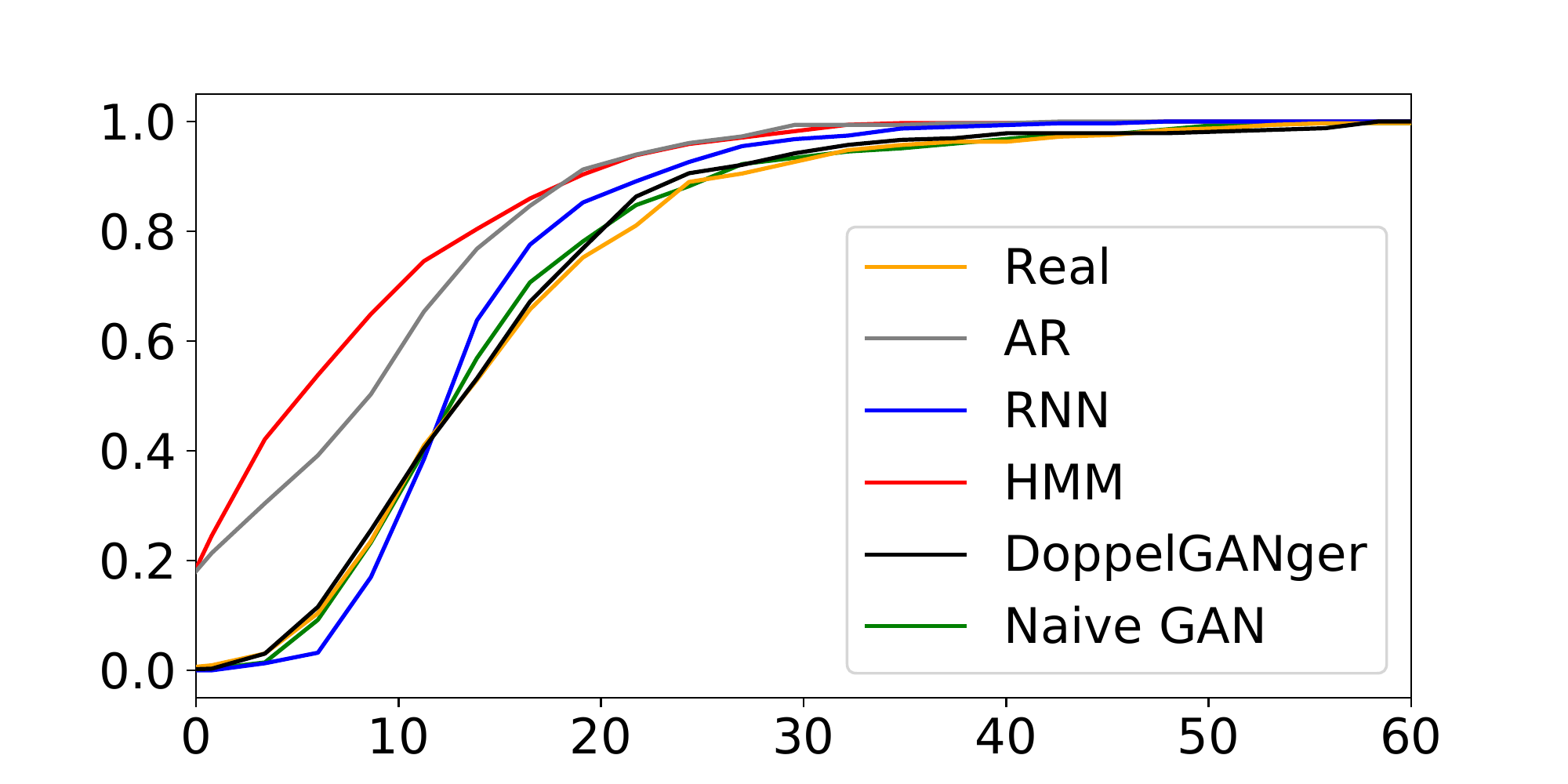}
		\put(-120,-5){Bandwidth (GB)}
		\put(-190,65){\rotatebox{90}{CDF}}
		\put(-175,35){(b)}
	\end{minipage}
	\vspace{-0.3cm}
	\caption{Total bandwidth usage in 2 weeks in \fccshort~dataset for (a) DSL and (b) cable users.}
	\label{fig:fcc-cdf-DSL}
	\label{fig:fcc-cdf-cable}
\end{figure}

\myparatightest{\Measurement{}-\metadata{} correlations}
We compute the CDF of total bandwidth for DSL and cable users in \fccshort~dataset. 
Figures \ref{fig:fcc-cdf-DSL}(a) and \ref{fig:fcc-cdf-cable}(b) plot the full CDFs. 
Most of the baselines capture the fact that cable users consume more bandwidth than DSL users. 
However, \name~appears to excel in  regions of the distribution with less data, e.g., very small bandwidth levels.
In both cases, \name~captures the bandwidth distribution better than the other baselines. 
This means that \name~has a high fidelity on \measurement~distribution, and also successfully capture how \metadatas~(i.e., DSL and cable) influence the \measurements.

\begin{figure}
	\centering
	\begin{minipage}{1.0\linewidth}
		\centering
		\includegraphics[width=0.22\linewidth]{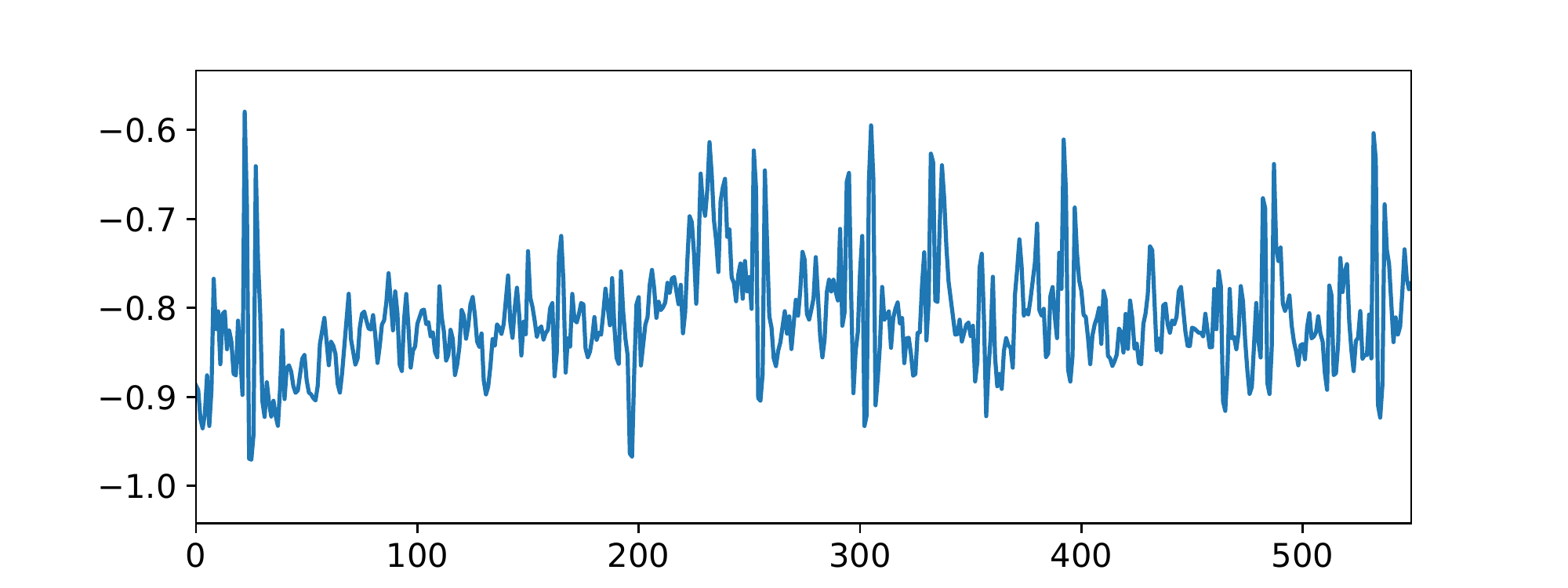}
		\includegraphics[width=0.22\linewidth]{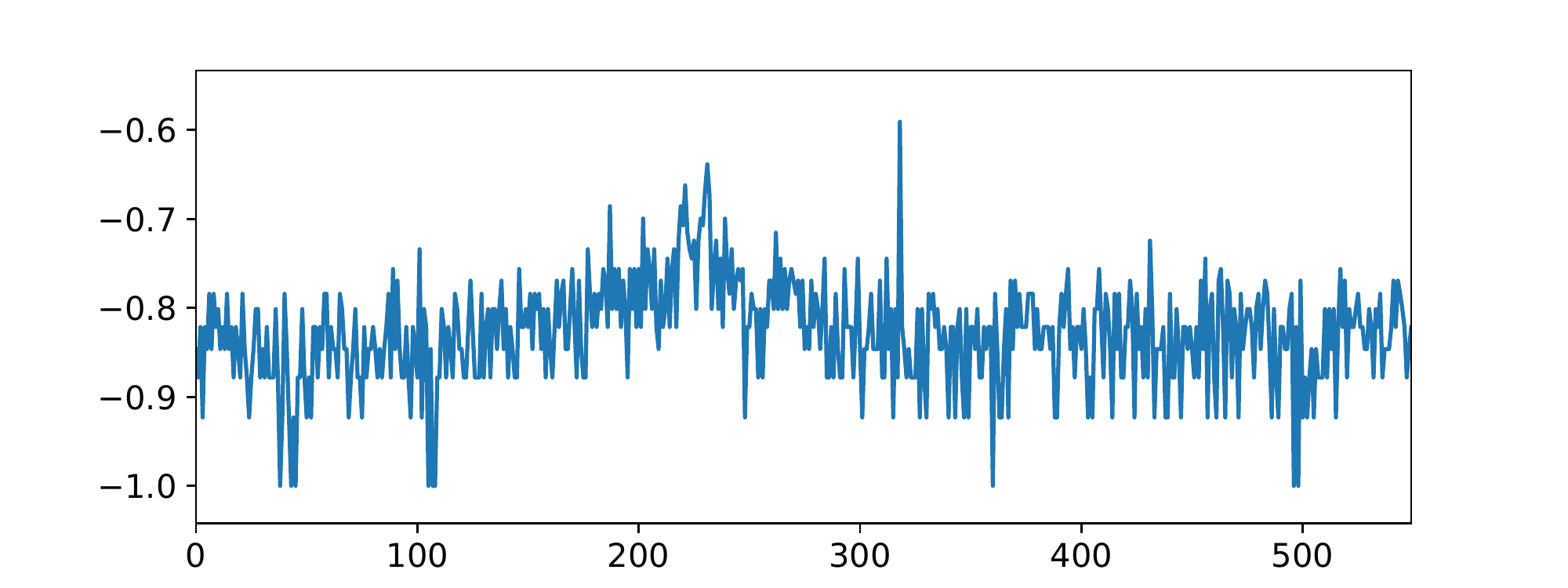}
		\includegraphics[width=0.22\linewidth]{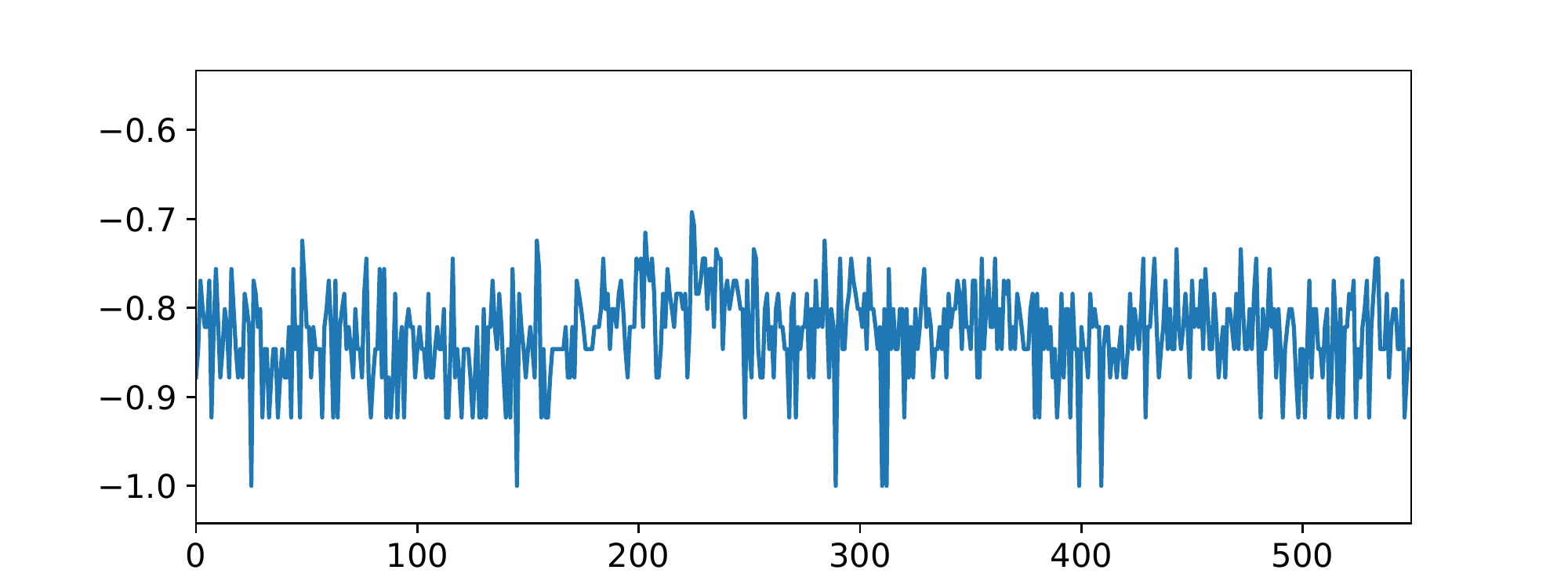}
		\includegraphics[width=0.22\linewidth]{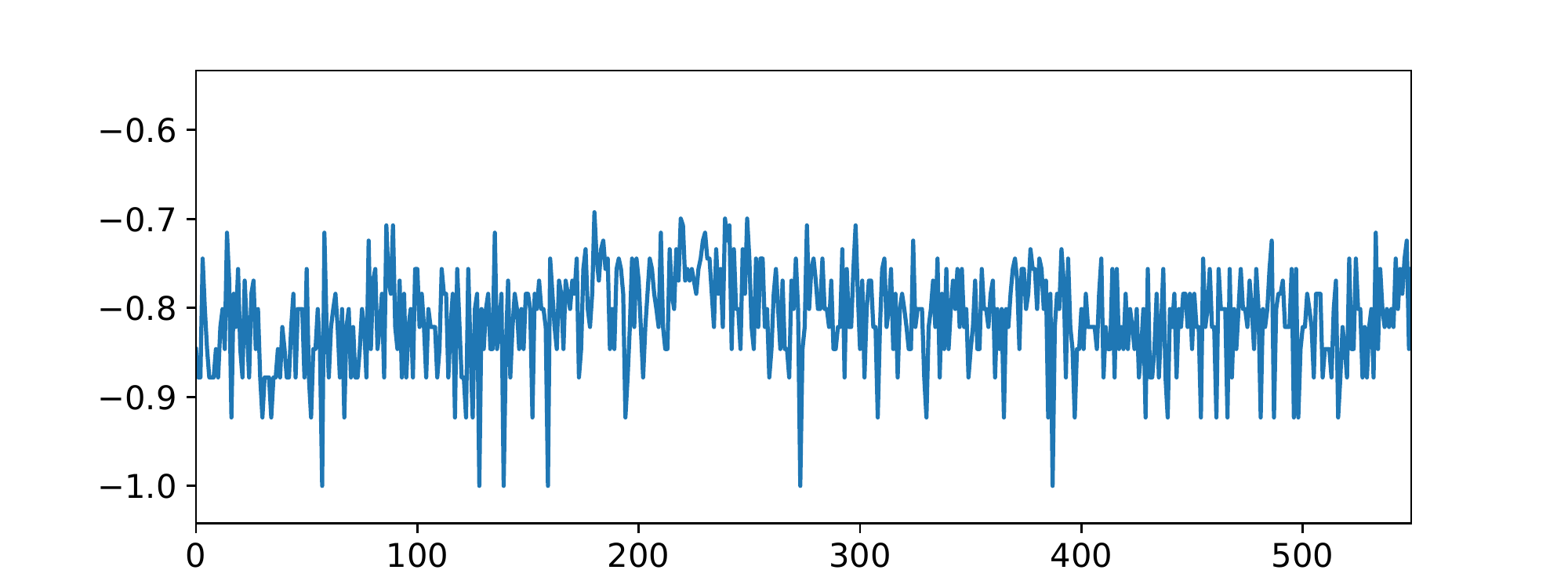}
		\put(-230,10){(a)}
		\put(-230, 25){Generated sample}
		\put(-157, 25){1st nearest}
		\put(-102, 25){2nd nearest}
		\put(-45, 25){3rd nearest}
	\end{minipage}
	\begin{minipage}{1.0\linewidth}
		\centering
		\includegraphics[width=0.22\linewidth]{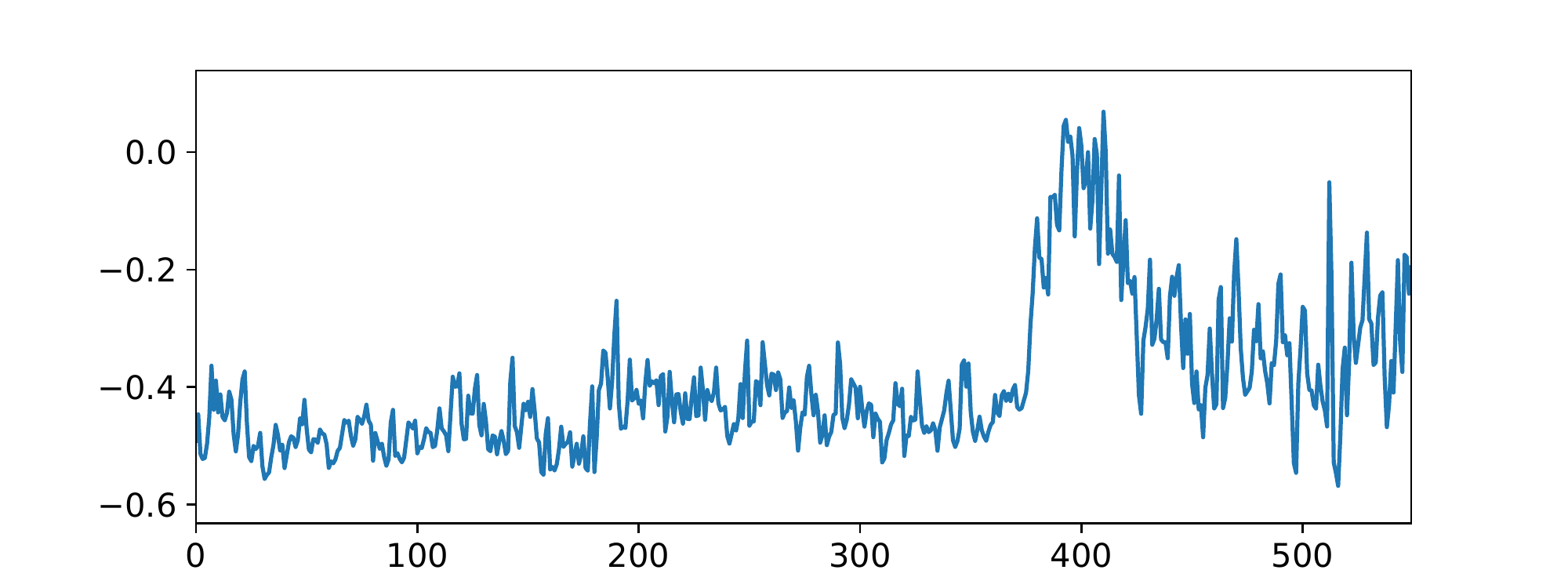}
		\includegraphics[width=0.22\linewidth]{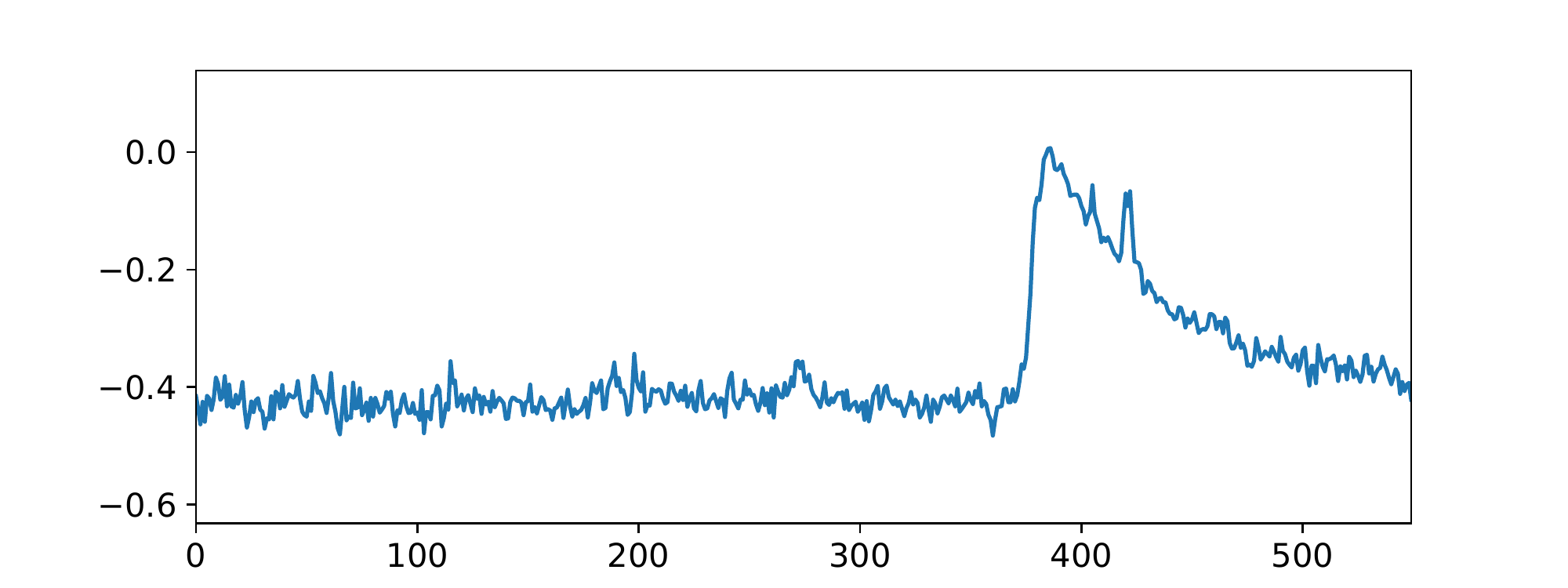}
		\includegraphics[width=0.22\linewidth]{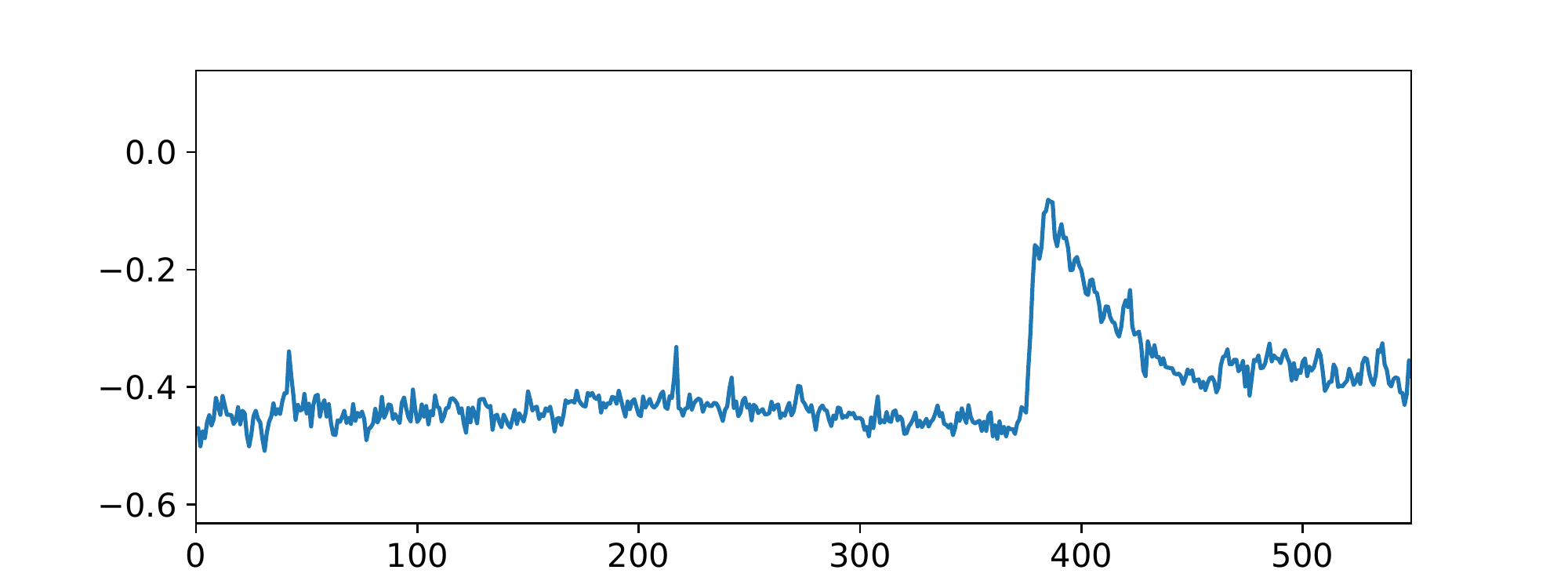}
		\includegraphics[width=0.22\linewidth]{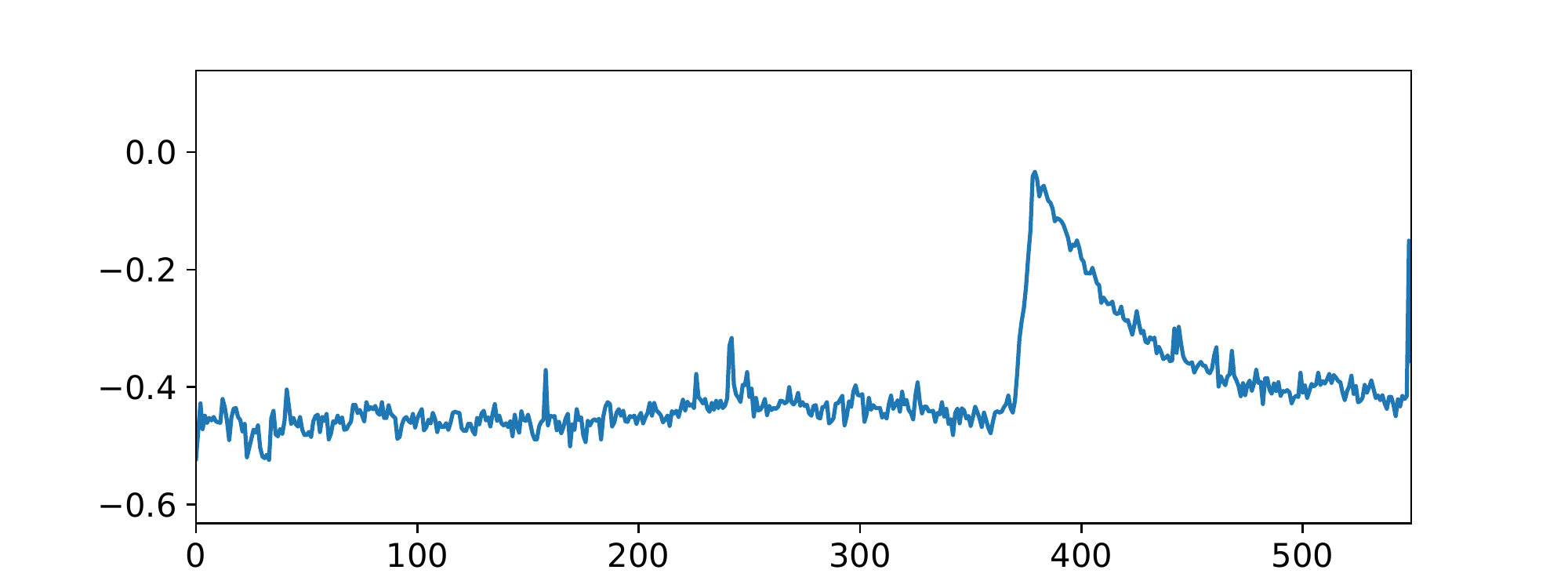}
		\put(-230,10){(b)}
	\end{minipage}
	\begin{minipage}{1.0\linewidth}
		\centering
		\includegraphics[width=0.22\linewidth]{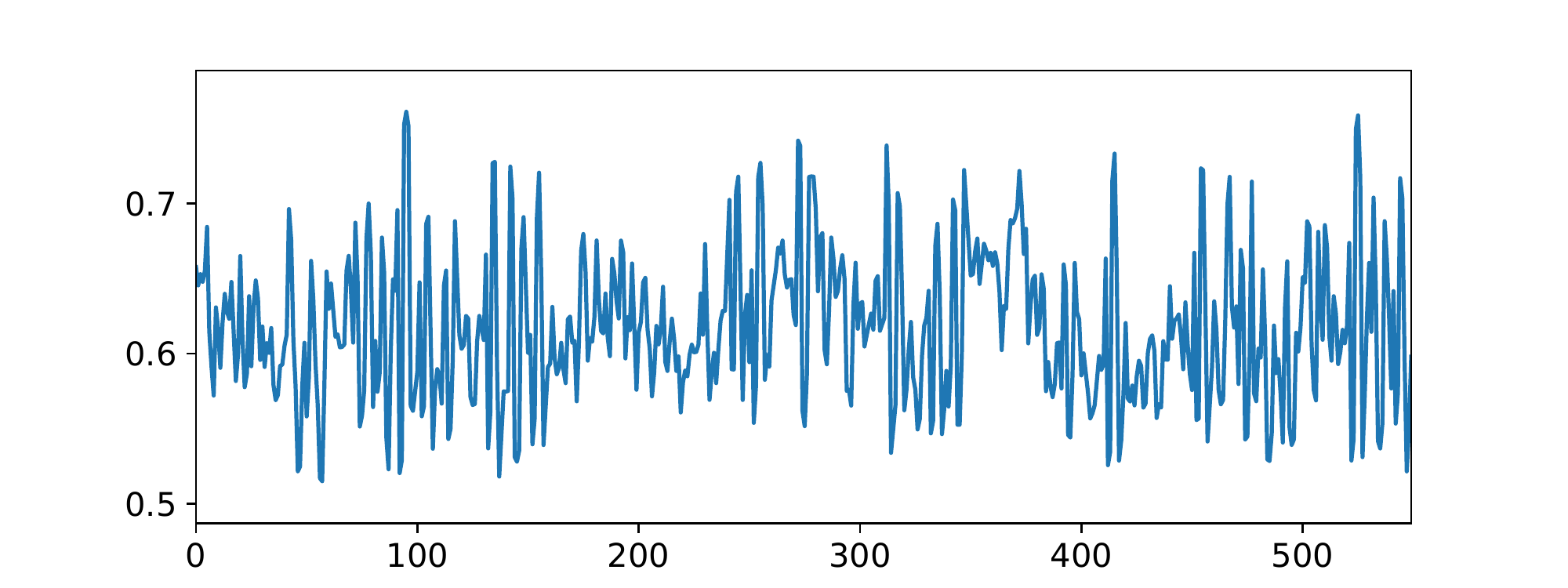}
		\includegraphics[width=0.22\linewidth]{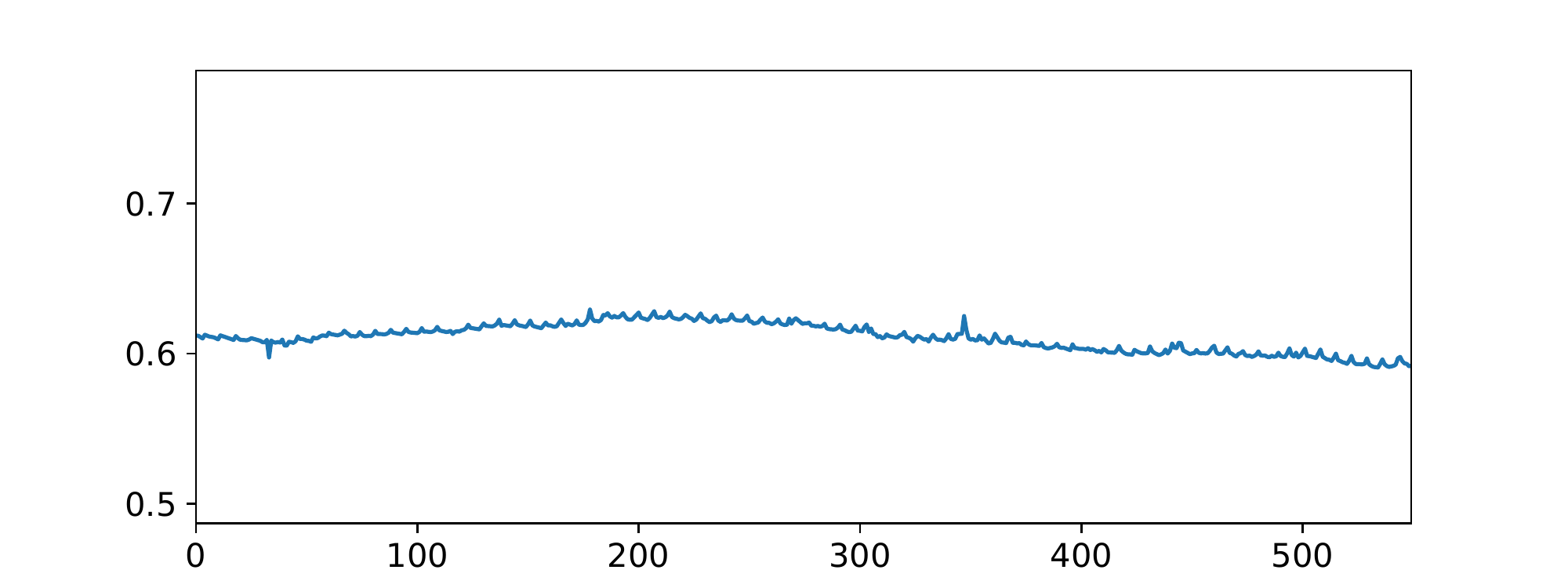}
		\includegraphics[width=0.22\linewidth]{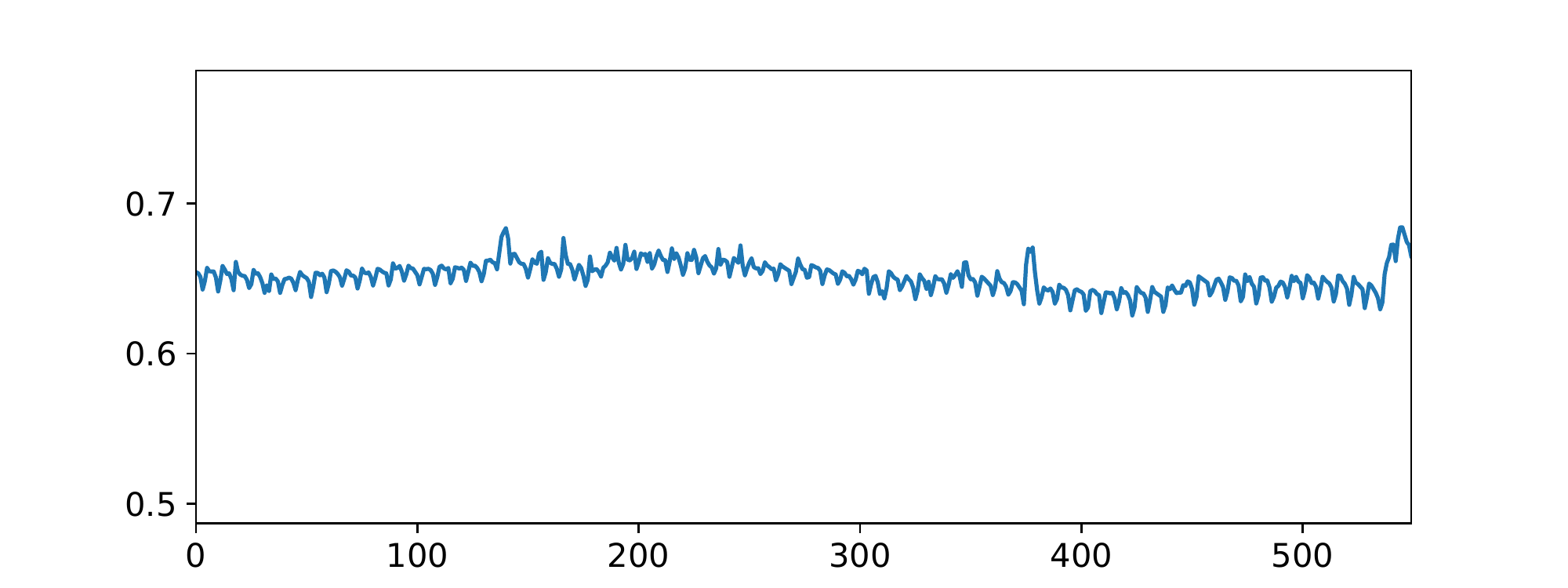}
		\includegraphics[width=0.22\linewidth]{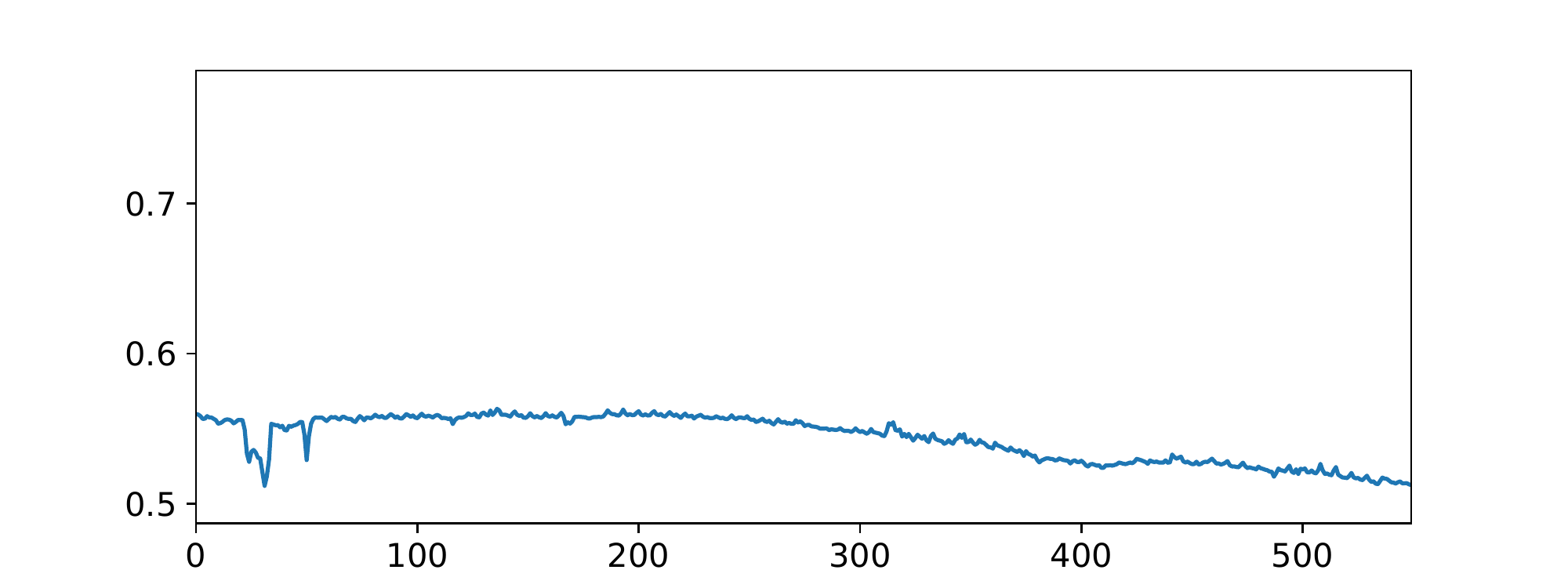}
		\put(-230,10){(c)}
	\end{minipage}

	\caption{Three time series samples selected uniformly at random from the synthetic dataset generated using \name~and the corresponding top-3 nearest neighbours (based on square error) from the real \wikishort~dataset. The time series shown here is daily page views (normalized).}
	\label{fig:web-memorization}
\end{figure}

\begin{figure}
	\centering
	\begin{minipage}{1.0\linewidth}
		\centering
		\includegraphics[width=0.22\linewidth]{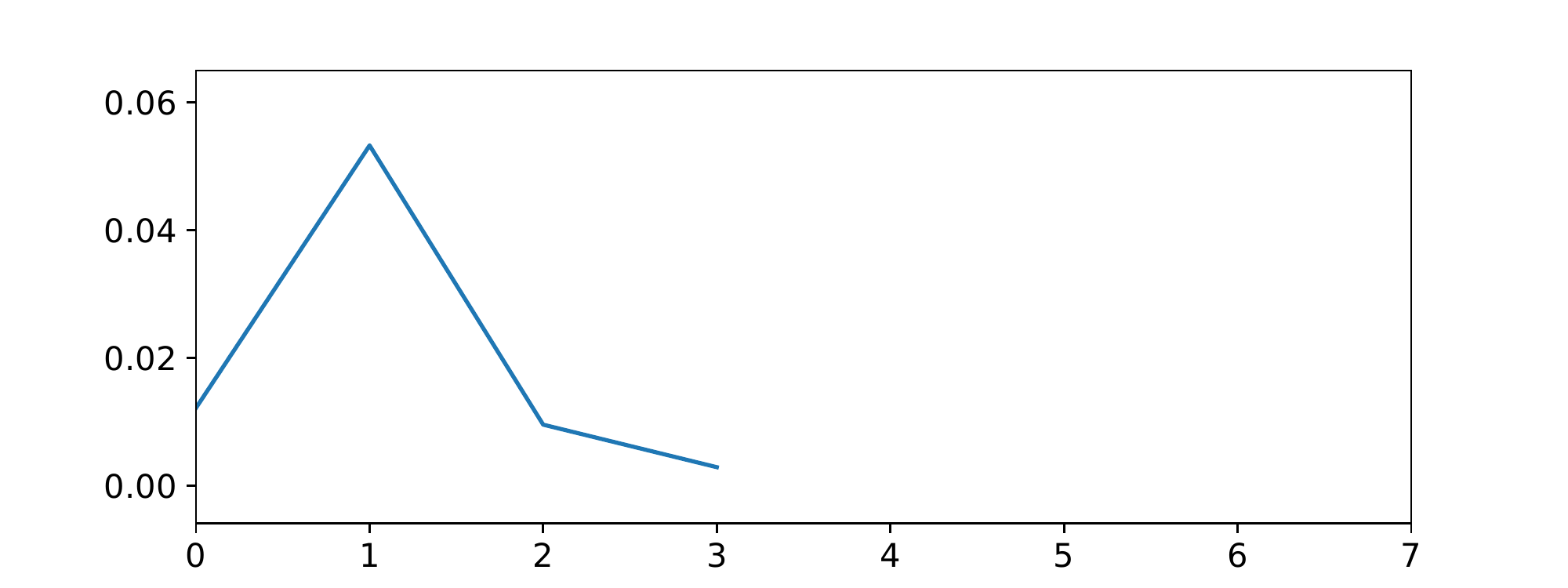}
		\includegraphics[width=0.22\linewidth]{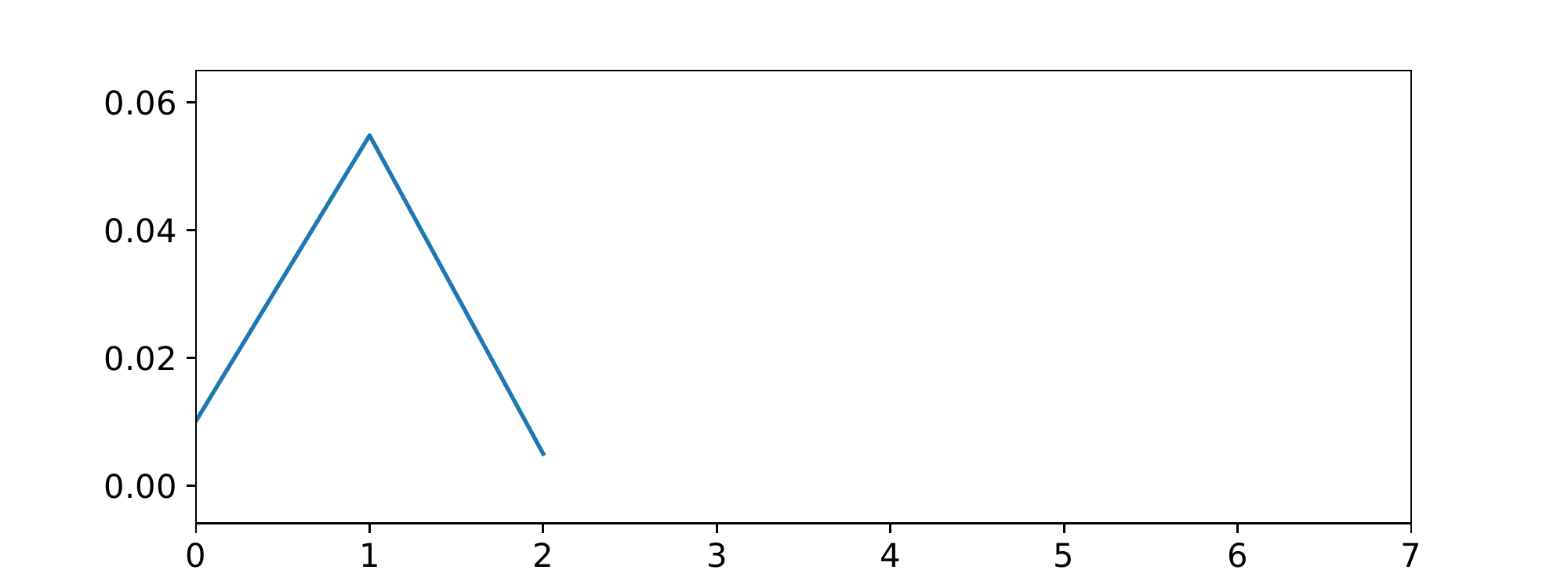}
		\includegraphics[width=0.22\linewidth]{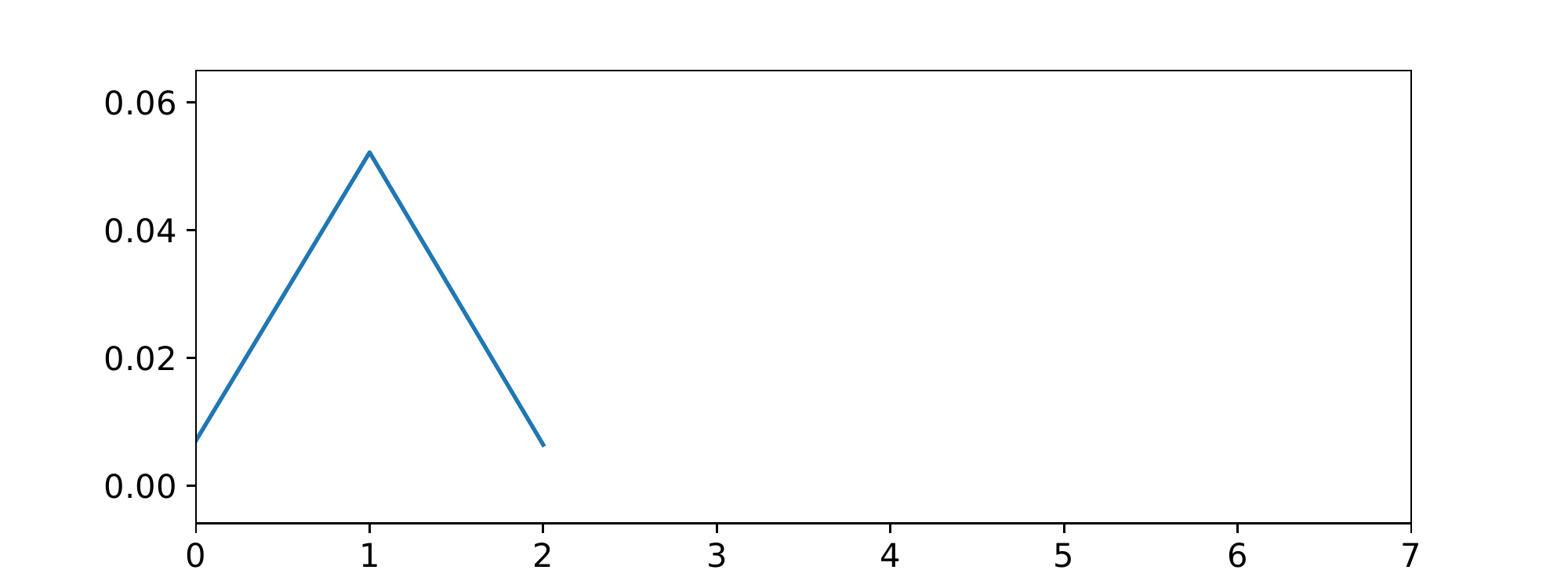}
		\includegraphics[width=0.22\linewidth]{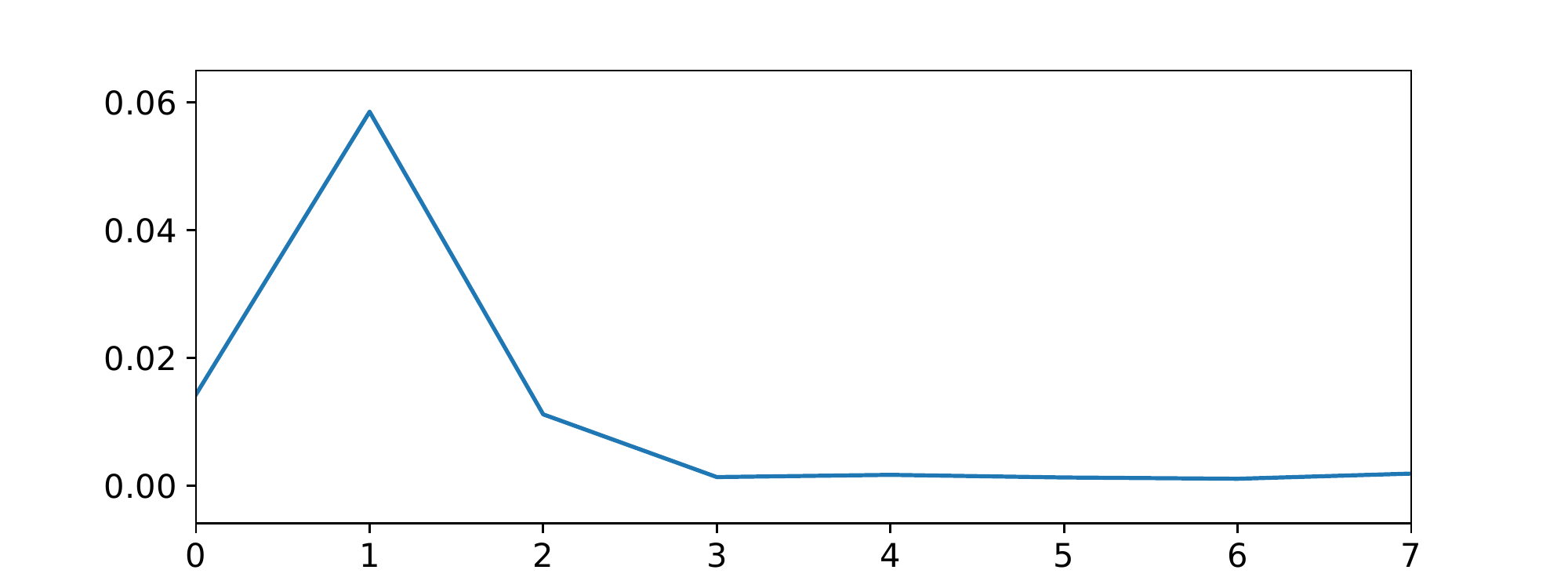}
		\put(-230,10){(a)}
		\put(-230, 25){Generated sample}
		\put(-157, 25){1st nearest}
		\put(-102, 25){2nd nearest}
		\put(-45, 25){3rd nearest}
	\end{minipage}
	\begin{minipage}{1.0\linewidth}
		\centering
		\includegraphics[width=0.22\linewidth]{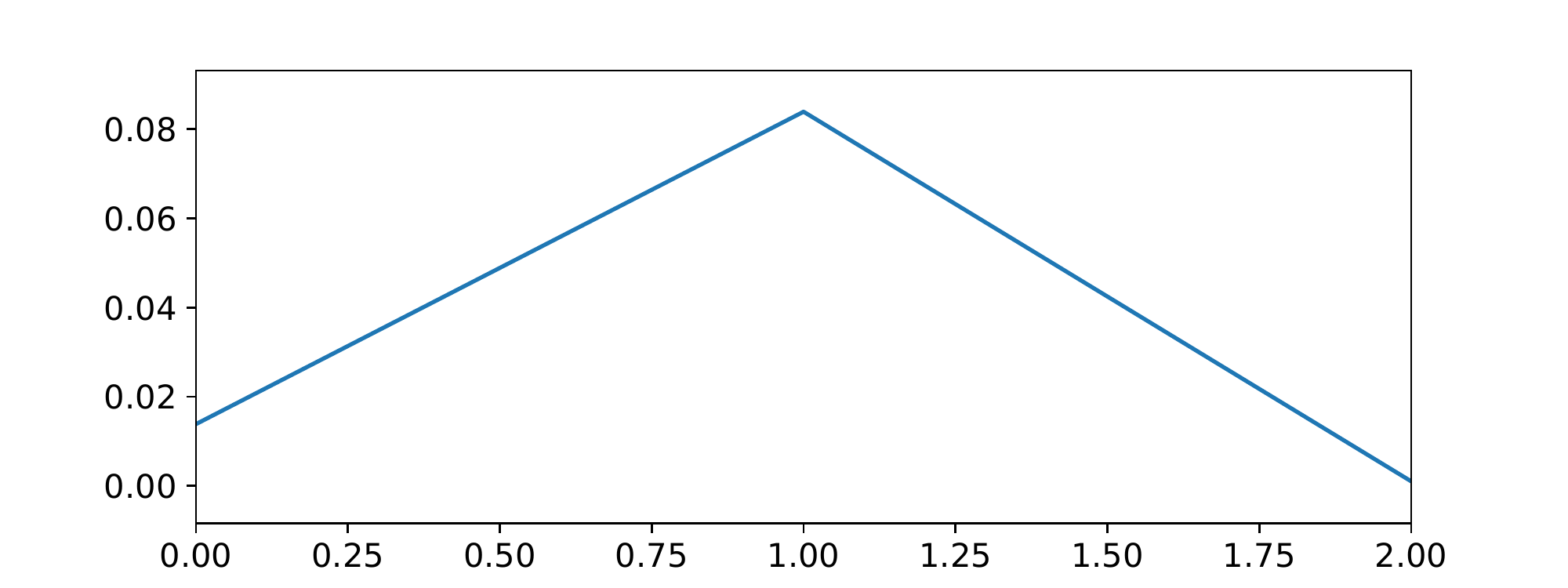}
		\includegraphics[width=0.22\linewidth]{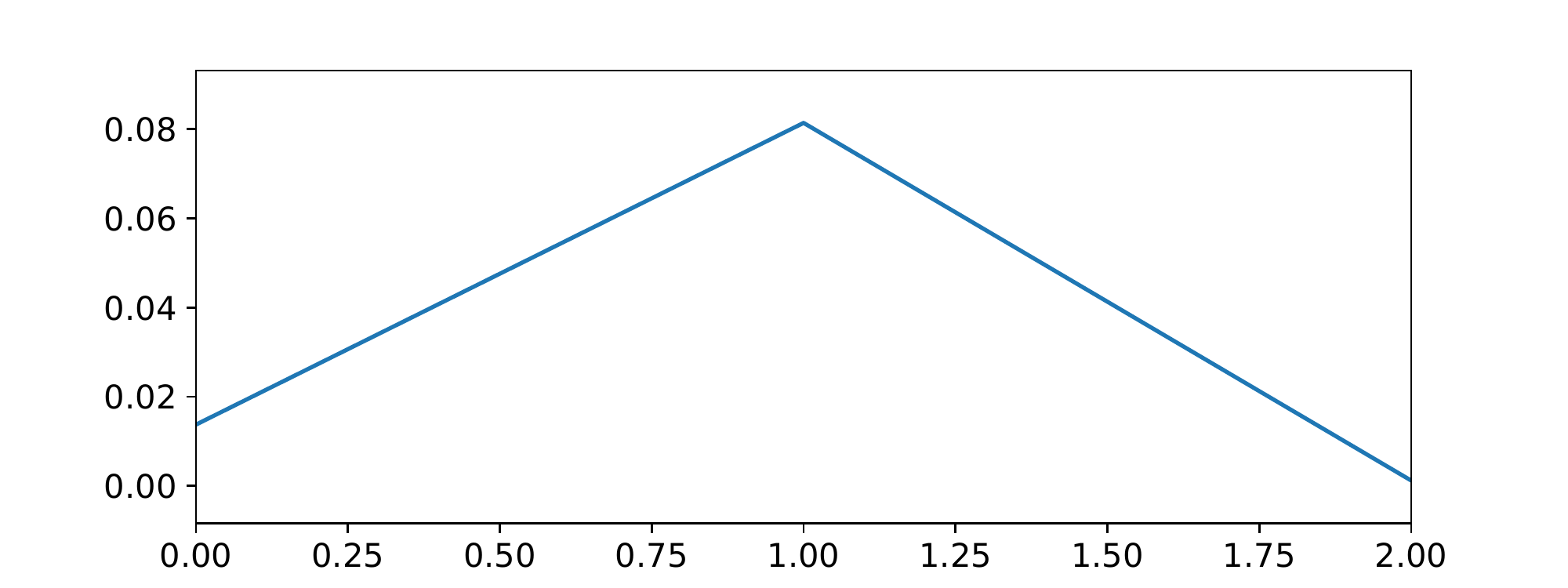}
		\includegraphics[width=0.22\linewidth]{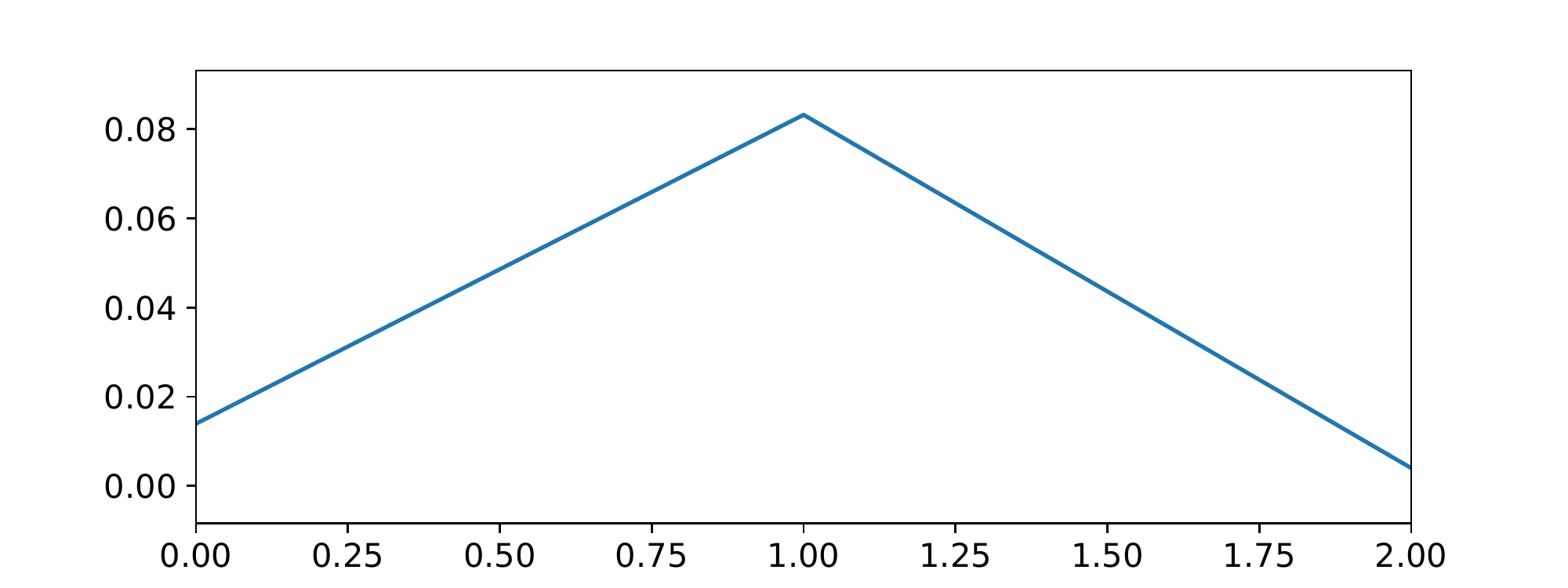}
		\includegraphics[width=0.22\linewidth]{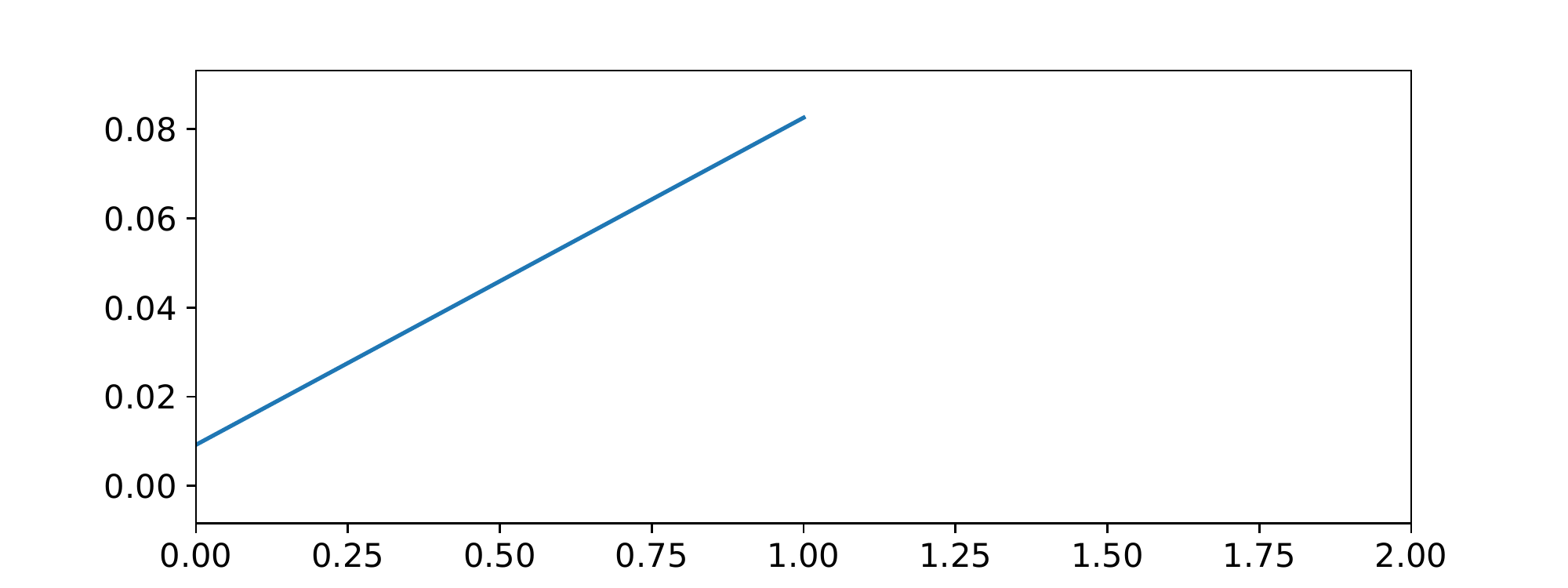}
		\put(-230,10){(b)}
	\end{minipage}
	\begin{minipage}{1.0\linewidth}
		\centering
		\includegraphics[width=0.22\linewidth]{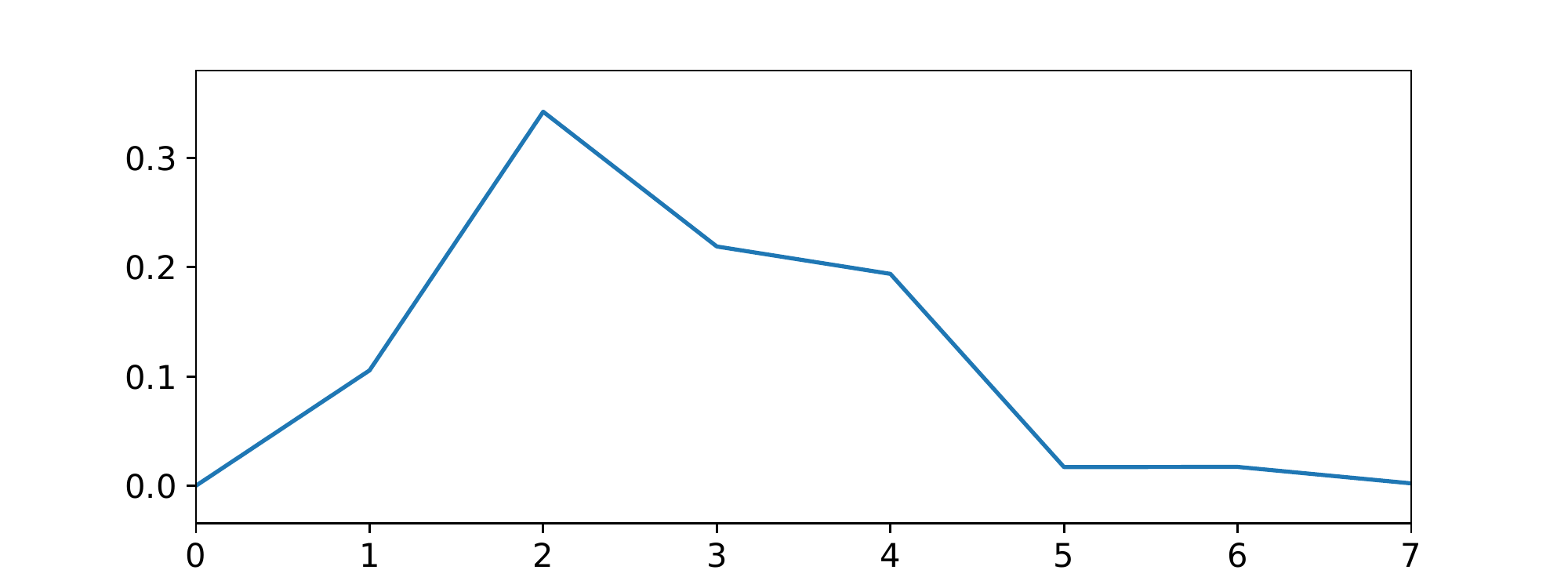}
		\includegraphics[width=0.22\linewidth]{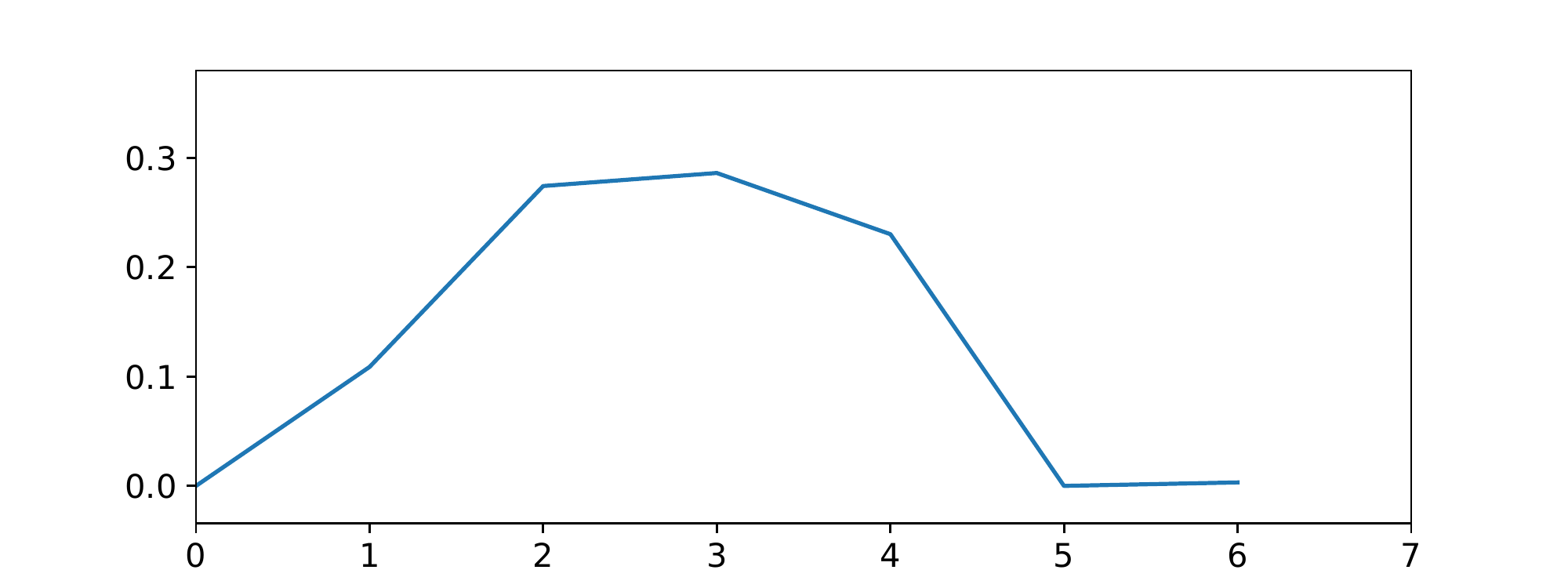}
		\includegraphics[width=0.22\linewidth]{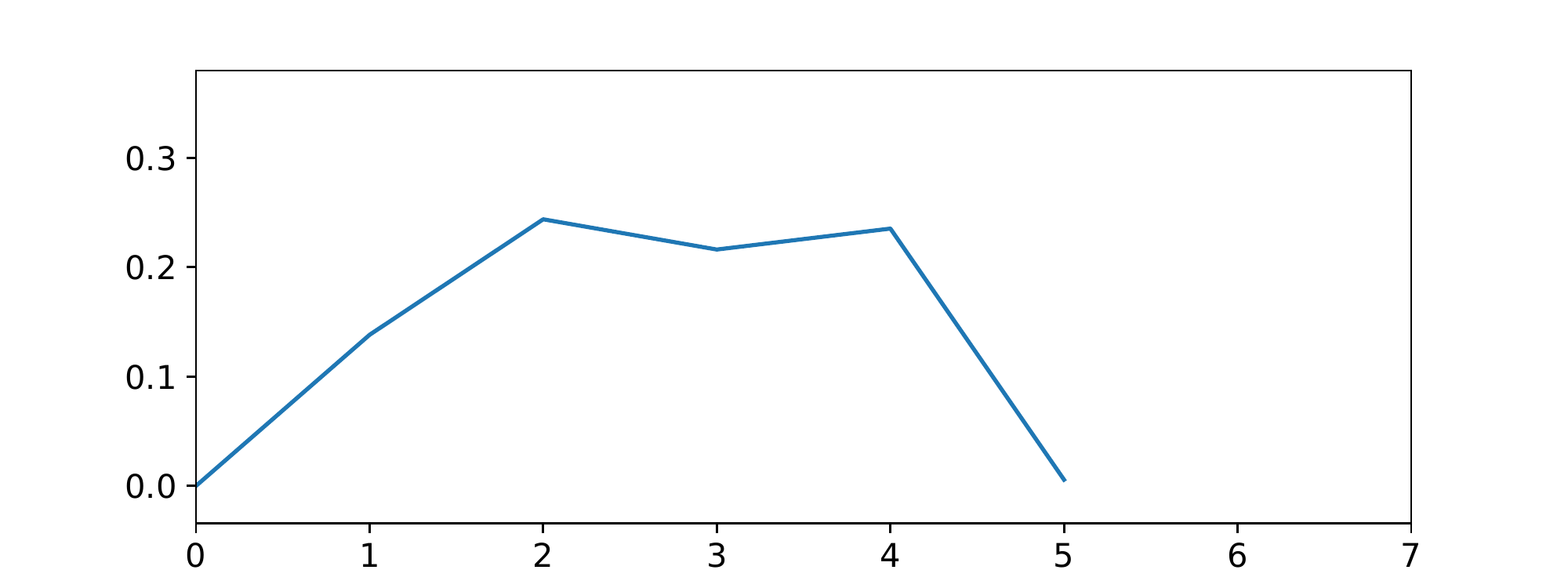}
		\includegraphics[width=0.22\linewidth]{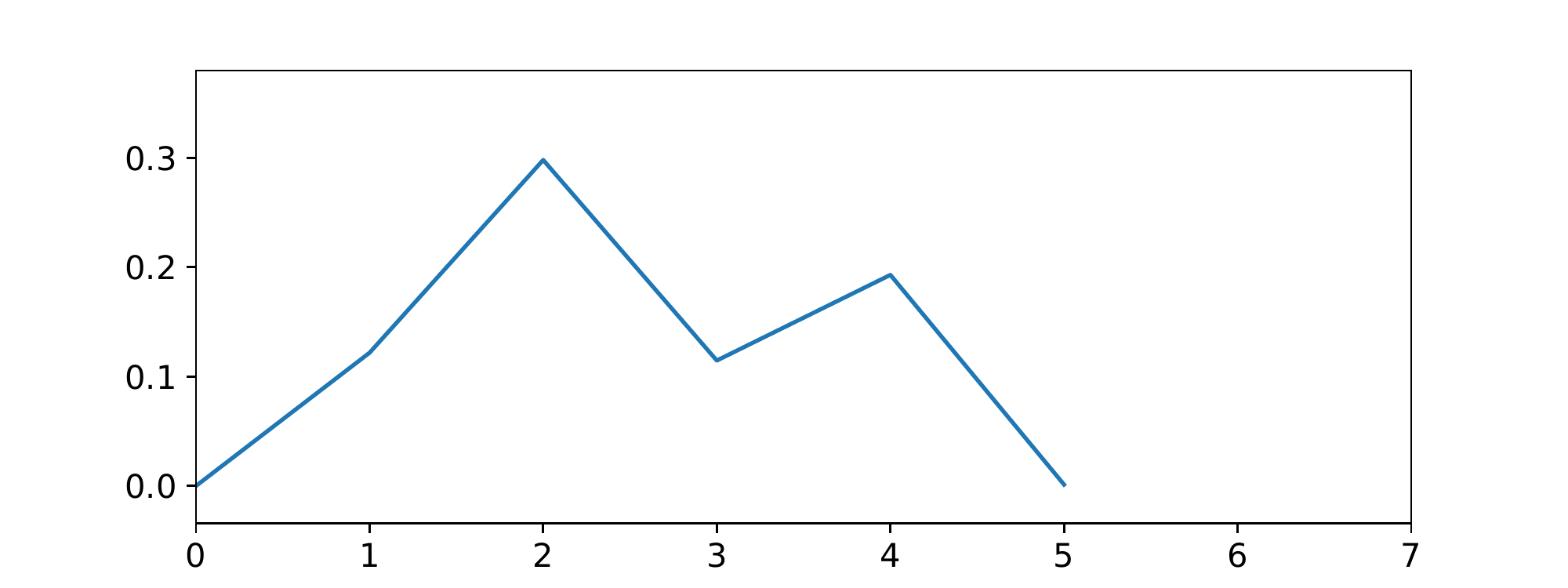}
		\put(-230,10){(c)}
	\end{minipage}
	
	\caption{Three time series samples selected uniformly at random from the synthetic dataset generated using \name~and the corresponding top-3 nearest neighbours (based on square error) from the real \clustershort~dataset. The time series shown here is CPU rate (normalized).}
	\label{fig:google-memorization}
\end{figure}

\begin{figure}
	\centering
	\begin{minipage}{1.0\linewidth}
		\centering
		\includegraphics[width=0.22\linewidth]{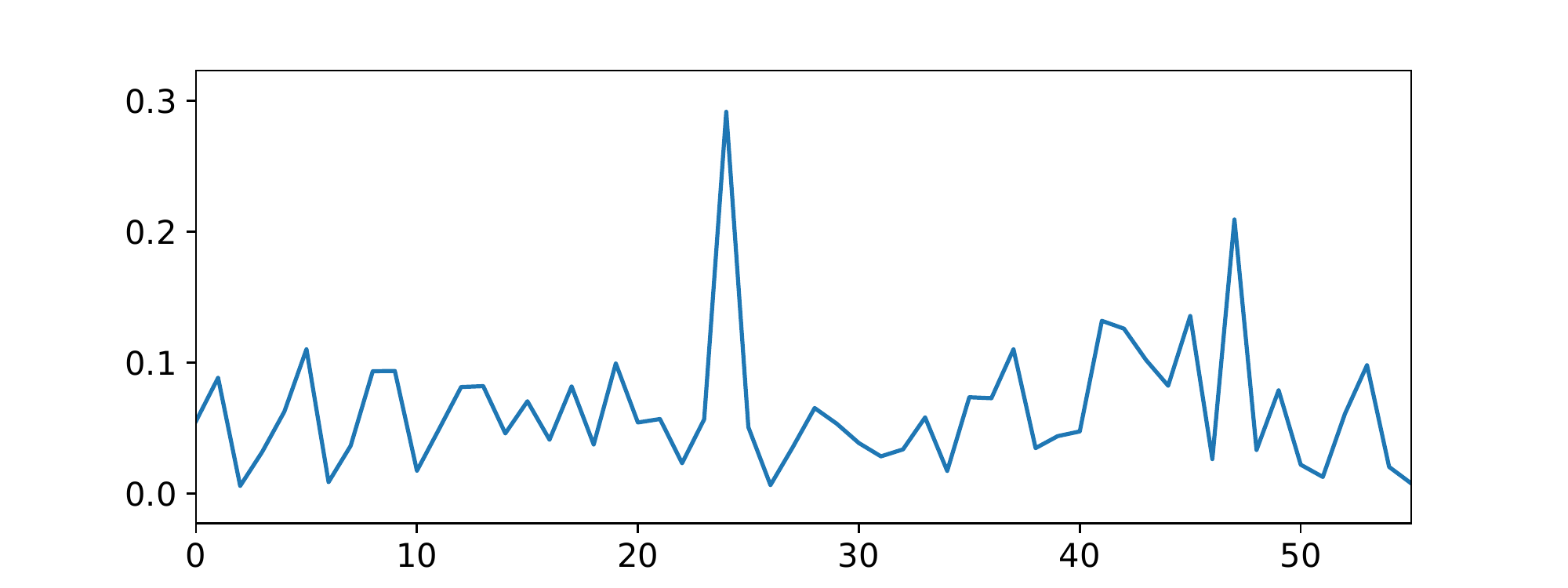}
		\includegraphics[width=0.22\linewidth]{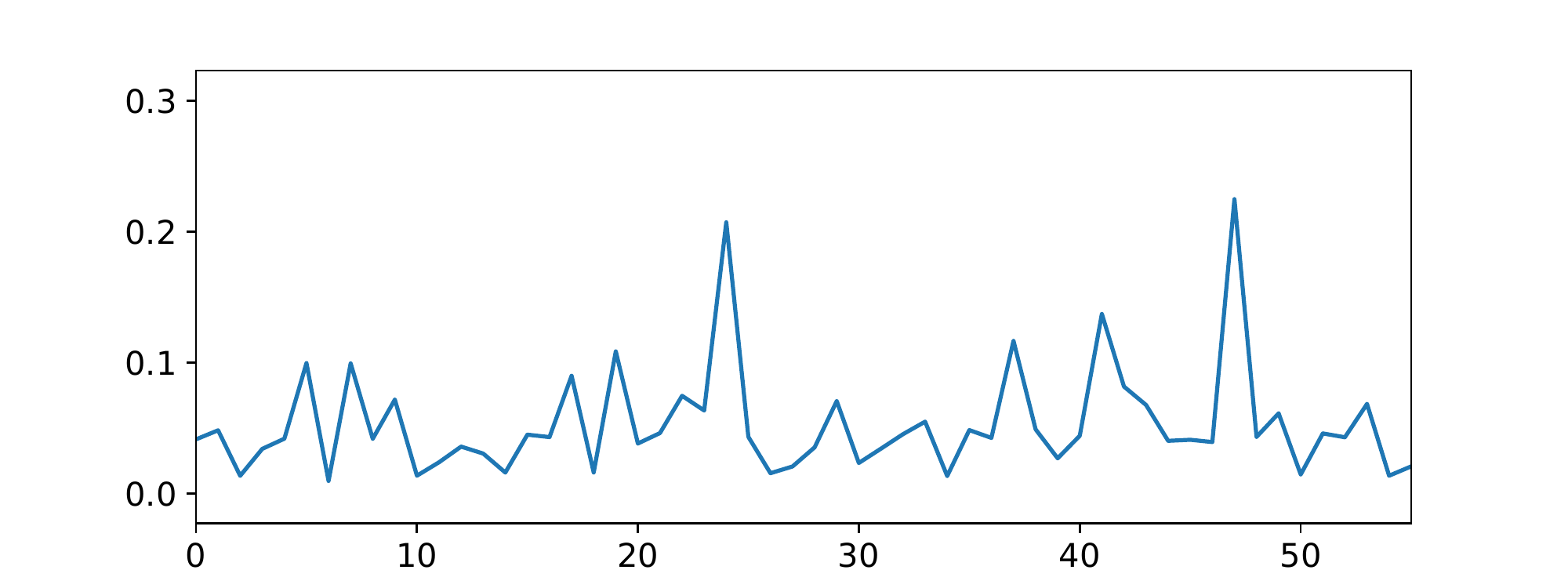}
		\includegraphics[width=0.22\linewidth]{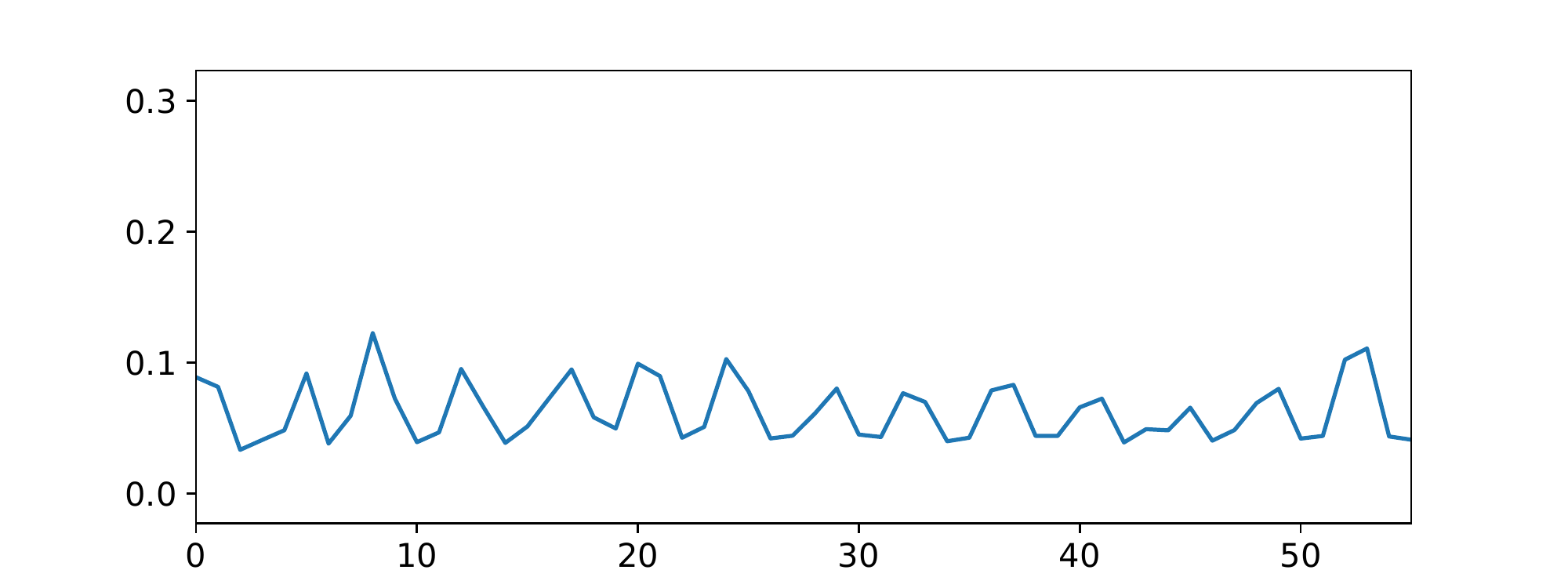}
		\includegraphics[width=0.22\linewidth]{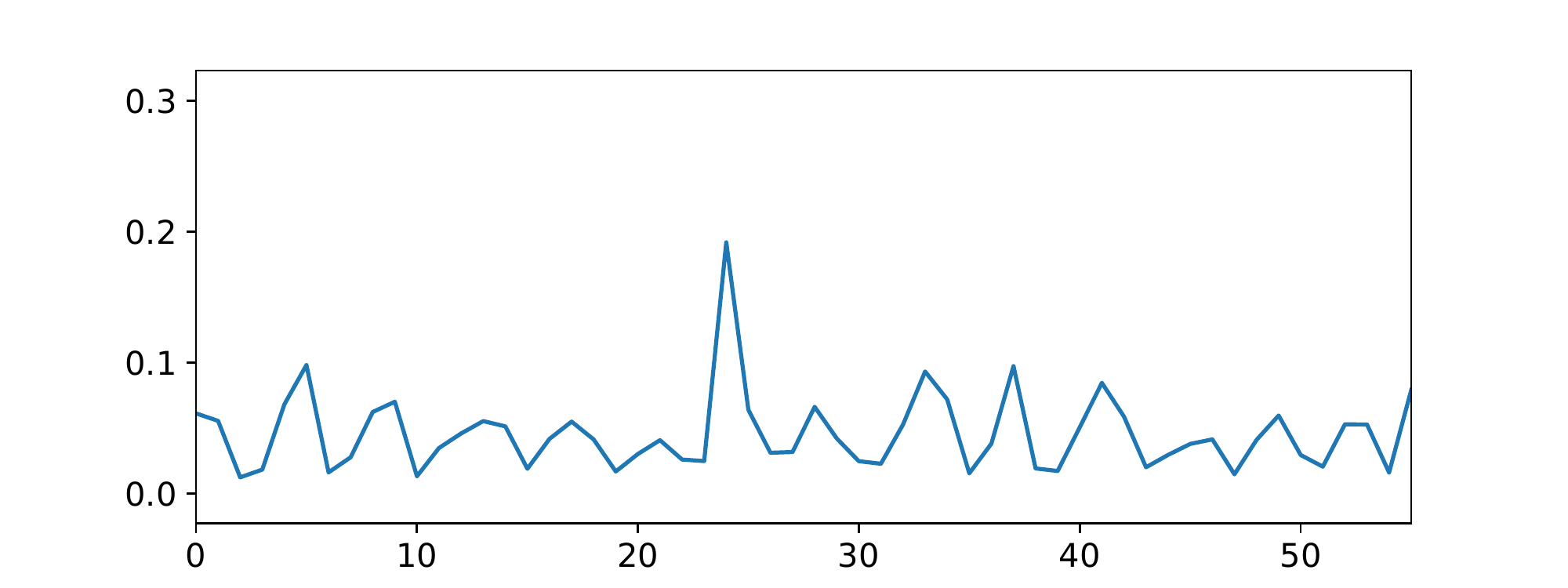}
		\put(-230,10){(a)}
		\put(-230, 25){Generated sample}
		\put(-157, 25){1st nearest}
		\put(-102, 25){2nd nearest}
		\put(-45, 25){3rd nearest}
	\end{minipage}
	\begin{minipage}{1.0\linewidth}
		\centering
		\includegraphics[width=0.22\linewidth]{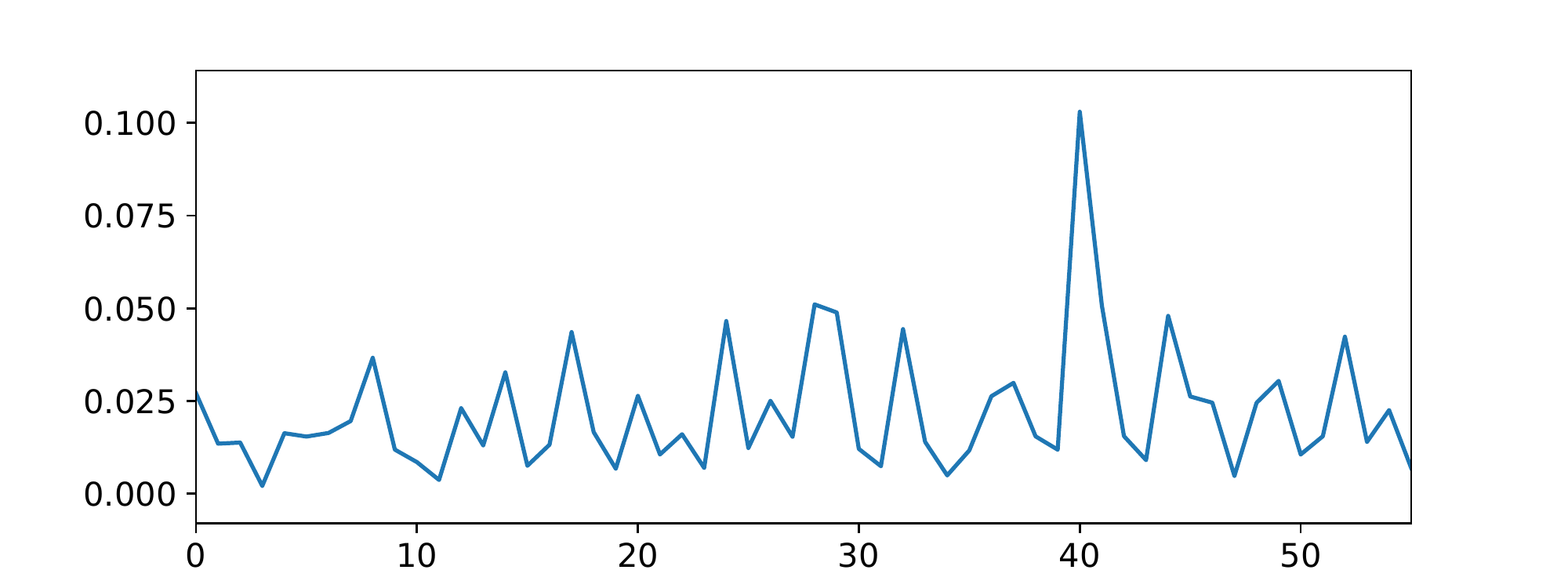}
		\includegraphics[width=0.22\linewidth]{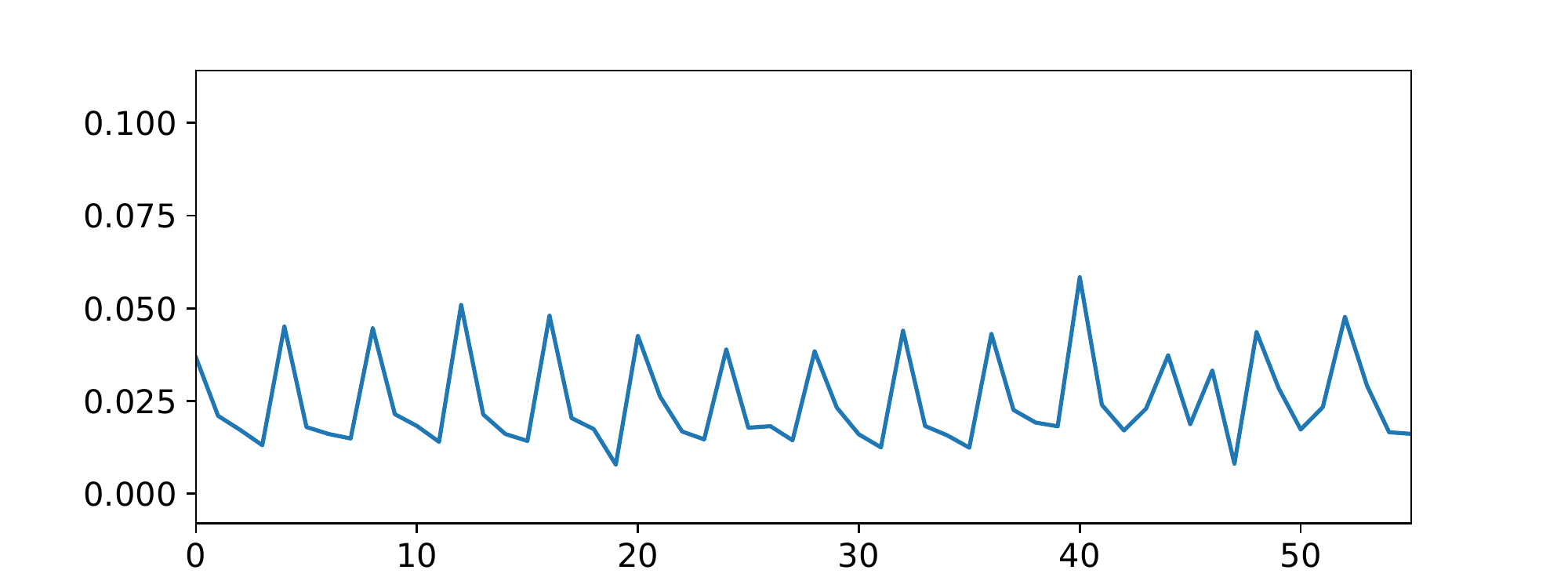}
		\includegraphics[width=0.22\linewidth]{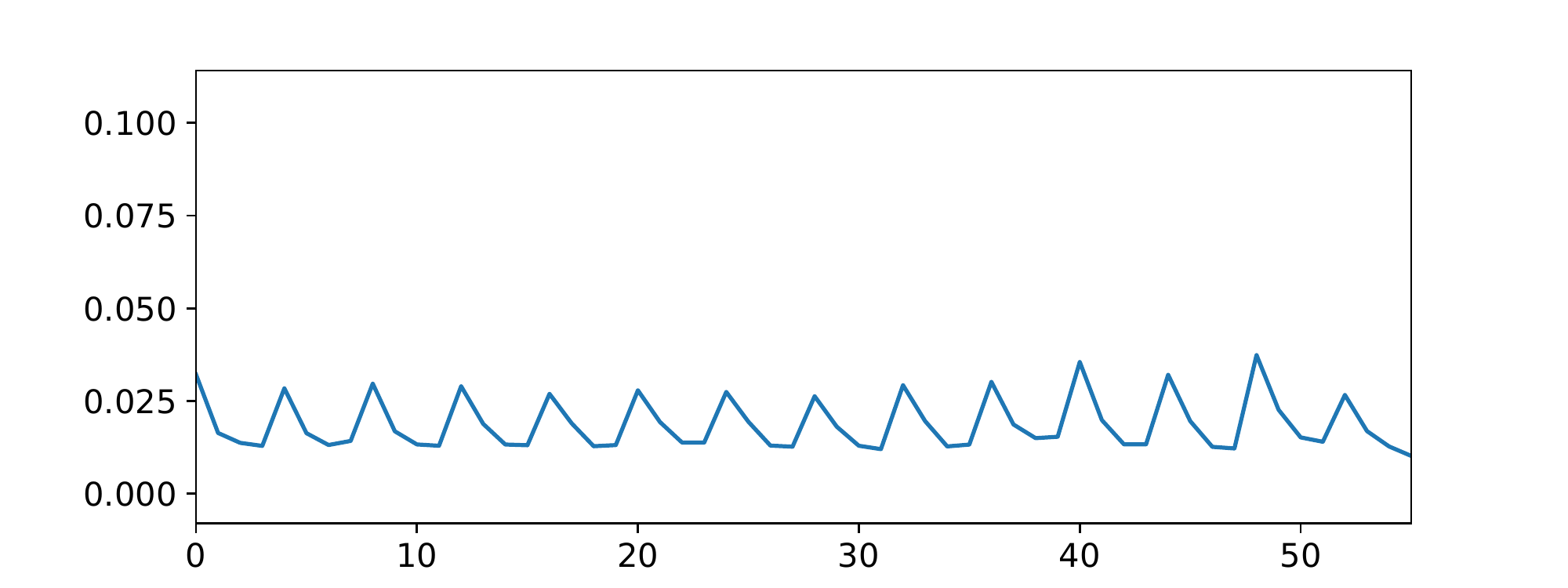}
		\includegraphics[width=0.22\linewidth]{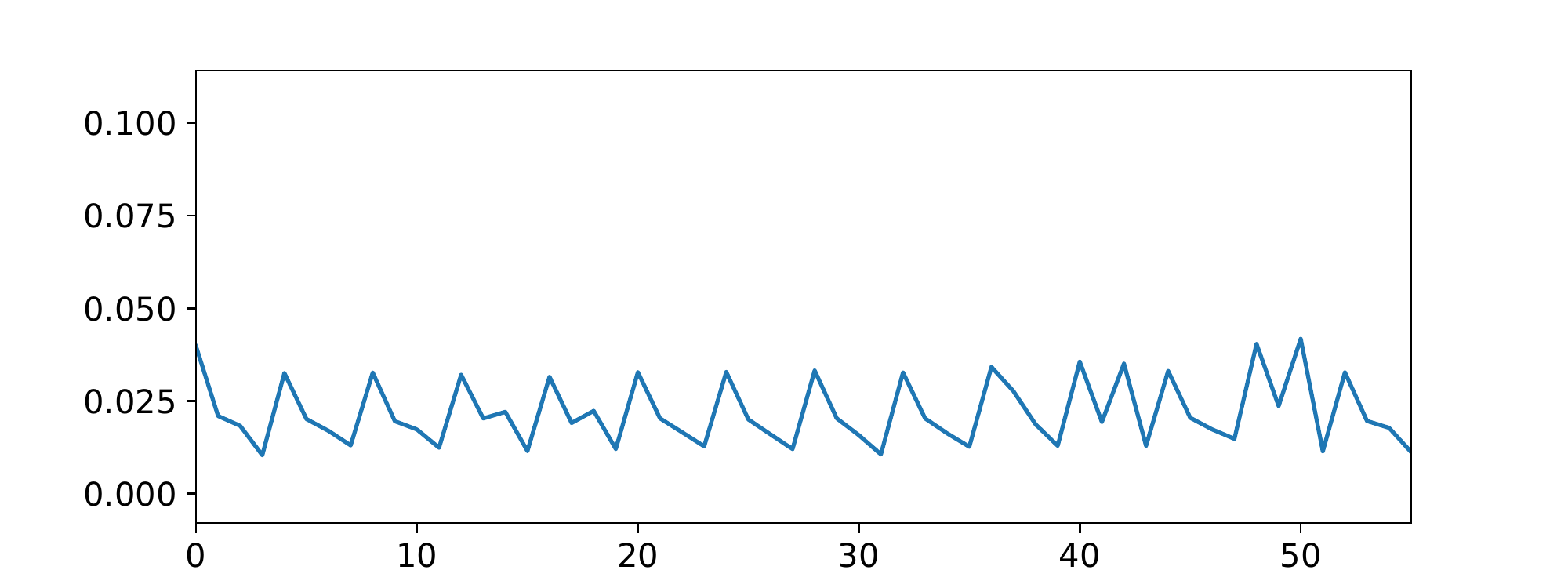}
		\put(-230,10){(b)}
	\end{minipage}
	\begin{minipage}{1.0\linewidth}
		\centering
		\includegraphics[width=0.22\linewidth]{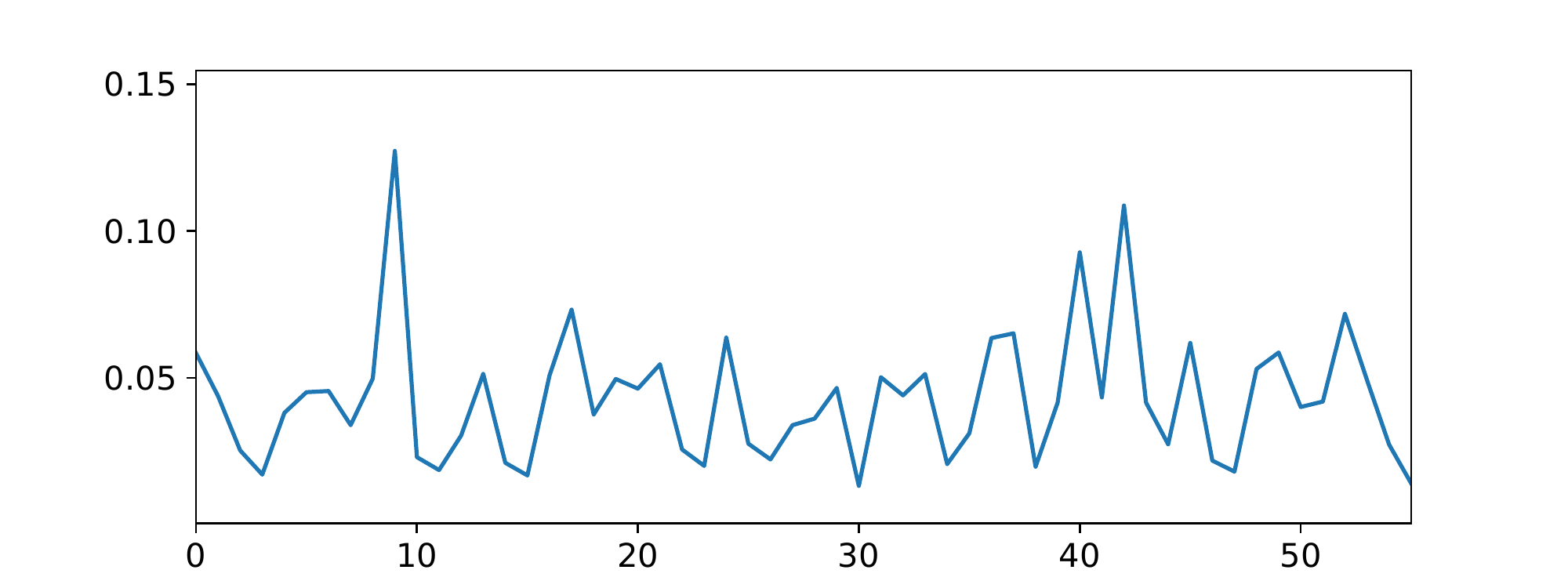}
		\includegraphics[width=0.22\linewidth]{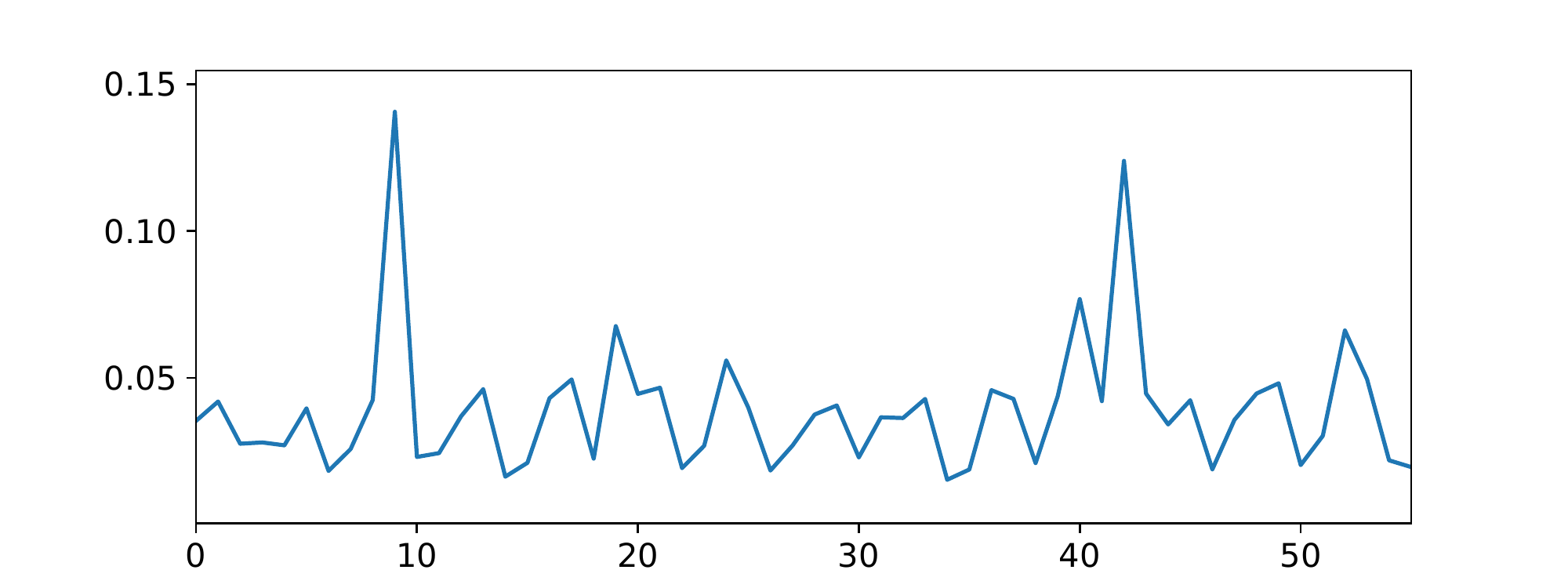}
		\includegraphics[width=0.22\linewidth]{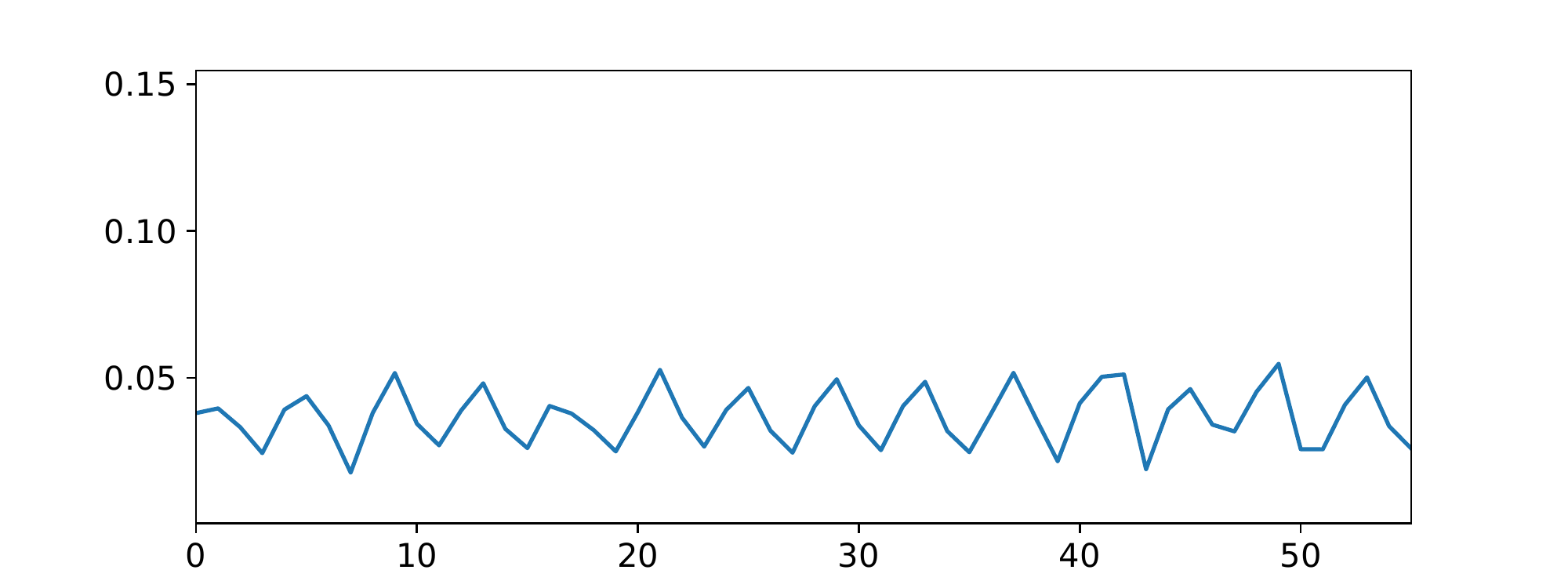}
		\includegraphics[width=0.22\linewidth]{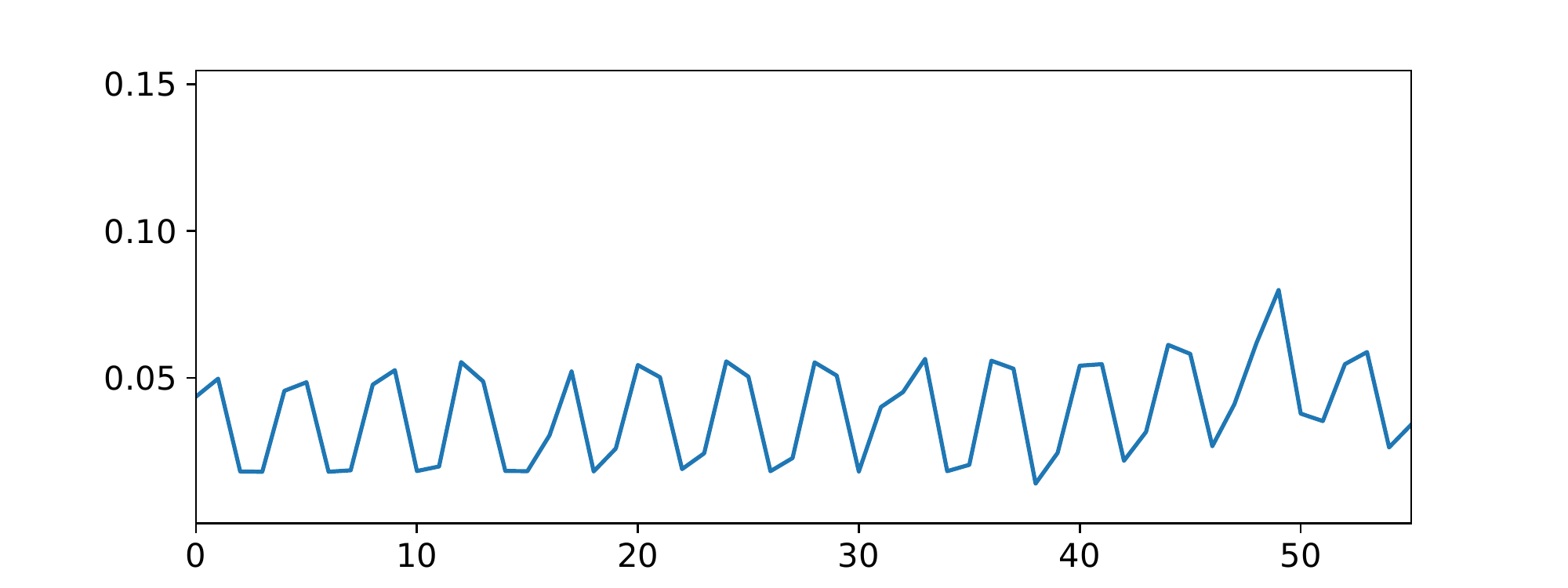}
		\put(-230,10){(c)}
	\end{minipage}
	
	\caption{Three time series samples selected uniformly at random from the synthetic dataset generated using \name~and the corresponding top-3 nearest neighbours (based on square error) from the real \fccshort~dataset. The time series shown here is traffic byte counter (normalized).}
	\label{fig:fcc-memorization}
\end{figure}

\myparatightest{\name~does not simply memorize training samples}
Figure \ref{fig:web-memorization}, \ref{fig:google-memorization}, \ref{fig:fcc-memorization} show the some generated samples from \name~and their nearest (based on squared error) samples in training data from the three datasets. The results show that \name~is not memorizing training samples. To achieve the good fidelity results we have shown before, \name~must indeed learn the underlying structure of the samples.

\section{Additional Case Study Results}
\label{app:downstream}

\myparatightest{Predictive modeling}
For the \wikishort~ dataset, the predictive modeling task involves forecasting of the page views for next 50 days, given those for the first 500 days.
We want to learn a (relatively) parsimonious model that can take an arbitrary length-500 time series as input and predict the next 50 time steps. 
For this purpose, we train various regression models: an MLP with five hidden layers (200 nodes each), and MLP with just one hidden layer (100 nodes), a linear regression model, and a Kernel regression model using an RBF kernel.
To evaluate each model, we compute the so-called \emph{coefficient of determination}, $R^2$, which captures how well a regression model describes a particular dataset.\footnote{For a time series with points $(x_i,y_i)$ for $i=1,\ldots, n$ and a regression function $f(x)$, $R^2$ is defined as
	$
	R^2 = 1 - \frac{\sum_i (y_i-f(x_i))^2}{\sum_i (y_i-\bar y)^2}
	$
	where $\bar y = \frac{1}{n}\sum_i y_i$ is the mean y-value.
	Notice that $-\infty \leq R^2 \leq 1$, and a higher score indicates a better fit.}

\begin{figure}[t]
	\centering
	\includegraphics[width=0.8\linewidth]{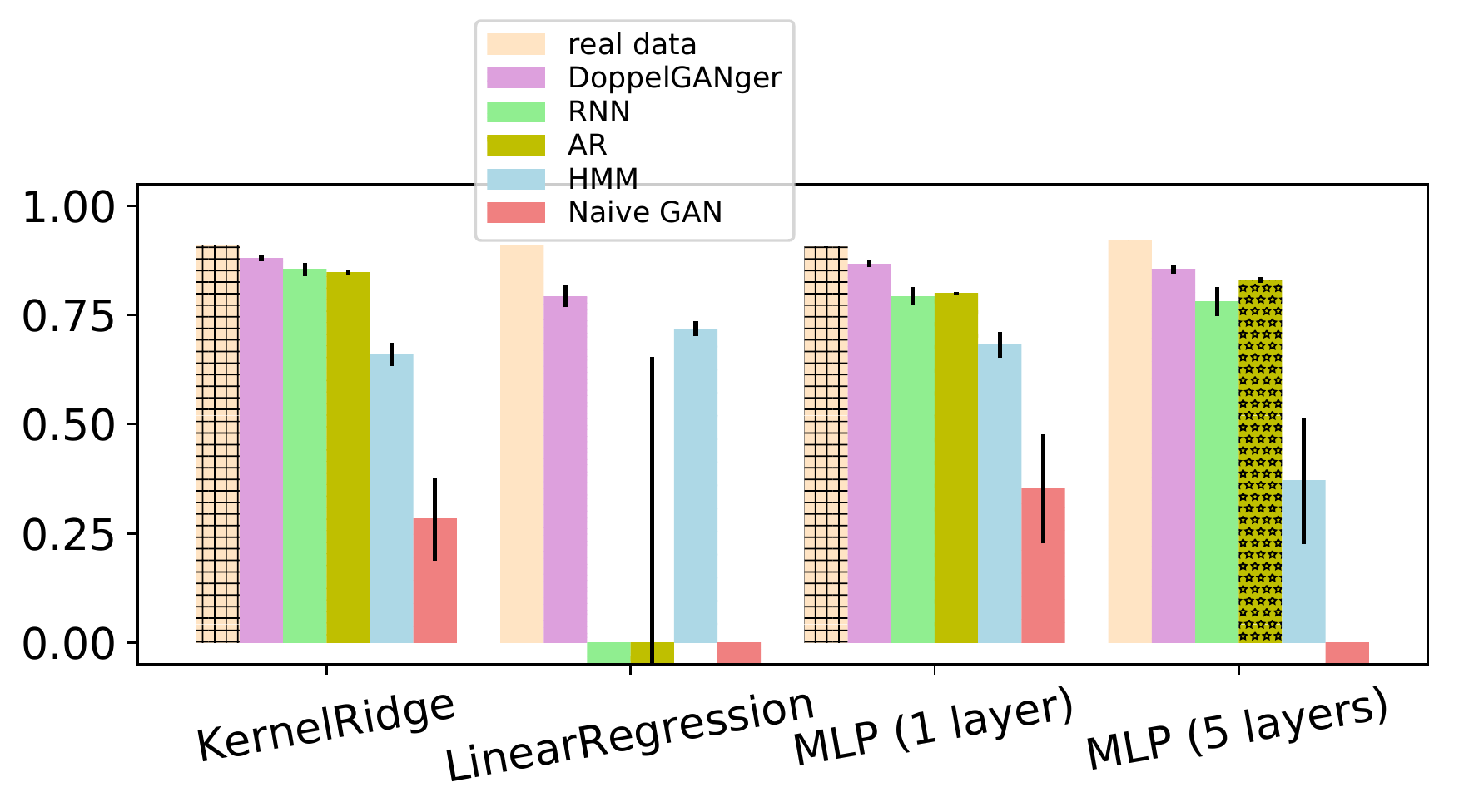}
	\caption{Coefficient of determination for \wikishort~time series forecasting. Higher is better.}
	\label{fig:web-acc}
\end{figure}

Figure \ref{fig:web-acc} shows the $R^2$ for each of these models for each of our generative models and the real data. Here we train each regression model on  generated data (B) and test it on real data (A'), hence it is to be expected that real data performs best. 
It is clear that \name~performs better than other baselines for all regression models. 
Note that sometimes RNN, AR, and naive GANs baselines have large negative $R^2$ which are therefore not visualized in this plot.

\begin{figure}[H]
	\centering
	\includegraphics[width=0.8\linewidth]{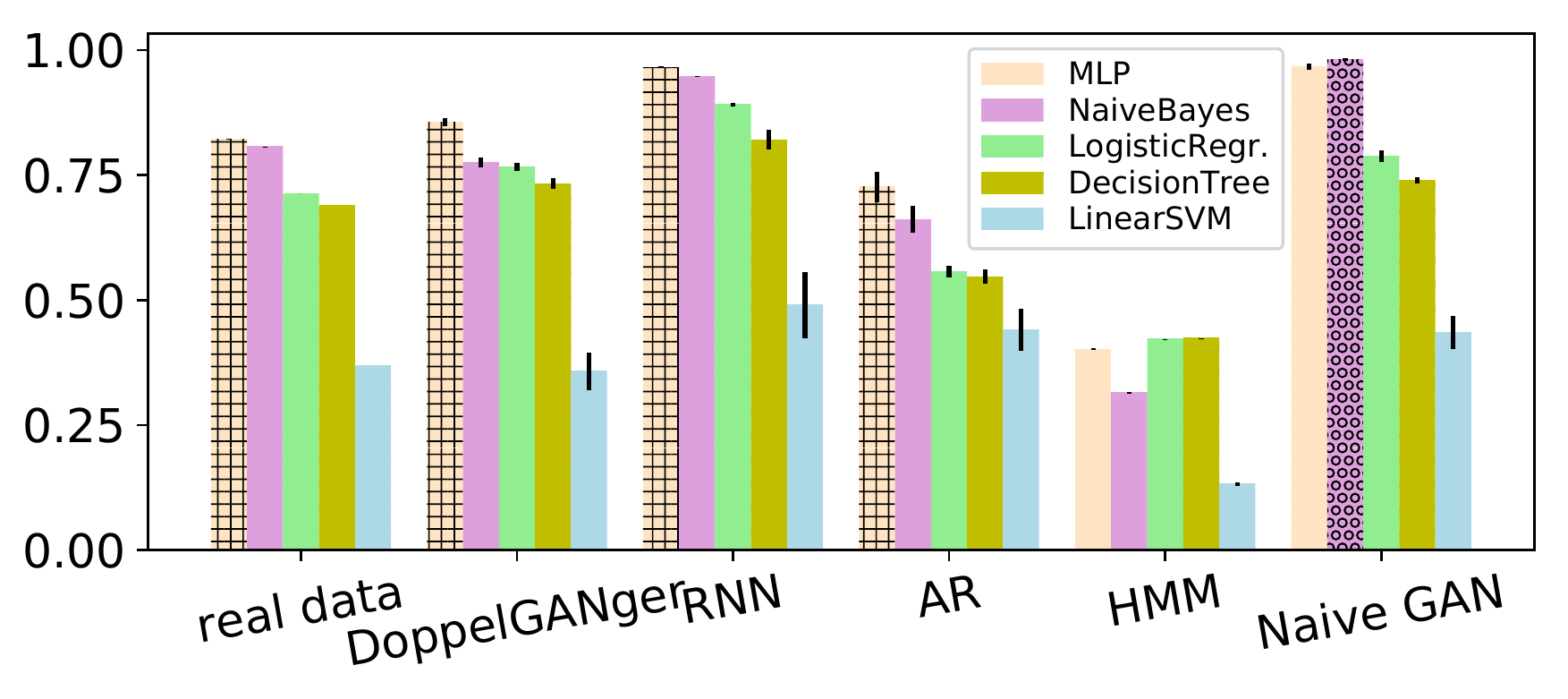}
	\caption{Ranking of end event type prediction algorithms on  \clustershort{} dataset.}
	\label{fig:google-rank}
\end{figure}

\begin{figure}[H]
	\centering
	\includegraphics[width=0.8\linewidth]{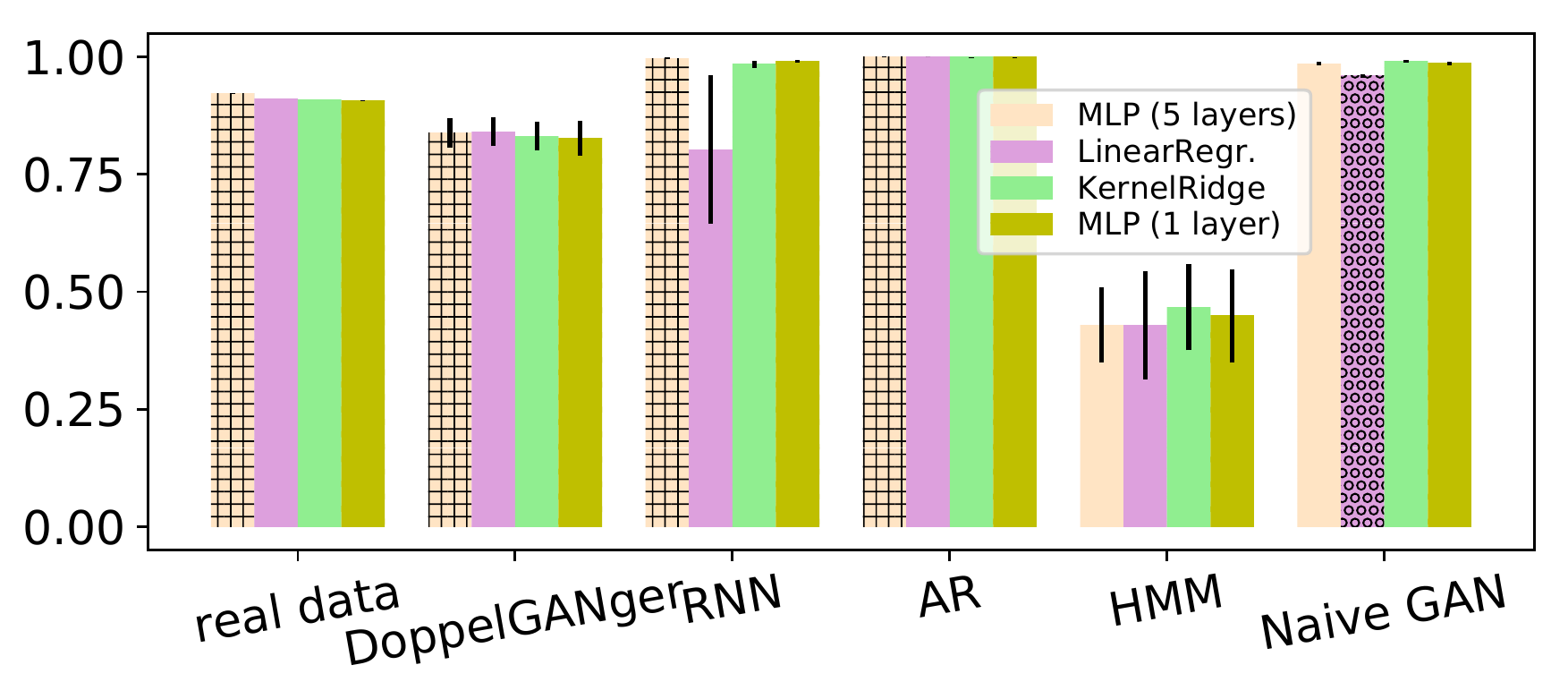}
	\caption{Ranking of traffic prediction algorithms on  \wikishort{} dataset.}
	\label{fig:web-rank}
\end{figure}

\myparatightest{Algorithm comparison}
Figure \ref{fig:google-rank}, \ref{fig:web-rank} show the ranking of prediction algorithms on \name's and baselines' generated data. Combined with \ref{tbl:rank}, we see that \name{} and AR are the best for preserving ranking of prediction algorithms.

\end{document}